\documentclass[3p,twocolumn]{elsarticle}

\usepackage{lineno,hyperref}
\usepackage{url}
\usepackage{hyperref}
\usepackage{graphicx}
\usepackage{subfigure}
\usepackage{booktabs}
\usepackage{threeparttable}
\usepackage{multirow}
\usepackage{url}
\usepackage{float}
\usepackage{amssymb}
\usepackage{amsmath}
\usepackage{amsmath}
\usepackage{bm}
\usepackage{blindtext}
\usepackage{mathrsfs}
\modulolinenumbers[5]

\journal{Neurocompting}

\begin{document}

\begin{frontmatter}

\title{DeepAVO: Efficient Pose Refining with Feature Distilling for Deep Visual Odometry}


\author{Ran Zhu}

\author{Mingkun Yang}

\author{Wang Liu}

\author{Rujun Song}
\author{Bo Yan}

\author{Zhuoling Xiao\corref{mycorrespondingauthor}}
\cortext[mycorrespondingauthor]{Corresponding author}
\ead{zhuolingxiao@uestc.edu.cn}

\address{School of Information and Communication Engineering}
\address{University of Electronic Science and Technology of China, Chengdu, 611731, China}


\begin{abstract}
The technology for Visual Odometry (VO) that estimates the position and orientation of the moving object through analyzing the image sequences captured by on-board cameras, has been well investigated with the rising interest in autonomous driving. This paper studies monocular VO from the perspective of Deep Learning (DL). Unlike most current learning-based methods, our approach, called DeepAVO, is established on the intuition that features contribute discriminately to different motion patterns. Specifically, we present a novel four-branch network to learn the rotation and translation by leveraging Convolutional Neural Networks (CNNs) to focus on different quadrants of optical flow input. To enhance the ability of feature selection, we further introduce an effective channel-spatial attention mechanism to force each branch to explicitly distill related information for specific Frame to Frame (F2F) motion estimation. Experiments on various datasets involving outdoor driving and indoor walking scenarios show that the proposed DeepAVO outperforms the state-of-the-art monocular methods by a large margin, demonstrating competitive performance to the stereo VO algorithm and verifying promising potential for generalization.
\end{abstract}


\end{frontmatter}

%\linenumbers

\section{Introduction}

From Unmanned Ground Vehicles (UGVs) to Micro Aerial Vehicles (MAVs), it is essential to know where autonomous robots are and to perceive the surrounding area. Global Positioning System (GPS) provides information about the position of the sensor in the world coordinate. However, a precise self-localization purely relying on the GPS is not sufficient for challenging environments like indoor scenarios and urban canyons. In this situation, a more precise measure or an alternative localization system is required in the real application for autonomous driving.

The camera is a small, light-weighted sensor that provides rich information about the environment around the sensing platform. Moreover, it can recover the ego-motion from image sequences by exploiting the consistency between consecutive frames \cite{fraundorfer2012visual}. Therefore, the concept of Visual Simultaneous Localization And Mapping (V-SLAM) and Visual Odometry (VO) are proposed to solve the well-known positioning problem, which estimates vehicles' position relative to its start point. As an essential task in robotics and computer vision communities, VO has been widely applied to various applications, ranging from autonomous driving and space exploration to virtual and augmented reality. From the perspective of the camera used, the VO methods consist of two types: stereo VO and monocular VO. This work aims at investigating the monocular VO, for a single camera is cheaper, lighter, and more general than a stereo rig. Especially when the ratio of stereo baseline to depth is minimal, the stereo VO degenerates to the monocular one.

Over the past thirty years, enormous work has been done to develop an accurate and robust VO system. The traditional VO algorithms can be divided into the feature-based method and the direct method. Feature-based methods typically consist of camera calibration, feature detection, feature matching, outlier rejection (e.g., RANSAC), motion estimation, scale estimation, and optimization (e.g., Bundle Adjustment). Unfortunately, how to detect appropriate features for recovering specific motions remains a challenging problem. Handcrafted feature descriptors such as SIFT \cite{Lowe2004Distinctive}, ORB \cite{2012ORB} designed for general visual tasks, lack the response to motions. Instead, extra information guided by geometric prior such as planar structures \cite{2017Visual} and vanishing points \cite{2015Real}, is used for camera pose estimation in specific environments, providing promising performance but limited generalization ability. Unlike feature-based methods, direct methods track the motion of the pixel and obtain pose prediction by minimizing the photometric error, so it is extremely vulnerable to light changes. Moreover, the absolute scale estimation in the traditional monocular VO must use some extra information (e,g., the height of the camera) or prior knowledge.

The emerging Deep Learning (DL), a data-driven approach, has yielded impressive achievement in computer vision. Rather than handcrafted features, DL that has the ability to extract deep features from the plain input, encodes the high-level priors to regress camera poses. Compared with traditional VO, learning-based VO has the advantage of low computation cost and no need for internal camera parameters. A few methods on DL have been proposed for camera motion recovery, such as DeepVO \cite{wang2017deepvo}, ESP-VO \cite{wang2018end}, NeuralBundler \cite{2019Pose}, CL-VO \cite{2019Learning}, DAVO \cite{2020Dynamic}. While achieving promising performances, they do not take into account the different responses of visual cues and the effect of pixels movement in different directions in the input image to the camera motion, thus may output trajectories with large error. For learning-based VO, it should focus more on geometric constraints than the ``appearance'' information when harnessing Convolutional Neural Networks (CNNs) to extract features. Optical flow, as the representation of the geometric structure, has been proved useful for estimating Frame to Frame (F2F) ego-motion. \cite{muller2017flowdometry} takes the raw optical flow calculated by Flownet \cite{dosovitskiy2015flownet} as the input of the pose prediction network, which adopts the structure of FlowNetS as the underlying CNN. Therefore, we take the optical flow as input to the proposed model.

Guided by the previous considerations, we explore a novel strategy for performing visual ego-motion estimation in this work. Inspired by P-CNN VO\cite{costante2015exploring}, we extend the neural network into four branches focusing on pixels movement in different directions in the optical flow and then regress the global feature concatenated from the four outputs to obtain F2F motion estimation. In particular, features extracted by each branch have been distilled by using the attention mechanism to refine estimation. In this paper, many quantitative and qualitative experiments in terms of precision, robustness, and computation speed are conducted. The results demonstrate that the proposed model outperforms many current monocular methods and provides a competitive performance against the classic stereo VO. In summary, our key contributions are as follows:

\begin{itemize}
\item Novel visual perception guiding ego-motion estimation: By considering the four quadrants in optical flow and fusing the distilling module into each branch encoder, the learning-based DeepAVO model pays more attention to the visual cues that are effective for ego-motion estimation.

\item Lightweight VO framework with enhanced tracking performance: The proposed DeepAVO model framework yields more robust and accurate results compared with competing monocular VOs. The F2F VO calculation can be done within 12 ms, making it practical and valuable in real-world applications.

\item Extensive fresh scenes validation: The DeepAVO produces promising pose estimation and maintains high-precision tracking results on various datasets involving outdoor driving and indoor walking scenarios. Outstanding improvements in the accuracy and robustness of VO are further demonstrated.
\end{itemize}

Our method outperforms state-of-the-art learning-based methods and produces competitive results against classic algorithms. Additionally, it works well in the new dataset, where learning-based algorithms tend to fail due to different feature characteristics. The rest of this paper is organized as follows: Section \ref{sec:related works} reviews some related works, and Section \ref{sec:approach} describes the proposed architecture in detail. The performance of our approach is compared with many current methods in Section \ref{sec:experiments}. Finally, we conclude the paper in Section \ref{sec:conclusion}.

\section{Related Works}
\label{sec:related works}
Visual odometry has been studied for decades, and many excellent approaches have been proposed. In this section, we discuss various algorithms and their differences from others. There are mainly two types of algorithms in terms of the technique and framework adopted: geometry-based and learning-based methods.

\subsection{Methods based on Geometry}
Traditionally, the VO problem that relies on geometric constraints extracted from imagery can be solved by minimizing reprojection errors or photometric errors. Thus, they can be further categorized into feature-based and direct methods.

\subsubsection{Sparse Feature based Methods}
The standard approach is to extract a sparse set of salient features(e.g., points, lines) in each image; match them in successive frames, such as the algorithms in ORB-SLAM2 \cite{mur2017orb} and LIBVISO2 \cite{Geiger2012Stereoscan}; robustly recover camera motion using epipolar geometry; finally, refine the pose through reprojection error minimization. The majority of traditional VO algorithms \cite{SCARAMUZZA2011Visual} follows this procedure, independent of the applied optimization framework. A reason for the success of these methods is the availability of robust feature detectors and descriptors that allow matching between images even at the large inter-frame movement.

\subsubsection{Direct Methods}
Feature extraction and matching that are key to determining the performance of sparse feature-based methods are computationally expensive. However, outliers and mismatch often cause VO algorithms to suffer from drifts over time. Direct Methods \cite{Irani1999About} estimate structure and motion directly from the intensity values in consecutive images under the assumption of photometric consistency, e.g., DTAM in \cite{2011DTAM}, DSO in \cite{wang2017stereo}. The local intensity gradient magnitude and direction are used in the optimization compared to sparse feature-based methods that only use salient features without benefiting from rich information in the whole image. Besides, semi-direct approaches achieve promising performance in the monocular VO \cite{Engel2016Direct}\cite{6906584}, which uses feature-correspondence to avoid time cost of feature extraction from each frame and increase accuracy in texture-less environments.
\begin{figure*}[!th]
	\centering
	\includegraphics[width=0.95\textwidth,height=0.5\textwidth]{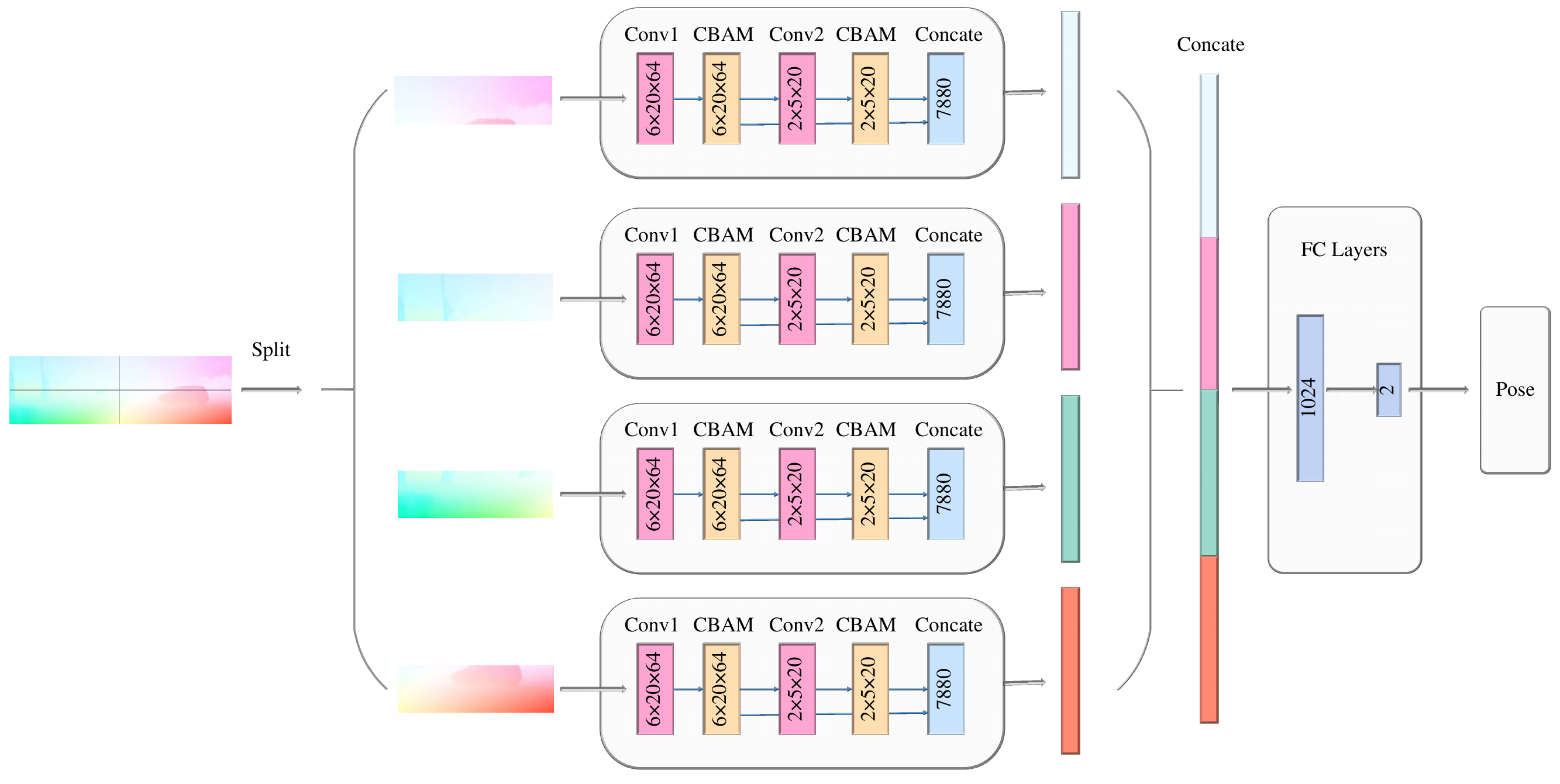}
	\caption{The architecture of the proposed DeepAVO based monocular VO system. In this figure, the details in our system are described. Note that an average pooling operation is omitted before feeding the four parts of optical flow into CNNs.}
	\label{fig:network architecture}
\end{figure*}

\subsection{Methods based on Learning}
Taking advantage of an overwhelming availability of data, DL is utilized to learn motion model and explore VO from sensor readings with deep learning techniques. Many approaches without explicitly applying geometric theory have been proposed to deal with the challenges in the classic monocular VO systems, such as feature extraction, depth estimation, scale correction, and data association.

Some work based on Machine Learning (ML) techniques has been proposed to solve the monocular VO problem. Taking optical flow data as input, \cite{2008Memory} that first tries to apply learning methods in solving the VO problem trains a K Nearest Neighbor (KNN) regressor for the monocular VO. \cite{2014Evaluation} proposes the SVR VO to regress ego-motion leveraging Support Vector Machine (SVM) by introducing Gaussian Processes (GP), of which the performance is far behind traditional methods. However, it has been widely demonstrated that traditional ML techniques are inefficient when encountering large or highly non-linear high-dimensional data. DL that automatically learns suitable feature representation from the large-scale dataset, provides more promising performance. In this paper, we mainly focus on DL-based monocular VO works.

\subsubsection{Unsupervised Methods}
Mimicking the conventional structure from motion, a number of algorithms that deal with the VO problem in an unsupervised manner have emerged. SfmLearner \cite{zhou2017unsupervised} recovers the depth of scenes and ego-motion from unlabeled sequences with view synthesis using photometric error as supervisory signals. Its successor \cite{2017UnDeepVO} extends this work to take stereo image pairs as input and recovers the absolute scale with the known camera baseline. GeoNet \cite{yin2018geonet} proposes an unsupervised learning framework for jointly estimating monocular depth, optical flow, and camera motion from video. NeuralBundler \cite{2019Pose} introduces a hybrid VO system that combines an unsupervised monocular VO with a pose graph optimization back-end. D3VO \cite{yang2020d3vo} incorporates the deep predictions of depth, pose, and uncertainty into a direct visual odometry and defeats several popular conventional VO/VIO systems, such as DSO \cite{wang2017stereo}, VINS-Mono \cite{qin2018vins}.

These unsupervised methods learn from large amounts of unlabeled data. Although it breaks through the limitation of the requirement for large amounts of labelled data in supervised learning, it can only process a limited number of consecutive frames due to the fragility of photometric losses, resulting in high geometric uncertainty and severe error accumulation.

\subsubsection{Supervised Methods}
Recently, DL techniques such as CNNs and RNNs have been utilized for pose estimation. DeMoN \cite{Ummenhofer2016DeMoN} jointly estimates depth and motion from two consecutive images by formulating structure from motion as a supervised learning problem. \cite{muller2017flowdometry} takes the raw optical flow calculated by Flownet \cite{dosovitskiy2015flownet} as the input of the pose prediction network, which adopts the structure of FlowNetS as the underlying CNN. P-CNN VO \cite{costante2015exploring} exploits the best visual features and proposes a VO, which outperforms other contemporary methods. Moreover, it is robust for the blur, luminance, and contrast anomalies conditions. DeepVO \cite{wang2017deepvo} recovers camera poses from image sequences by harnessing LSTM \cite{Hochreiter1997Long} to learn historical information for current motion prediction. Based on DeepVO, ESP-VO \cite{wang2018end} extends into a unified framework to directly infer poses and uncertainties. CL-VO \cite{2019Learning} introduces Curriculum Learning strategy for learning the geometry of monocular VO by gradually making the learning objective more difficult during training. DAVO \cite{2020Dynamic} dynamically adjusts the attention weights on different semantic categories for different motion scenarios to estimate the ego-motion of a monocular camera.

The methods above take the visual cues in the whole image equally. However, the movement characteristics of different parts in images captured by the camera and the attention to motion features extracted by the network are ignored.

\section{System Model}
\label{sec:approach}

In this section, we introduce our framework (Fig. \ref{fig:network architecture}) in detail. Considering the significance of geometric structure for the VO task, we calculate the optical flow discussed in \ref{subsec:optical flow calculation} from the consecutive RGB images. The \emph{Encoder} module in \ref{subsec:encoder} extracts high-level representations, which are further distilled by the attention mechanism in \ref{subsec:distilling}. We design the loss function considering both the rotational and translational errors in \ref{subsec:loss_function}.

\subsection{Optical Flow Calculation}
\label{subsec:optical flow calculation}
The essence of the ego-motion estimation is quite different from other computer vision tasks, which focuses more on geometric motion between images in the video. To ensure that the proposed framework could learn geometric feature representations, optical flow calculation from consecutive images is conducted. The optical flow depicts the pixel movement in the image captured by the vehicle-mounted camera. In optical flow, the image from the camera changes over time, and the image can be seen as a function of time: $\bm{I}(t)$. Then, for a pixel located at $(x,y)$ at time $t$, its intensity value (i.e., the grayscale) can be written as $\bm{I}(x,y,t)$.

The optical flow calculation is based on the assumption of photometric consistency. That is, the pixel intensity value of the same spatial point is fixed in each image. For the pixel located at $(x,y)$ at time $t$, supposing that it moves to $(x+\mathrm{d}x, y+\mathrm{d}y)$ at time $t+\mathrm{d}t$, it has:
\begin{equation}\label{2}
  \bm{I}(x+\mathrm{d}x, y+\mathrm{d}y, t+\mathrm{d}t) = \bm{I}(x, y, t).
\end{equation}

We can perform the first-order Taylor expansion on the left side of Equation (\ref{2}):
\begin{equation}\label{3}
  \bm{I}(x+\mathrm{d}x, y+\mathrm{d}y, t+\mathrm{d}t) \approx \bm{I}(x, y, t) + \frac{\partial{\bm{I}}}{\partial{x}}\mathrm{d}x + \frac{\partial{\bm{I}}}{\partial{y}}\mathrm{d}y + \frac{\partial{\bm{I}}}{\partial{t}}\mathrm{d}t.
\end{equation}Based on the photometric consistency, the grayscale at the next moment is equal to the previous, thus:
\begin{equation}\label{4}
\frac{\partial{\bm{I}}}{\partial{x}}\mathrm{d}x + \frac{\partial{\bm{I}}}{\partial{y}}\mathrm{d}y + \frac{\partial{\bm{I}}}{\partial{t}}\mathrm{d}t = 0.
\end{equation}Divide by $dt$, Equation (\ref{4}) is further formulated as:
\begin{equation}\label{5}
  \frac{\partial{\bm{I}}}{\partial{x}}\frac{\mathrm{d}x}{\mathrm{d}t} + \frac{\partial{\bm{I}}}{\partial{y}}\frac{\mathrm{d}y}{\mathrm{d}t} = -\frac{\partial{\bm{I}}}{\partial{t}}
\end{equation}where $\frac{\mathrm{d}x}{\mathrm{d}t}$ and $\frac{\mathrm{d}y}{\mathrm{d}t}$ are the moving speed of pixels on the x-axis and y-axis, respectively, denoted as $u$, $v$. $\frac{\partial{\bm{I}}}{\partial{x}}$ is the gradient of the image in the x-axis direction at this point and the other term $\frac{\partial{\bm{I}}}{\partial{y}}$ is the gradient in the y-axis direction, denoted as $\bm{I}_{x}$, $\bm{I}_{y}$, respectively. $\bm{I}_{t}$ is the change of the image grayscale with respect to time. Equation (\ref{5}) can be written in a matrix:
\begin{equation}\label{6}
  [\bm{I}_{x}, \bm{I}_{y}]\begin{bmatrix} u\\v \end{bmatrix} = -\bm{I}_{t}.
\end{equation}

\begin{figure}
	\centering
	\subfigure[Frame 1]{
		\label{fig:f1}
		\includegraphics[width=0.226\textwidth,height=0.14\columnwidth]{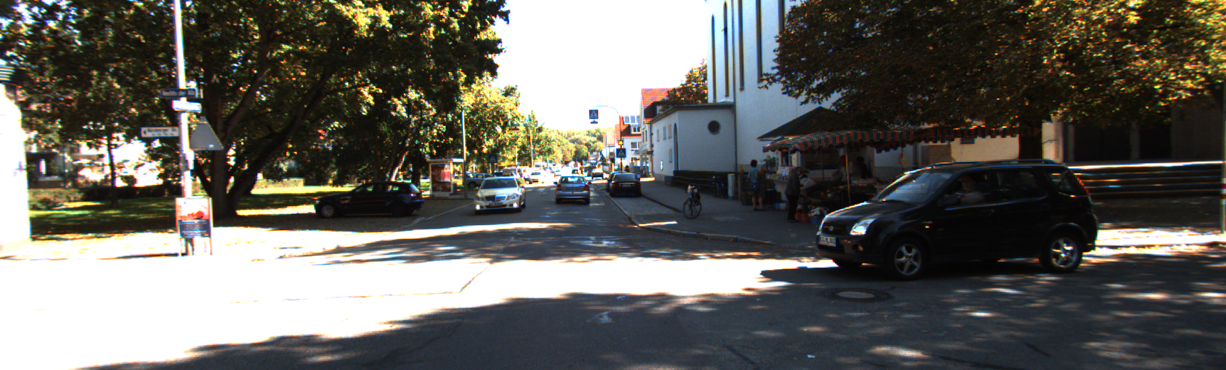}}
	\subfigure[Frame 2]{
		\label{fig:f2}
		\includegraphics[width=0.226\textwidth,height=0.14\columnwidth]{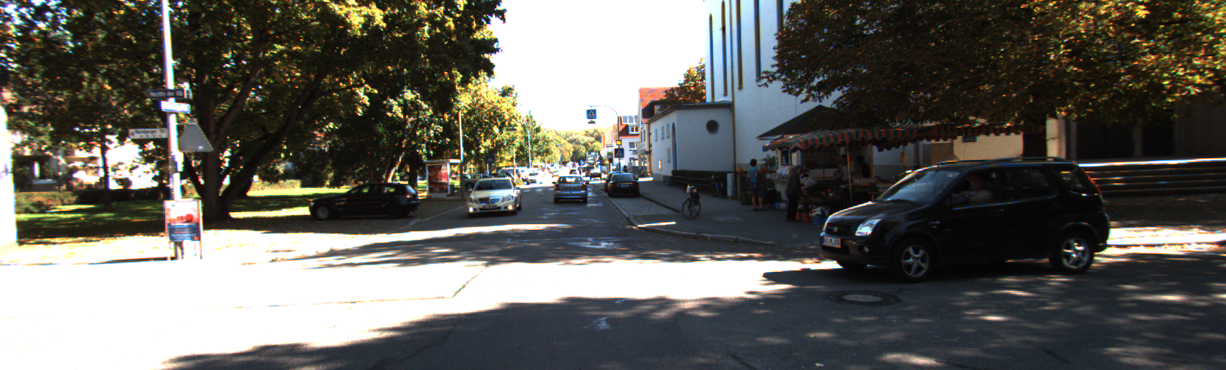}}
	\subfigure[Dense optical Flow]{
		\label{fig:denseflo}
		\includegraphics[width=0.226\textwidth,height=0.14\columnwidth]{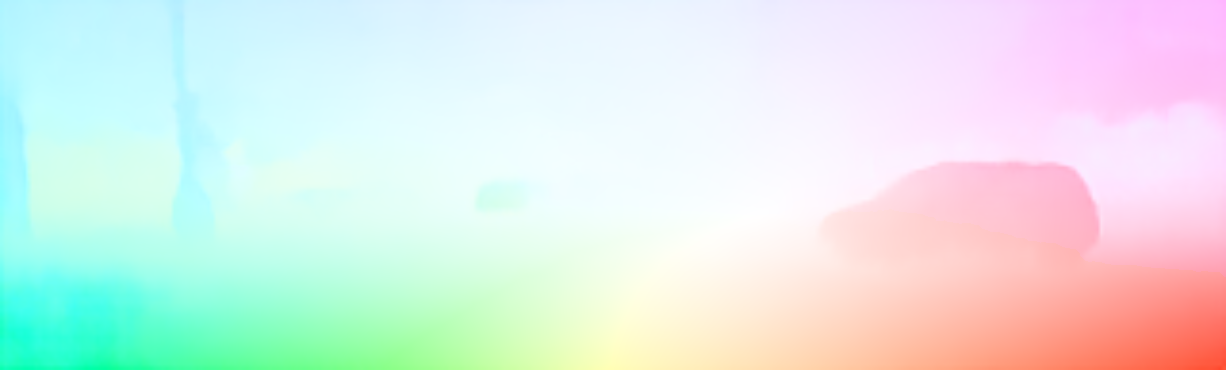}}
	\subfigure[Parse optical Flow]{
		\label{fig:sparseflo}
		\includegraphics[width=0.226\textwidth,height=0.14\columnwidth]{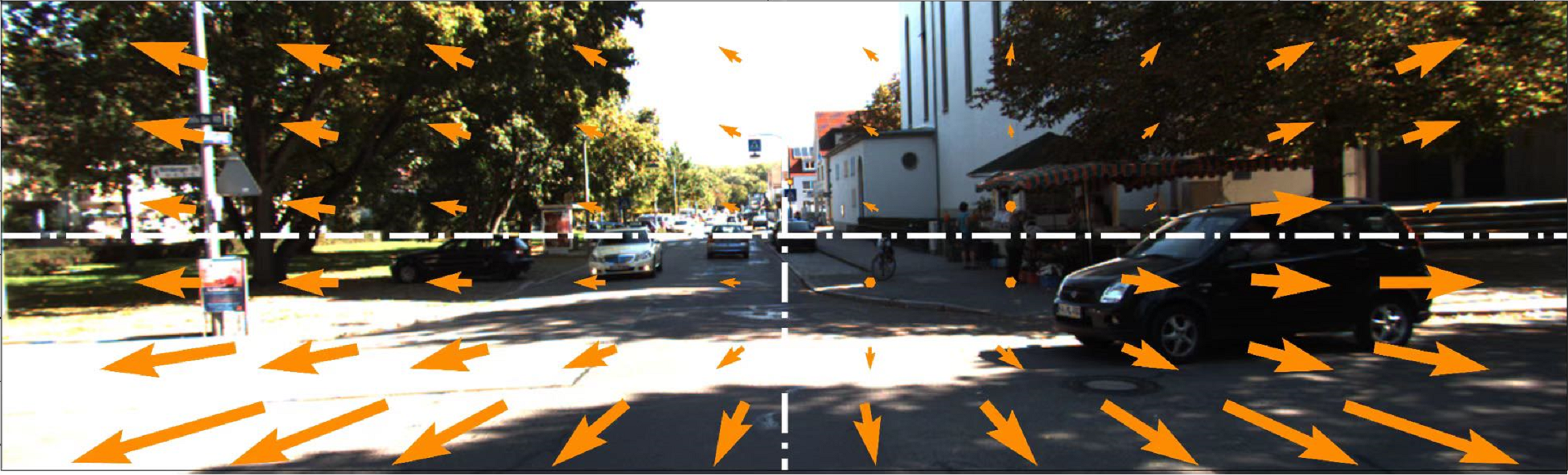}}
	\caption{Original frames and visualization of optical flow. (a) and (b) are the two frames in the KITTI Seq 08. (c) and (d) is the corresponding dense optical flow and sparse optical flow acquired from PWC-Net.}
    \label{Fig.optical flow}
\end{figure}

In order to calculate the pixel motion $u$, $v$, the traditional method is to find the least squares solution by introducing the Lucas-Kanade (LK) method. In this way, we can get the moving speed of pixels between images.

However, traditional optical flow algorithms for high-precision VO are widely applied, while most of them are computationally intense and cannot meet the real-time requirements of the system. Considering the performance of the proposed model and the network calculation, we utilize a learning-based optical flow extractor PWC-Net\cite{sun2018pwc}, which is known as a compact but effective CNN model using simple and well-established principles: pyramidal processing, wrapping, and the use of a cost volume. Not only does PWC-Net reduce the model size, but it also improves performance. We use the Pytorch version of the network framework released by the original paper\cite{sun2018pwc} to calculate the pixel motion, as shown in Fig. \ref{Fig.optical flow}. The process can be described as:
\begin{equation}\label{10}
  \bm{Flo}_{t} = \mathcal{F}(\bm{i}_{t-1}, \bm{i}_{t})
\end{equation}where $\bm{Flo}_{t}{\in}\mathbb{R}^{C{\times}H{\times}W}$ denotes the optical flow at time $t$ by function $\mathcal{F}$ from two consecutive images $\bm{i}_{t-1}$ and $\bm{i}_{t}$. $H$, $W$, and $C$ represent the height, width, and channel of obtained optical flow where $C = 2$.

\subsection{Encoder}
\label{subsec:encoder}

\begin{figure}[!h]
	\centering
	\includegraphics[width=0.5\textwidth,height=0.29\textwidth]{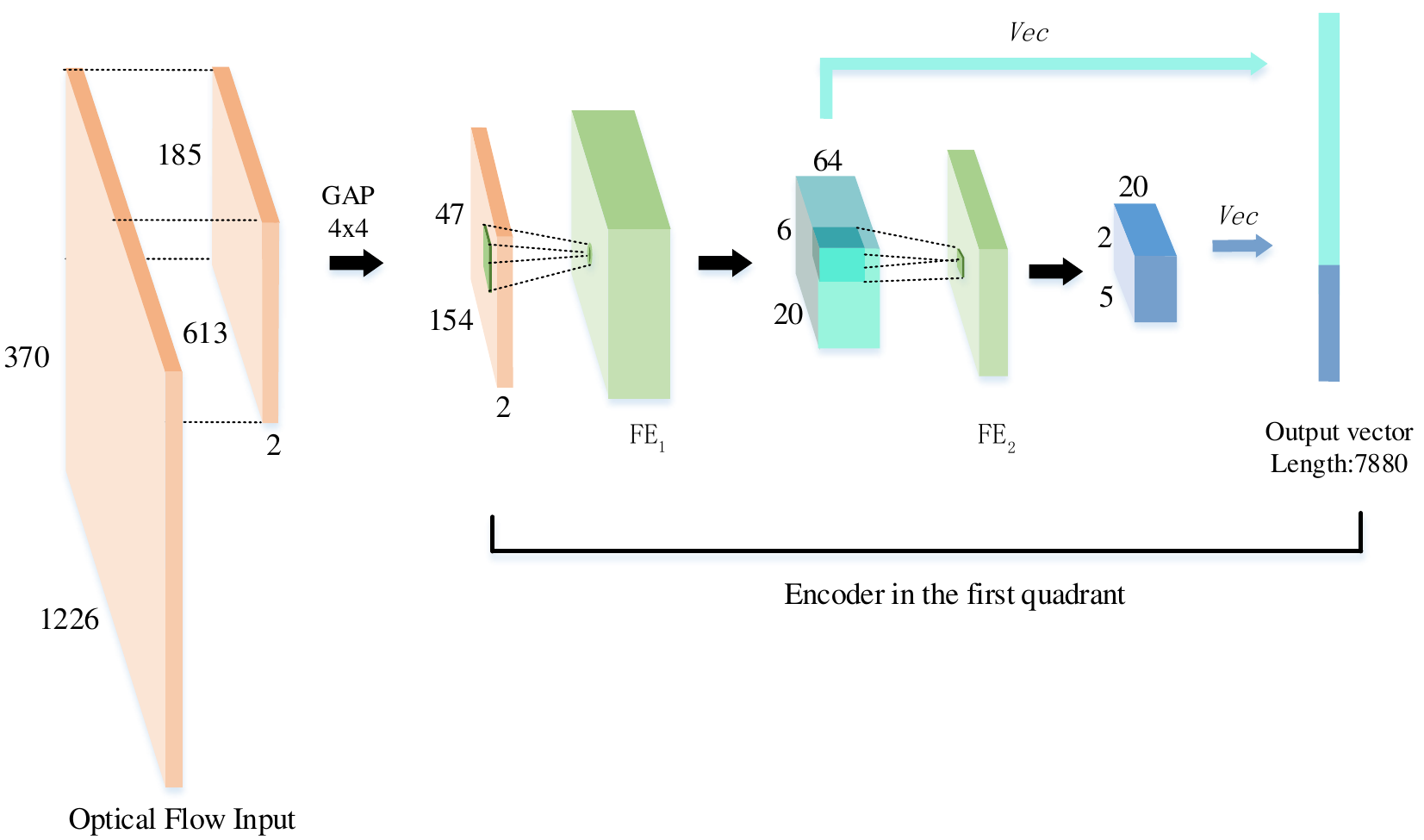}
	\caption{The core architecture of the proposed network. The image is divided into four quadrants, and each one passes through a chain of feature extractors ($FE_{1}$, $FE_{2}$). To produce more robust visual features, we concatenate the output of $FE_{1}$ and $FE_{2}$.}
	\label{fig:encoder}
\end{figure}

While many state-of-the-art models (e.g., VGGNet\cite{simonyan2014very}, ResNet\cite{he2016deep}, and GoogleNet\cite{szegedy2015going}) have yielded remarkable performance in computer vision tasks, such as image classification, motion recognition, it is impractical to simply adopt them to the VO task rooted in the geometry of images. The VO task is on the basics of geometric constraints between video frames, so the devised neural network should concern itself with pixel motion characteristics in optical flow. 

For the image sequences captured by the on-board camera, the pixel movement at the edge of the image is more intense, as shown in Fig. \ref{fig:sparseflo}, and can be roughly divided into four directions. Therefore, in the $\emph{Encoder}$, four parallel CNNs of the proposed DeepAVO are responsible for focusing on the pixel motion in different directions to exploit local visual cues.

To balance the performance and computation complexity of the model, each quadrant is down-sampled $4$ times by using the $Global$ $Average$ $Pooling$ ($GAP$) and then fed into a series of CNN filters to extract motion features. Each branch contains the same core architecture shown in Fig. \ref{fig:encoder}, and the detailed configuration is outlined in Table. \ref{tab:t1}. Four parallel core neural networks are trained simultaneously as a whole DeepAVO. Two blocks of the core architecture, to be specific, extract features in different levels: $FE_{1}$ extracts the  coarser ones and $FE_{2}$ extracts the finer details. The output of two blocks are concatenated as the final feature map of the branch:
\begin{equation}\label{11}
\begin{split}
  X_{t}^{i} = Vec\ (FE_{1}^{i}\ (\bm{Flo}_{t}^{i})) \oplus Vec\ (FE_{2}^{i}\ (FE_{1}^{i}\ (\bm{Flo}_{t}^{i}))), \\ i =1, 2, 3, 4
\end{split}
\end{equation}where $Vec$ reshapes a $3D$ feature map into a vector for following concatenation operation $\oplus$. $X_{t}^{i}$ denotes the feature vector that is encoded from the optical flow $\bm{Flo}_{t}^{i}$ in the corresponding $ith$ quadrant at time $t$. $FE_{1}^{i}$ and $FE_{2}^{i}$ denote two feature extractors of the $ith$ branch, respectively.

\begin{figure*}[!ht]
	\centering
	\includegraphics[width=\textwidth,height=0.16\textheight]{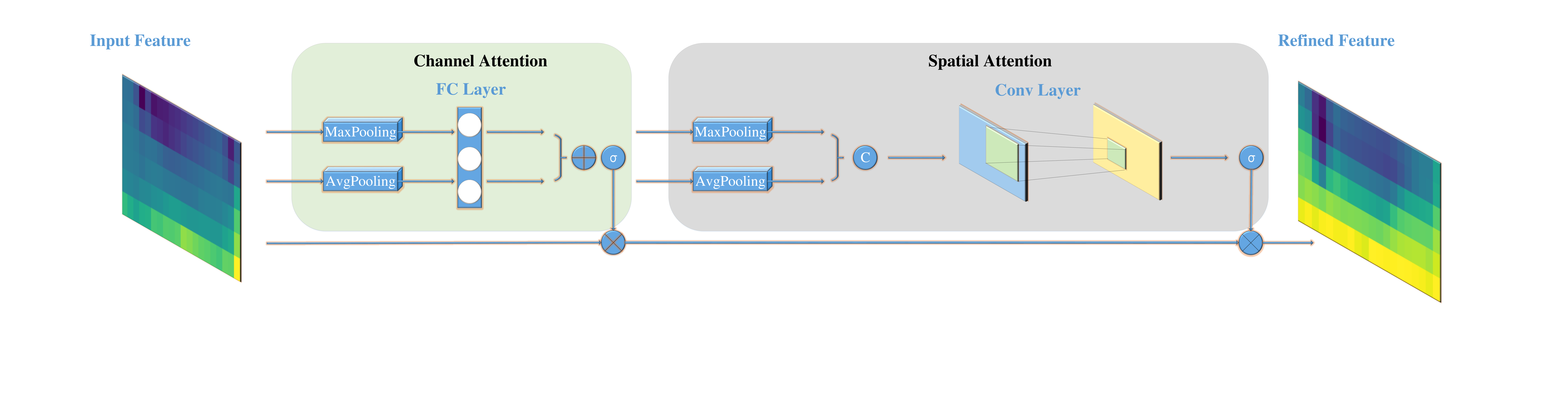}
	\caption{The overview of CBAM \cite{woo2018cbam}. The mechanism has two sequential sub-modules: channel and spatial. The intermediate feature map is adaptively refined using this mechanism at each $FE$ block in each branch. $\sigma$ means the sigmoid function, and C denotes concatenation operation.}
	\label{fig:CBAM}
\end{figure*}

While Four quadrants depict the same motion, the pose estimation can not rely on a single quadrant because the limited motion information in one quadrant causes the ambiguity between simple turning and forward movement. Hence, we concatenate four branches outputs into a feature vector containing the global information. The fully connected layers, shown in Fig. \ref{fig:network architecture}, give the F2F pose prediction using features of all four quadrants at the same resolutions.

\begin{table}[!h]
\renewcommand{\arraystretch}{1.2} 
\setlength{\tabcolsep}{0.4mm}
\caption{Configuration of each branch CNN.}
\begin{tabular}{lcccc}
\hline
Layer & \multicolumn{1}{l}{\begin{tabular}[c]{@{}l@{}}Receptive\\ Field Size\end{tabular}} & \multicolumn{1}{l}{Padding} & \multicolumn{1}{l}{Stride} & \multicolumn{1}{l}{\begin{tabular}[c]{@{}l@{}}Number of\\ Channels\end{tabular}} \\ \hline
GAP            & 4 $\times$ 4                                                                                       & 2                                    & 4                                   & 2                                                                                         \\
Conv1          & 9 $\times$ 9                                                                                       & 4                                    & 2                                   & 64                                                                                        \\
Avgpooling1    & 4 $\times$ 4                                                                                       & 2                                    & 4                                   & 64                                                                                        \\
Conv2          & 3 $\times$ 3                                                                                       & 1                                    & 2                                   & 20                                                                                        \\
Avgpooling2    & 2 $\times$ 2                                                                                       & 1                                    & 2                                   & 20                                                                                        \\ \hline
\end{tabular}
\label{tab:t1}
\end{table}

\subsection{Distilling}
\label{subsec:distilling}

In terms of the image processing domain, the attention mechanism is proposed originally by DeepMind ("recurrent models of visual attention") for image classification \cite{mnih2014recurrent}. It improves the performance of the model by reducing the dependence on external information and capturing the internal correlation of data or features. 

The information redundancy of high-dimensional feature extracted by learning-based methods often leads to the lack of attention to essential information and the suppression of useless information. This always leads to unsatisfactory performance on learning tasks. Based on this problem, this paper introduces an attention mechanism to solve it. For the VO task, the attention mechanism enables the model to concentrate on pixels in distinct motion. Correspondingly, the weight of features in the foreground and blurred part is decreased. Our approach benefits from effective feature learning by incorporating an attention module to selectively distill features from the channel and  spatial dimensions for current F2F pose inference.

There are many attention mechanisms, such as CBAM\cite{woo2018cbam}, SENet\cite{hu2018squeeze}, and Non-local neural networks\cite{wang2018non} (Nloc). Among them, SENet improves the representation ability of the model by modelling the relationship between channels, that is, assigning weights to the various channel features extracted by the previous layer. CBAM that adds the spatial attention mechanism on the basis of SENet, focuses on essential features and restrains unnecessary ones to refine the distribution and processing of information. Nloc directly integrates global information, bringing richer semantic information to the following layers, but it will increase computation. The ablation experiments on different attention mechanisms in Section \ref{sec:experiments} show that the proposed architecture combined with CBAM performs better.

CBAM, as a dual attention mechanism, generates the factors to recalibrate feature map in both the channel domain and spatial domain, as shown in Fig. \ref{fig:CBAM}. This process can be described as two operations:
\begin{equation}\label{12}
\bm{M}' = {\sigma}(MLP(AP(\bm{M}))+MLP(MP(\bm{M})))  \odot  \bm{M}
\end{equation}
\begin{equation}\label{13}
\bm{M}'' = {\sigma}(f^{7{\times}7}[AP(\bm{M}'), MP(\bm{M}')])  \odot  \bm{M}'
\end{equation}where $\odot$ denotes element-wise multiplication, ${\sigma}$ is the sigmoid function, $f^{7{\times}7}$ is a $7{\times}7$ convolutional layer, $AP$, $MP$, and $MLP$ mean average pooling, max pooling, and a dense layer. $\bm{M}{\in}\mathbb{R}^{C{\times }H{\times}W}$ is a feature map. $\bm{M}'{\in}\mathbb{R}^{C{\times}H{\times}W}$  and $\bm{M}''{\in}\mathbb{R}^{C{\times}H{\times}W}$ are the channel-refined and spatial-refined feature maps, respectively.

\begin{figure}[!h]
	\centering
	\subfigure[The first quadrant of Frame 1]{
		\label{Fig.img}
		\includegraphics[width=0.226\textwidth,height=0.14\columnwidth]{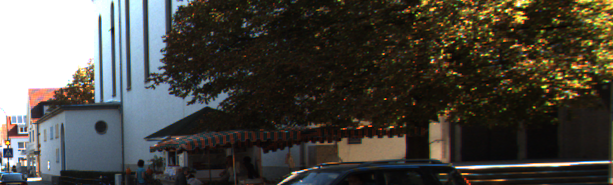}}
	\subfigure[The first quadrant of Frame 2]{
		\label{Fig.img-1}
		\includegraphics[width=0.226\textwidth,height=0.14\columnwidth]{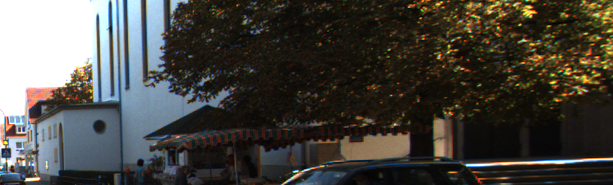}}
	\subfigure[Sub optical flow]{
		\label{Fig.img_flo}
		\includegraphics[width=0.226\textwidth,height=0.14\columnwidth]{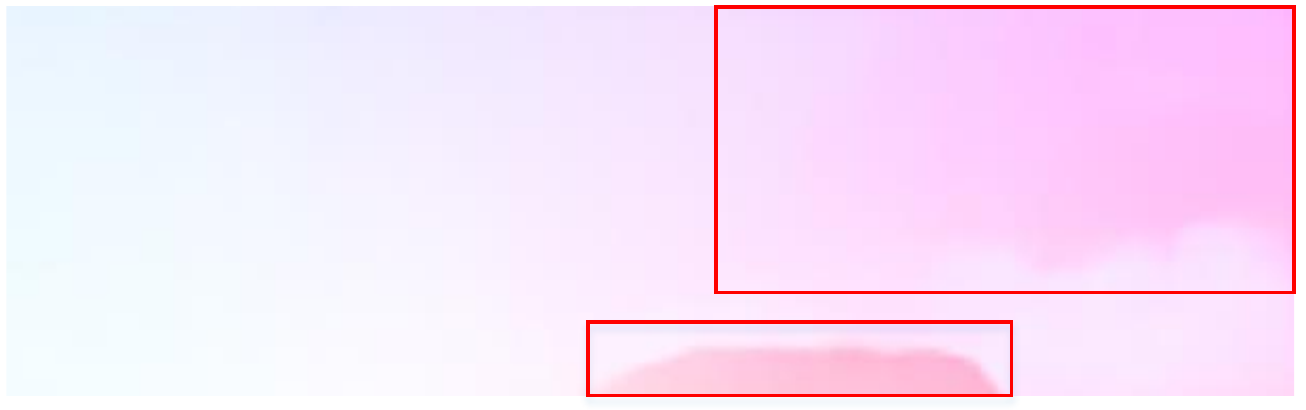}}
	\subfigure[The differential matrix]{
		\label{Fig.res_attention}
		\includegraphics[width=0.226\textwidth,height=0.14\columnwidth]{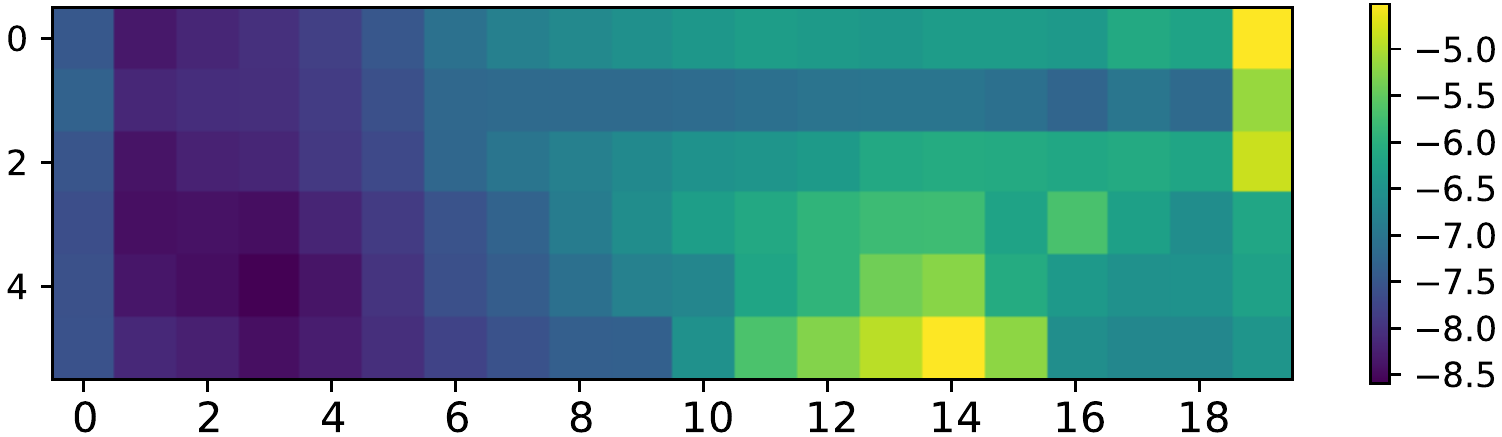}}
	\caption{Implementation of the CBAM \cite{woo2018cbam} in the first branch of the DeepAVO. (a) and (b) are the first quadrant of Fig. \ref{fig:f1} and Fig. \ref{fig:f2}. (c) is corresponding sub optical flow calculated from PWC-Net \cite{sun2018pwc}, and red boxes indicate the zone where pixel movement is intense. (d) is differential matrix between features after and before using CBAM in $FE_{1}$.}
	\label{Fig.attention}
\end{figure}
In this paper, the CBAM is implemented after the convolutional layers in FE1 and FE2. Fig. \ref{Fig.attention} presents how CBAM guides the VO. We calculate the difference between distilled feature map and the original feature map, called the differential matrix, which is visualized in Fig. \ref{Fig.res_attention}. Because activation function (i.e., $Sigmoid$) in Equation (12) and Equation (13) projects the attention maps into the range of 0 to 1, values of elements in the distilled feature map are smaller than original ones. Therefore, The zone where elements are closer to 0 (the brighter color in visualization) is given more attention. It can be observed that the CBAM focuses more on objects close to the camera (pixels with obvious motion), such as the stationary car at the crossroads and the trees on the roadside, corresponding to the red boxes in Fig. \ref{Fig.img_flo}. This demonstrates the CBAM has the ability to assist the $\emph{Encoder}$ in distilling the more effective representations from redundant features for pose estimation.

\subsection{Loss Function}
\label{subsec:loss_function}
KITTI dataset \cite{geiger2012we} was collected by a car whose motion model can be simplified as the motion on a 2-dimensional plane\cite{muller2017flowdometry}. The Y-axis for elevation is left out because the elevation differences are at least an order of magnitude smaller than the movement in the other axes. The dataset provides ground truth odometry information as a series of $3\times4$ transformation matrices that transform the first frame of a video sequence into the coordinate system of the current frame. The transformation matrix is formed by concatenating the rotation matrix (i.e., $\bm{R}_{t}$) and the translation vector (i.e., $\bm{T}_{t}$), which are defined in Equation (\ref{14}) and Equation (\ref{15}), respectively.
\begin{equation}\label{14}
  \bm{R}_{t} = \begin{bmatrix} R_{t,1}, \quad R_{t,2}, \quad R_{t,3} \\ R_{t,4}, \quad R_{t,5}, \quad R_{t,6} \\ R_{t,7}, \quad R_{t,8}, \quad R_{t,9} \end{bmatrix}
\end{equation}
\begin{equation}\label{15}
  \bm{T}_{t} = \begin{bmatrix} T_{t,X} \\ T_{t,Y} \\ T_{t,Z} \end{bmatrix}
\end{equation}

From this set of data, by decomposing the rotation matrix to find the difference between angles, the incremental angle change (i.e., $\Delta\varphi_{t}$) can be calculated, as shown in Equation (\ref{16}). The incremental distance change (i.e., $\Delta p_{t}$) is gained by calculating the Euclidean distance between the translational parts of the transformation matrices, as shown in Equation (\ref{17}).
\begin{equation}\label{16}
  \Delta \varphi_{t} = \arctan(-R_{t,3}, R_{t,1}) - \arctan(-R_{t-1,3}, R_{t-1,1})
\end{equation}

\begin{equation}\label{17}
  \Delta p_{t} = \sqrt{\sum(\bm{T}_{t} - \bm{T}_{t-1})^{2}}
\end{equation} For each optical flow input, the model regresses an angle and a distance to represent the displacement and orientation changes of the camera. This converts global transformation data into an ego-motion format in which small changes are accumulated over time.

The proposed network architecture based on the DeepAVO system can be considered to compute the conditional probability of the F2F poses $\bm{Y}_{t}$, given the optical flow data $\bm{Flo}_{t}$ at time $t$. To find the optimal parameters $\bm{\theta^{*}}$ for the model, DeepAVO maximizes conditional probability:
\begin{equation}\label{18}
  \bm{\theta^{*}} = \mathop{\arg\max}_{\bm{\theta}}p(\bm{Y}_{t}|\bm{Flo}_{t}; \bm{\theta})
\end{equation}

To learn the parameters $\bm{\theta}$, the Euclidean distance between the ground truth pose $(p_{t}, \varphi_{t})$ at time $t$ and its estimated one $(\widehat{p}_{t}, \widehat{\varphi}_{t})$ is minimized. The loss function is composed of Mean Square Error (MSE) of the position and orientation:
\begin{equation}\label{19}
\bm{\theta^{*}} = \mathop{\arg\max}_{\bm{\theta}}\frac{1}{N}\sum\limits_{t=1}^{N}\parallel\widehat{p}_{t}-p_{t}\parallel_{2}^{2} + \alpha\parallel\widehat{\varphi}_{t}-\varphi_{t}\parallel_{2}^{2}
\end{equation}where $ \left\|\ \right\|_{2} $ is 2-norm, and $N$ is the number of samples. ${\alpha}$ is a scale factor to balance the weights of translations and rotations. The better performance can be achieved by our model when setting ${\alpha}=100$. Detailed reasons and analysis are presented in Section \ref{subsubsec:balance}.

The displacements and angles computed for the optical flow are independent of the previous or next frame in the video sequence. However, The evaluation of the model needs to convert the pose predicted by the DeepAVO into the KITTI odometry benchmark format. The process can be described as:
\begin{equation}\label{20}
\begin{bmatrix} \bm{R}\mid \bm{T}\end{bmatrix}_{t} =
\left[
\begin{array}{cccc}
 \cos(\varphi_{t})& 0 &-\sin(\varphi_{t})  & T_{t,X}\\
 0& 1 &0& 0\\
 \sin(\varphi_{t})& 0 &\cos(\varphi_{t})  &T_{t,Z}
\end{array}
\right ]
\end{equation}where $\varphi_{t}$, and $T_{t,X}$, $T_{t,Z}$ are accumulated angle and distance, We update them as follows:
\begin{equation}\label{21}
\begin{cases}{\varphi}_{t} = {\varphi}_{t-1} + \Delta{\varphi}_{t-1}\\T_{t,X} = T_{t-1,X} + \Delta p_{t}\cos({\varphi}_{t})\\T_{t,Z} = T_{t-1,Z} + \Delta p_{t}\sin({\varphi}_{t})\end{cases}
\end{equation}At the start of every sequence, the camera position is initialized at the origin of an XZ coordinate system, with X and Z as the 2D movement plane. Starting from the origin, the next position is accumulated by applying the angle and displacement to the current position, thereby obtaining the absolute pose to origin to plot the driving path and evaluate the model performance.
\begin{figure*}[!ht]
	\centering
	\subfigure[Sequence 03]{
		\label{Fig.3.1}
		\includegraphics[width=0.2344\textwidth,height=0.375\columnwidth]{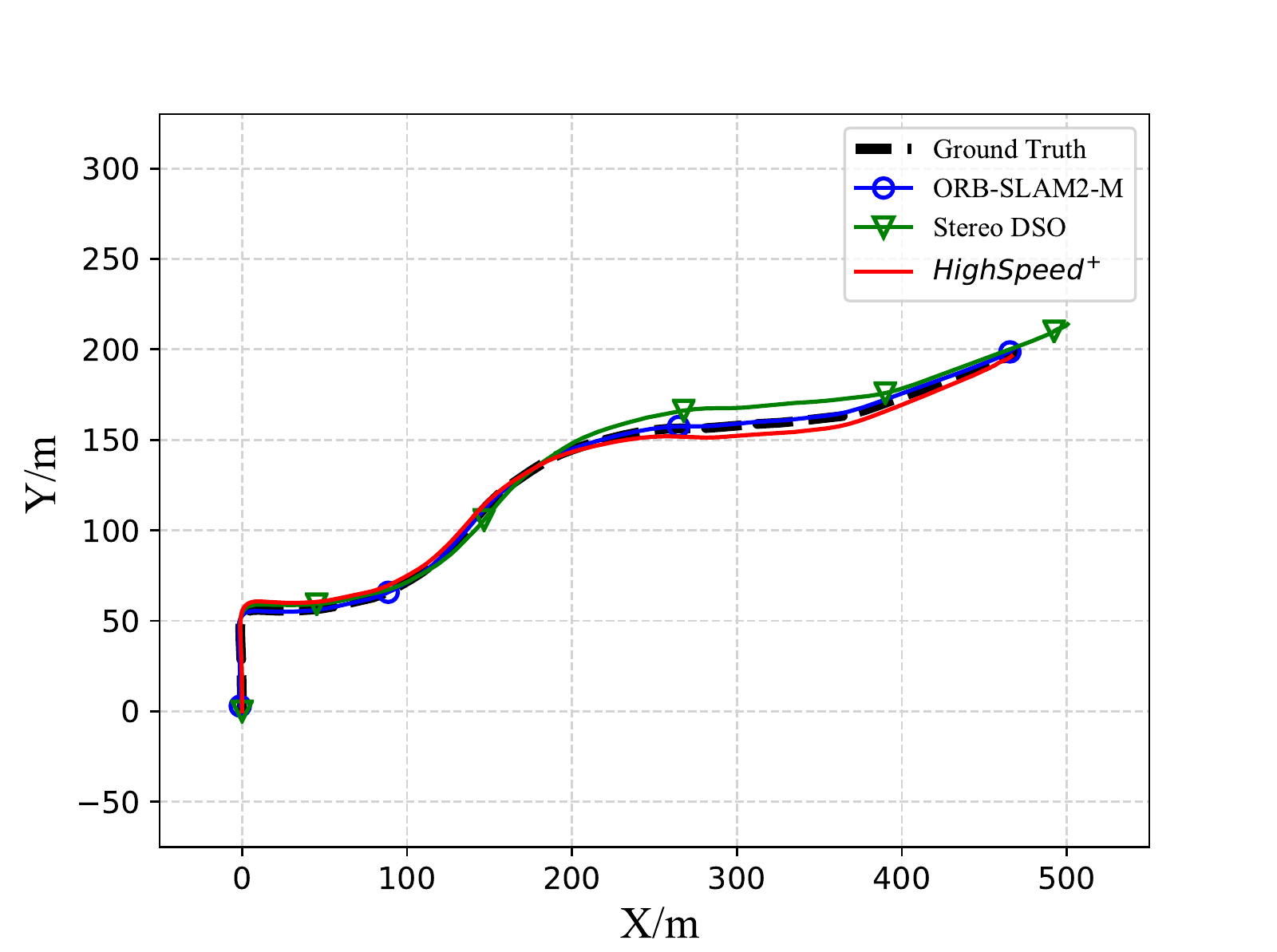}}
	\subfigure[Sequence 05]{
		\label{Fig.5.1}
		\includegraphics[width=0.2344\textwidth,height=0.375\columnwidth]{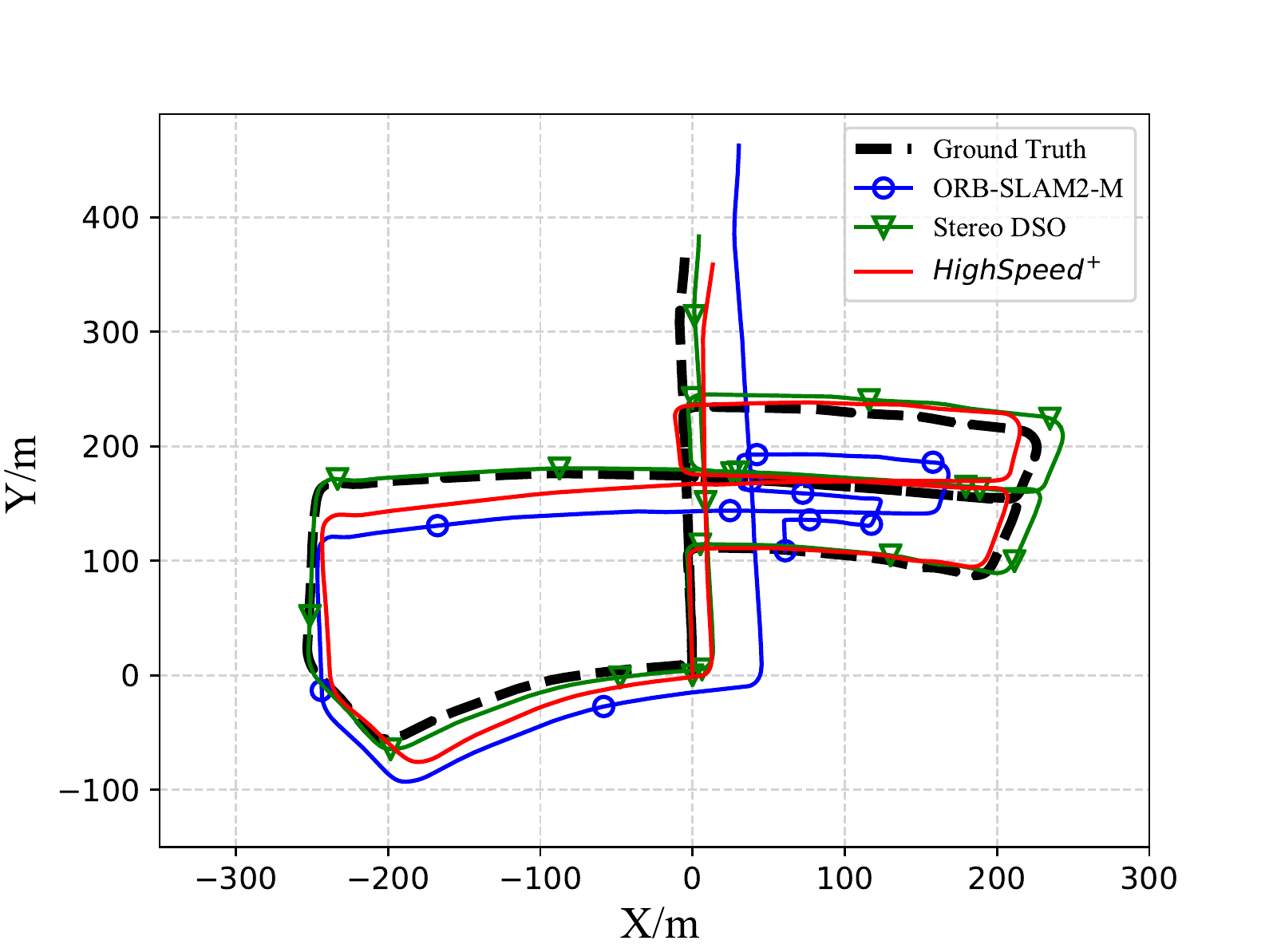}}
	\subfigure[Sequence 07]{
		\label{Fig.7.1}
		\includegraphics[width=0.2344\textwidth,height=0.375\columnwidth]{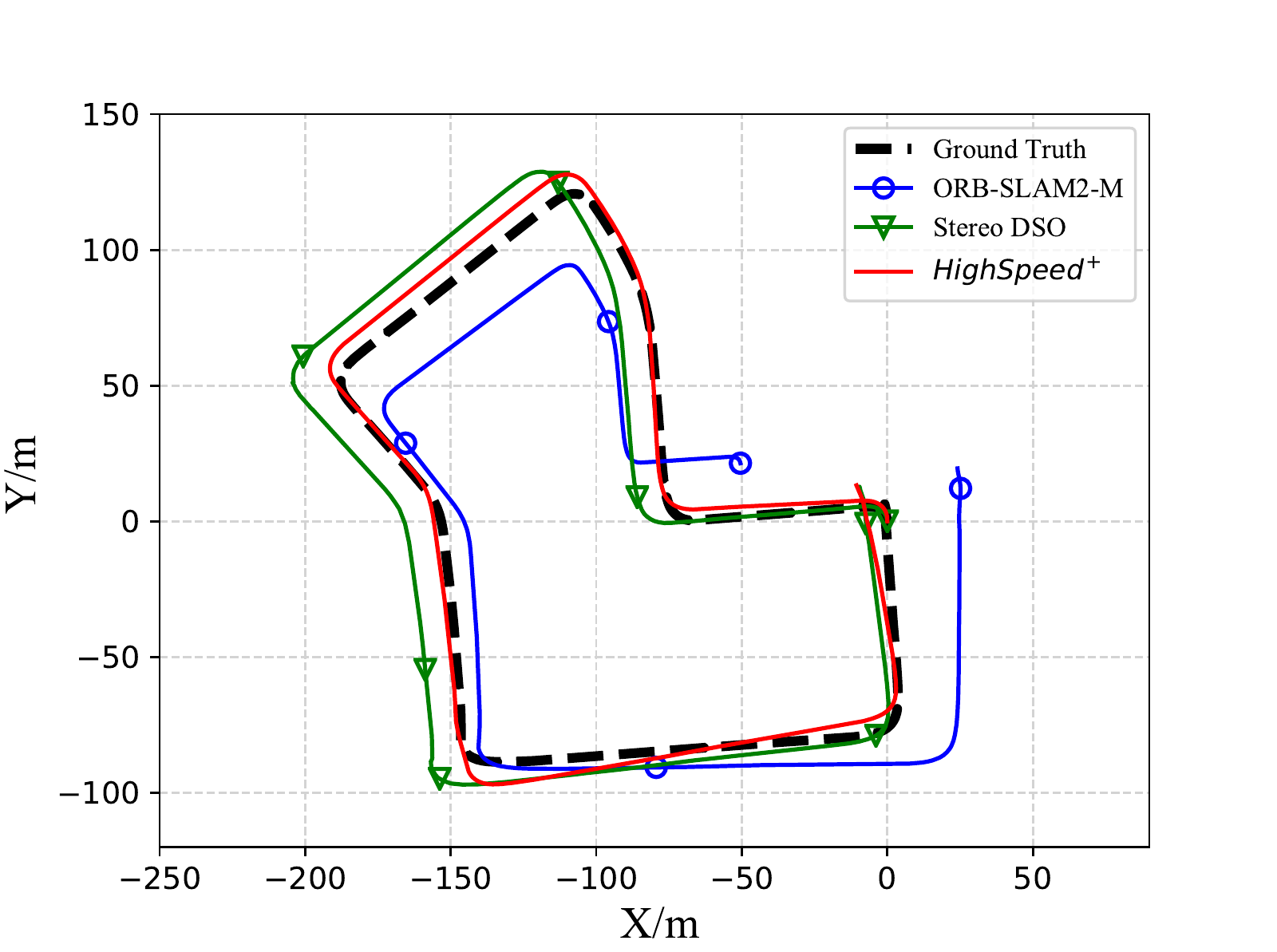}}
	\subfigure[Sequence 10]{
		\label{Fig.10.1}
        \includegraphics[width=0.2344\textwidth,height=0.375\columnwidth]{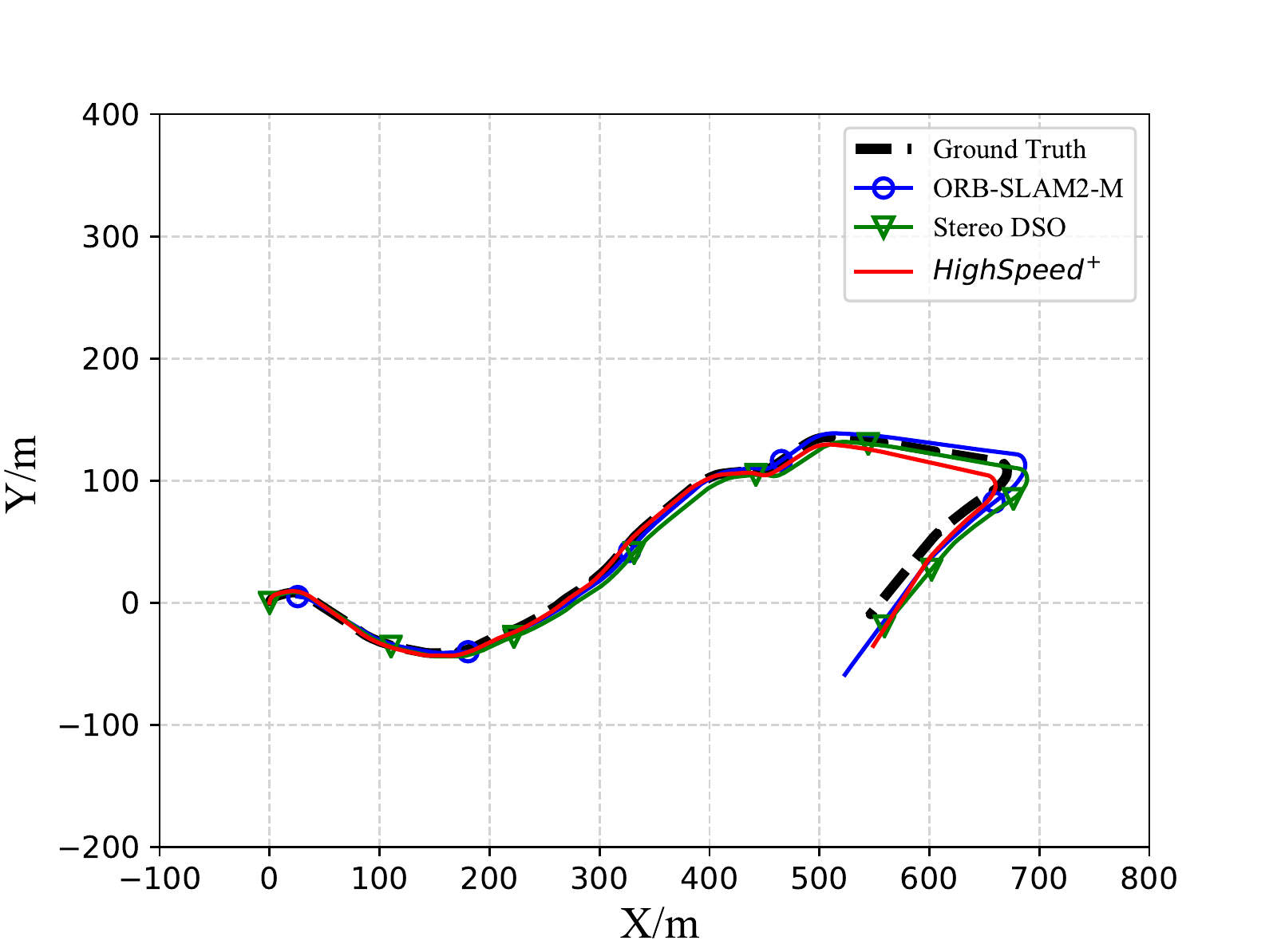}}
    \subfigure[Sequence 03]{
		\label{Fig.3.11}
		\includegraphics[width=0.2344\textwidth,height=0.375\columnwidth]{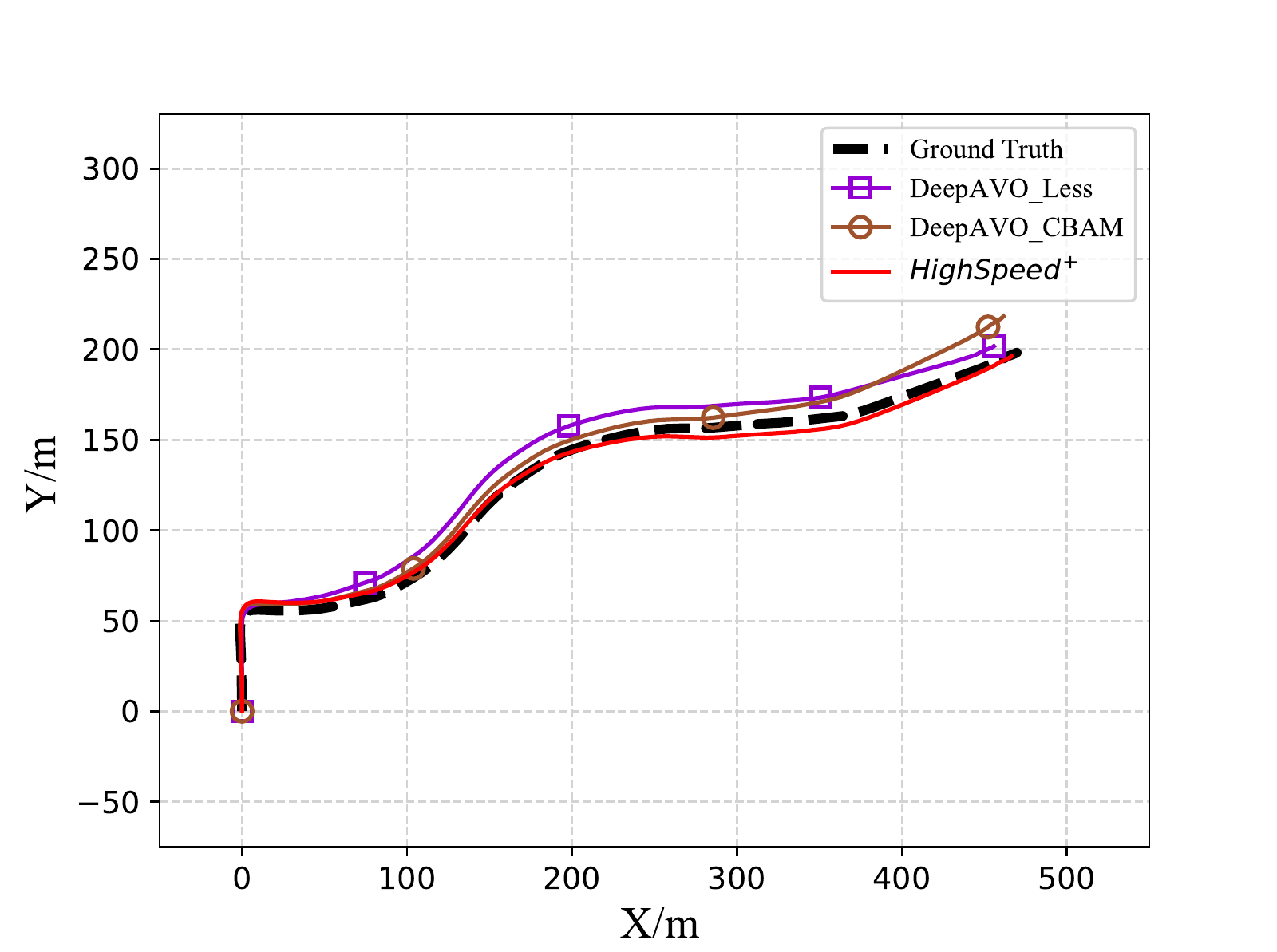}}
	\subfigure[Sequence 05]{
		\label{Fig.5.11}
		\includegraphics[width=0.2344\textwidth,height=0.375\columnwidth]{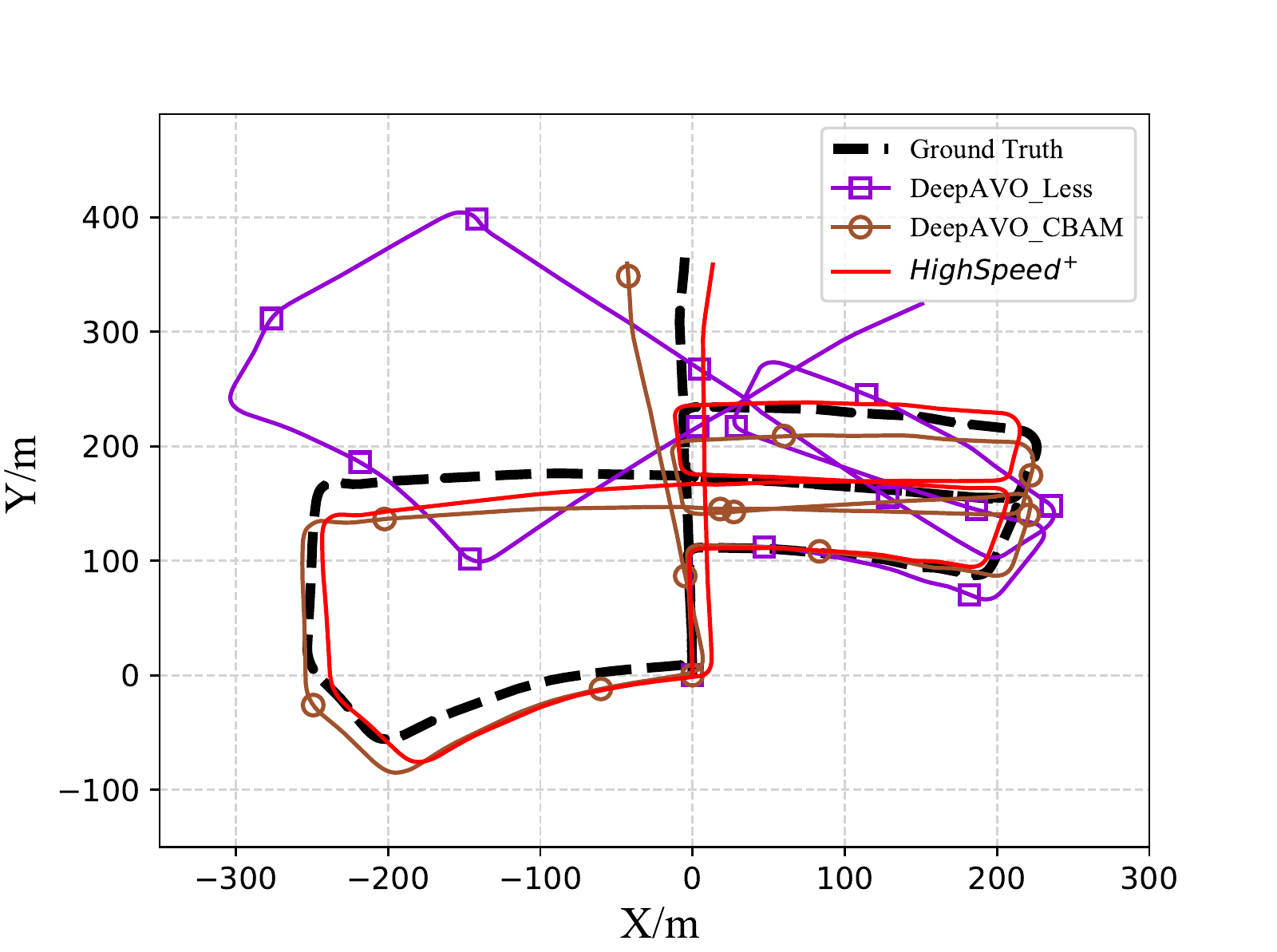}}
	\subfigure[Sequence 07]{
		\label{Fig.7.11}
		\includegraphics[width=0.2344\textwidth,height=0.375\columnwidth]{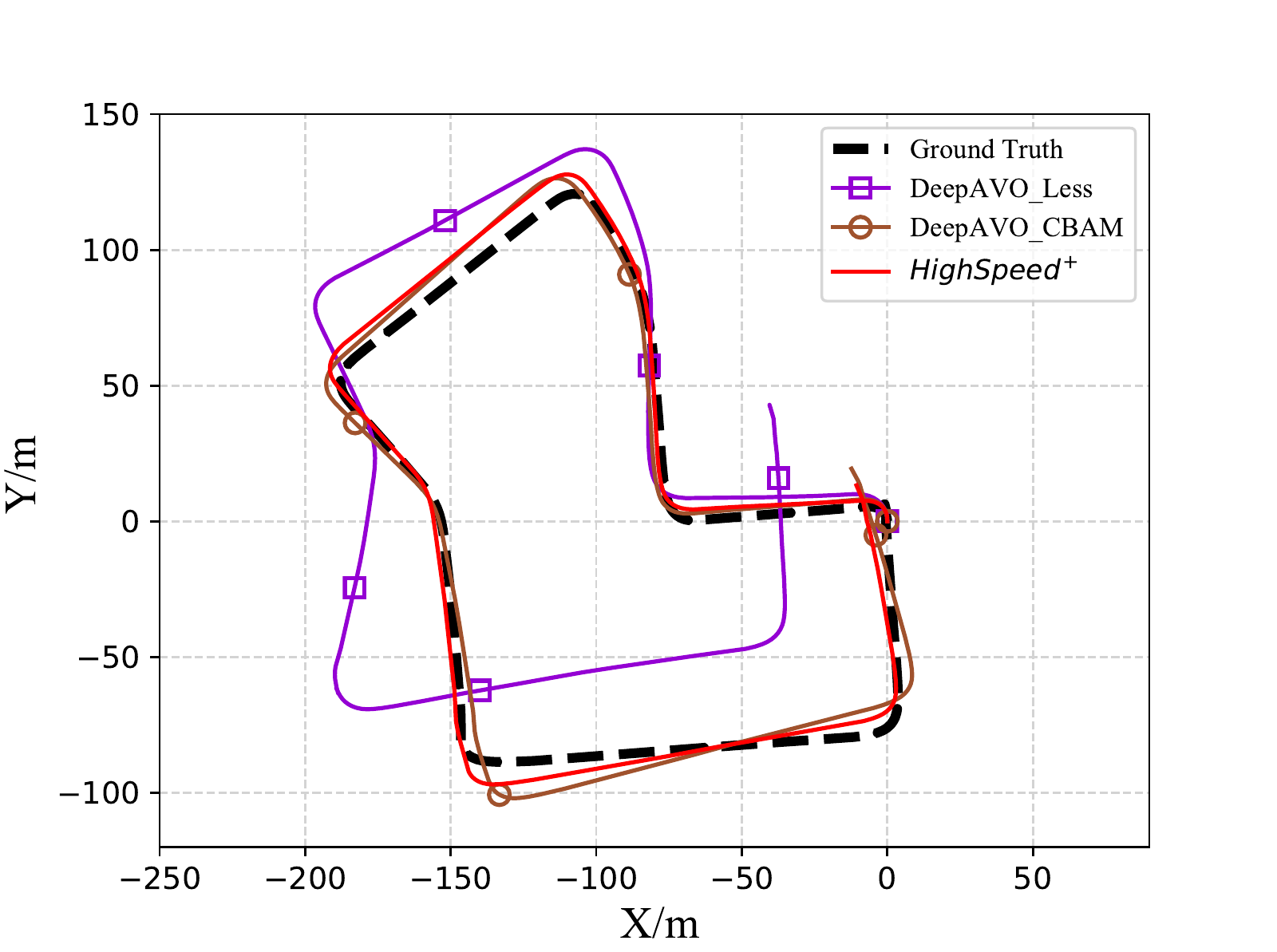}}
	\subfigure[Sequence 10]{
		\label{Fig.10.11}
        \includegraphics[width=0.2344\textwidth,height=0.375\columnwidth]{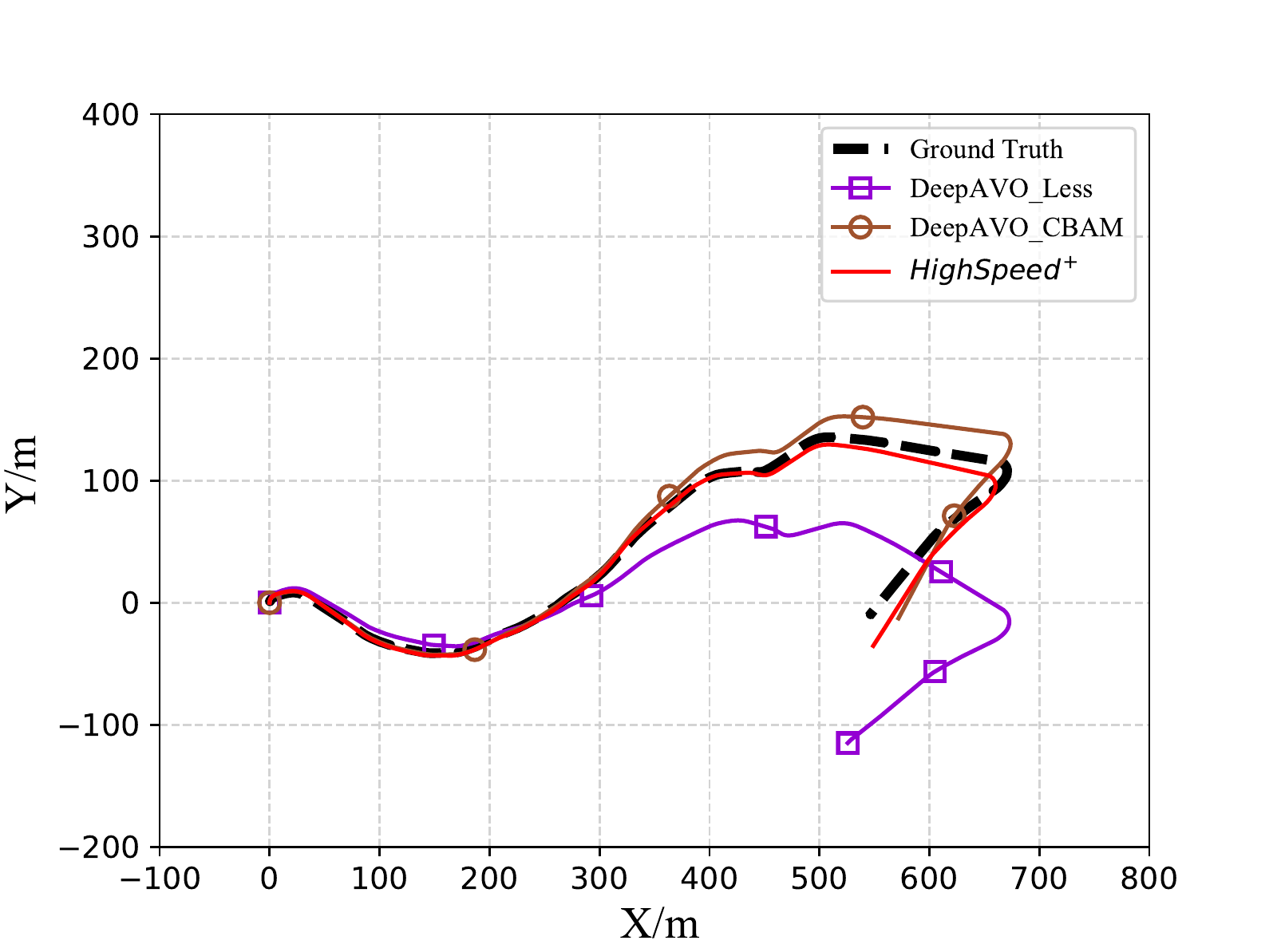}}
	  \caption{The trajectories of ground truth, ORB- SLAM2-M, Stereo DSO, and our model DeepAVO on Sequence 03, 05, 07, and 10 of the KITTI benchmark. Especially, this figure highlights the vital role of the attention module through the performance of the model with and without attention mechanism.}
	\label{Fig.test}
\end{figure*}

\section{Experiments}
\label{sec:experiments}
In this section, we first discuss the implementation details of our framework. Next, we evaluate the proposed DeepAVO by comparing it with various state-of-the-art algorithms in different scenarios, ranging from outdoor driving car (KITTI benchmark \cite{geiger2012we}, Malaga dataset \cite{blanco2014malaga}, ApolloScape dataset \cite{wang2018dels}) to self-collected indoor dataset. Finally, since the real-time operation is critical for robotic applications and learning-based methods are generally considered to be computationally expensive, we also discuss the real-time performance of the DeepAVO.

\subsection{Implementation}
\label{subsec:implementation}

\subsubsection{Dataset}
The KITTI dataset contains 22 video sequences captured in urban and highway environments at a relatively low sample frequency (10 fps) at the driving speed up to 90 km/h. It is very challenging for the VO monocular task. Sequence 00-10 associate with the ground truth measured and calibrated by multiple combined sensors, while the other 10 sequences (Sequence 11-21) are only provided with raw images. The size of raw images between different sequences does not remain the same. For example, the images of the Sequence 00-02 is 1241$\times$376 pixels, while the Sequence 04-11 is 1226$\times$370. In our experiments, the size of left RGB images is unified into 1226$\times$370 for training and testing.

\subsubsection{Training and Testing}
Two sets of experiments are conducted separately to evaluate the proposed method on the KITTI dataset. The first one is based on Sequence 00-10 to quantitatively and qualitatively analyze the model performance using ground truth since ground truth is only provided for these sequences. We adopt the same train/test split as  DeepVO \cite{wang2017deepvo} and ESP-VO \cite{wang2018end} by using Sequence 00, 01, 02, 08, 09 for training, which are relatively long. The trajectories are converted into optical flow data by PWC-Net \cite{sun2018pwc} for training. Then, the trained model is tested on Sequence 03, 04, 05, 06, 07, and 10 for evaluation.

Another experiment aims to evaluate the generalization of the DeepAVO: the ability of a learning-based method to maintain the performance in totally new environments. Therefore, models trained on all Sequence 00-10 are tested on Sequence 11-21, where there is no ground truth to train. In order to further analyze the generalization of the DeepAVO in the different datasets for a cross-dataset validation, the Malaga dataset \cite{blanco2014malaga}, Apollo dataset \cite{wang2018dels} and self-collected indoor dataset are used to test the model trained on Sequence 00-10 of the KITTI dataset.

\subsubsection{Network}
The network is implemented by the Tensorflow-1.9.0 framework \cite{abadi2016tensorflow} on an NVIDIA Geforce Titan XP GPU. Adam\cite{kingma2014adam} with $\beta_{1}=0.9, \beta_{2}=0.99$ is used as the optimizer to train the network for up to 70 epochs with a batch size of 48. Besides, Batch Normalization and Xavier weight initialization are used to make the network converge faster and better. The initial learning rate is set to 1$\times$$10^{-4}$ and reduce by half every 15 epochs. Dropout and early stopping technologies are introduced to prevent the model from overfitting.

\begin{table*}[!t]
\renewcommand{\arraystretch}{1.25} 
\setlength{\tabcolsep}{0.8mm}
\caption{Results on the KITTI dataset.}
\begin{tabular}{lcccccccccccccc}
\hline\hline
\multirow{3}{*}{Method} & \multicolumn{14}{c}{Sequence}                                                                                                                                                 \\
                        & \multicolumn{2}{c}{03} & \multicolumn{2}{c}{04} & \multicolumn{2}{c}{05} & \multicolumn{2}{c}{06} & \multicolumn{2}{c}{07} & \multicolumn{2}{c}{10} & \multicolumn{2}{c}{Avg} \\ \cline{2-15}
                        & $t_{rel}$  & $r_{rel}$ & $t_{rel}$  & $r_{rel}$ & $t_{rel}$  & $r_{rel}$ & $t_{rel}$  & $r_{rel}$ & $t_{rel}$  & $r_{rel}$ & $t_{rel}$  & $r_{rel}$ & $t_{rel}$  & $r_{rel}$  \\ \hline
\textbf{Traditional methods}\\ \hline
Stereo DSO\cite{wang2017stereo}       & 6.45         & 0.16        & 3.36         & 0.13        & 3.03         & 0.19        & 3.57         & 0.31        & 4.25         & 0.54        & 2.04         & 0.20         & 3.28        & 0.24  \\
ORB-SLAM2-M\cite{mur2017orb}          & 1.37         & 0.22        & 1.23         & 0.19        & 17.46        & 0.63        & 21.02        & 0.26        & 12.74        & 1.43        & 4.44         & 0.44         & 9.71        & 0.53  \\ \hline
\textbf{Learning-based VOs}\\ \hline
ESP-VO\cite{wang2018end}              & 6.72         & 6.46        & 6.33         & 6.08        & 3.35         & 4.93        & 7.24         & 7.29        & 3.52         & 5.02        & 9.77         & 10.20        & 6.12        & 6.15  \\
CL-VO\cite{2019Learning}              & 8.12         & 3.47        & 7.57         & 2.61        & 5.77         & 2.00        & 7.66         & 1.66        & 6.79         & 3.00        & 8.29         & 2.94         & 7.37        & 2.67       \\
NeuralBundler\cite{2019Pose}          & 4.51         & 2.82        & $\bm{2.3}$   & 0.87        & 3.91         & 1.64        & 4.6          & 2.85        & 3.56         & 2.39        & 12.9         & 3.17         & 5.30        & 2.29       \\
DAVO\cite{2020Dynamic}                & 5.50         & 2.71        & 6.03         & 2.37        & $\bm{2.28}$  &$\bm{1.14}$ &$\bm{4.19}$    & 1.69        & 4.11         & 2.61        &$\bm{4.26}$   &$\bm{1.70}$   & 4.40        & 2.04       \\ \hline
\textbf{Proposed methods}\\ \hline
DeepAVO\_Less                         & 6.56         & 2.59        & 3.95         & 1.40        & 7.41         & 3.36        & 13.72        & 5.32        & 8.47         & 4.80        & 12.32        & 3.99         & 9.16        & 3.83  \\
DeepAVO\_Nloc                         & 10.55        & 2.58        & 4.98         & 1.18        & 5.01         & 1.84        & 15.00        & 6.02        & 11.25        & 3.52        & 9.14         & 3.15         & 8.14        & 2.91  \\
DeepAVO\_SE                           & 7.75         & 2.14        & 4.52         & 1.44        & 3.85         & 1.66        & 8.15         & 2.58        & 6.24         & 4.95        & 6.58         & 2.50         & 5.39        & 2.26  \\ 
DeepAVO\_CBAM                         & $\bm{3.38}$  & 1.96        & 5.70         & 0.98        & 3.31         & 1.36        & 7.43         & 2.55        & $\bm{3.31}$  & 2.57        & 6.15         & 2.67         & 4.43        & 1.88 \\
$HighSpeed^{+}$                       & 3.64         & $\bm{1.89}$ & 3.88         & $\bm{0.60}$ & 2.57         & 1.16        & 4.96         & $\bm{1.34}$ & 3.36         &$\bm{2.15}$  & 5.49         & 2.49        & $\bm{3.52}$ & $\bm{1.50}$ \\ \hline\hline
\end{tabular}
    \begin{itemize}
      \item $t_{rel}$: average translational RMSE drift ($\%$) on length from 100, 200 to 800 m.
      \item $r_{rel}$: average rotational RMSE drift ($^\circ$/100m) on length from 100, 200 to 800 m.
      \item $HighSpeed^{+}$ means the DeepAVO\_CBAM model trained on the original training set and  the subsampling data of Sequence 00.The best results are highlighted without considering traditional methods.
    \end{itemize}
\label{tab:t2}
\end{table*}
\begin{figure*}[!ht]
	\centering
	\subfigure[Translation against path length]{
		\label{Fig.TP}
		\includegraphics[width=0.4688\textwidth,height=0.73\columnwidth]{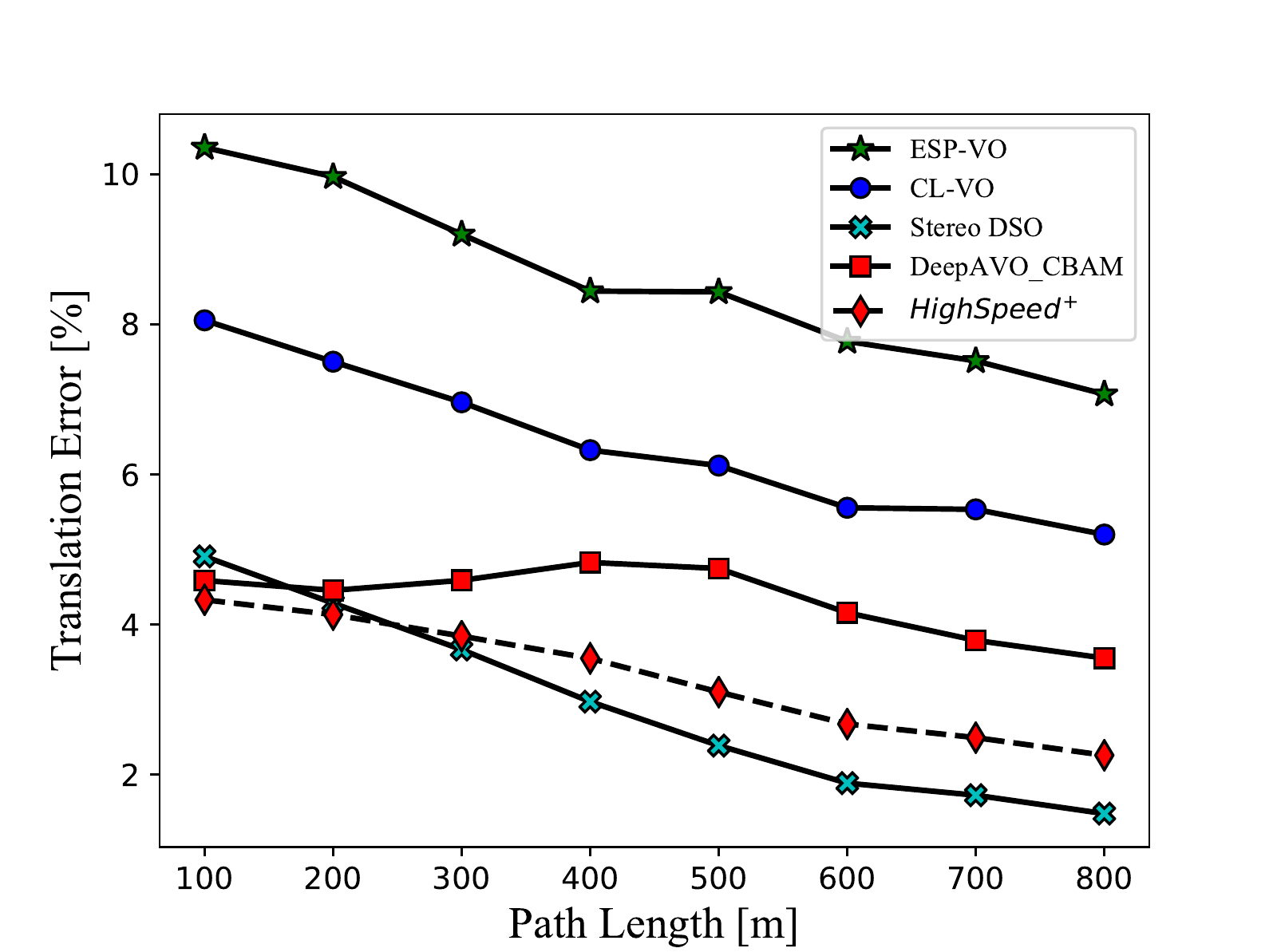}}
	\subfigure[Rotation against path length]{
		\label{Fig.RP}
		\includegraphics[width=0.4688\textwidth,height=0.73\columnwidth]{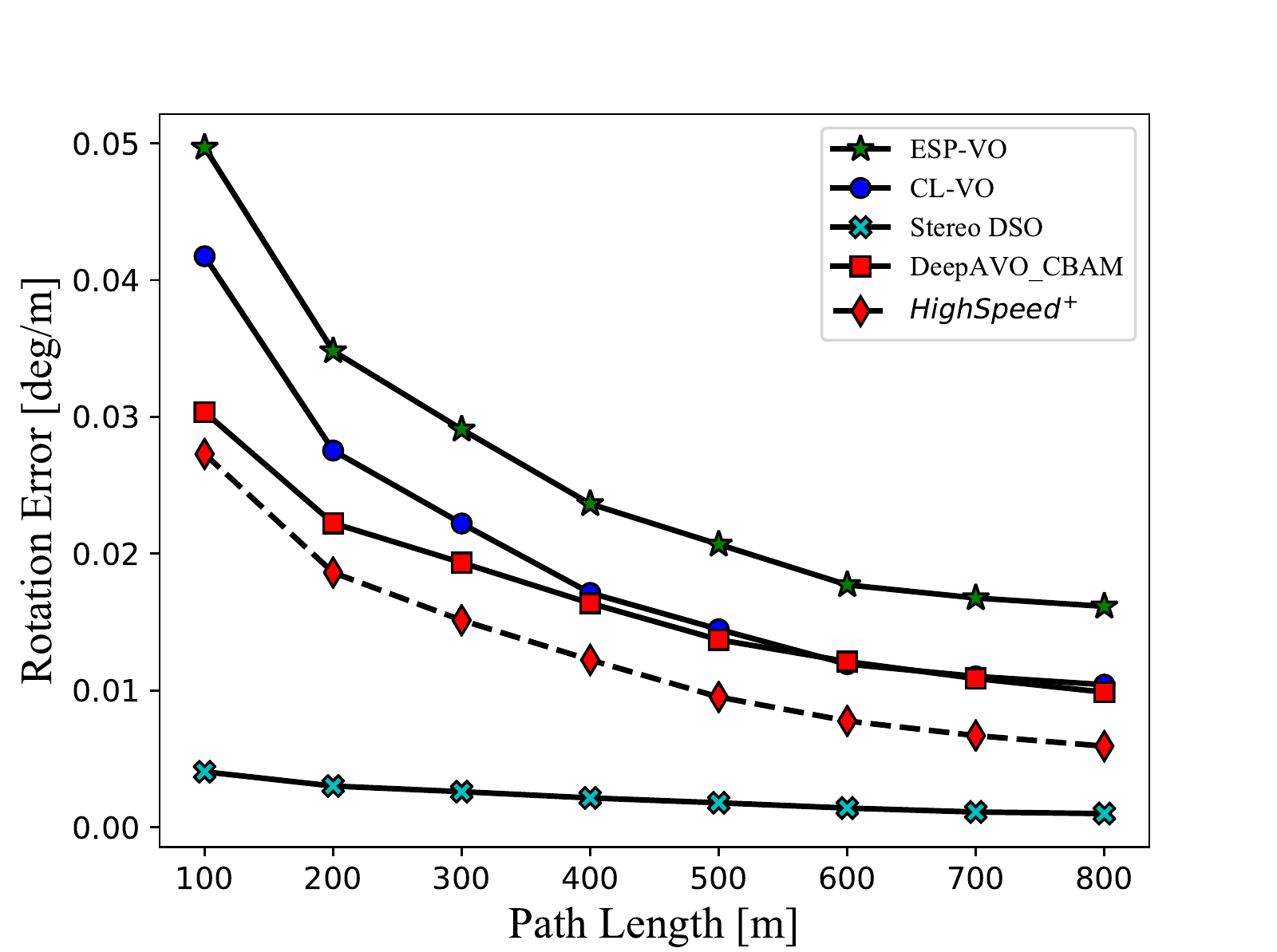}}
	\subfigure[Translation against speed]{
		\label{Fig.TS}
		\includegraphics[width=0.4688\textwidth,height=0.73\columnwidth]{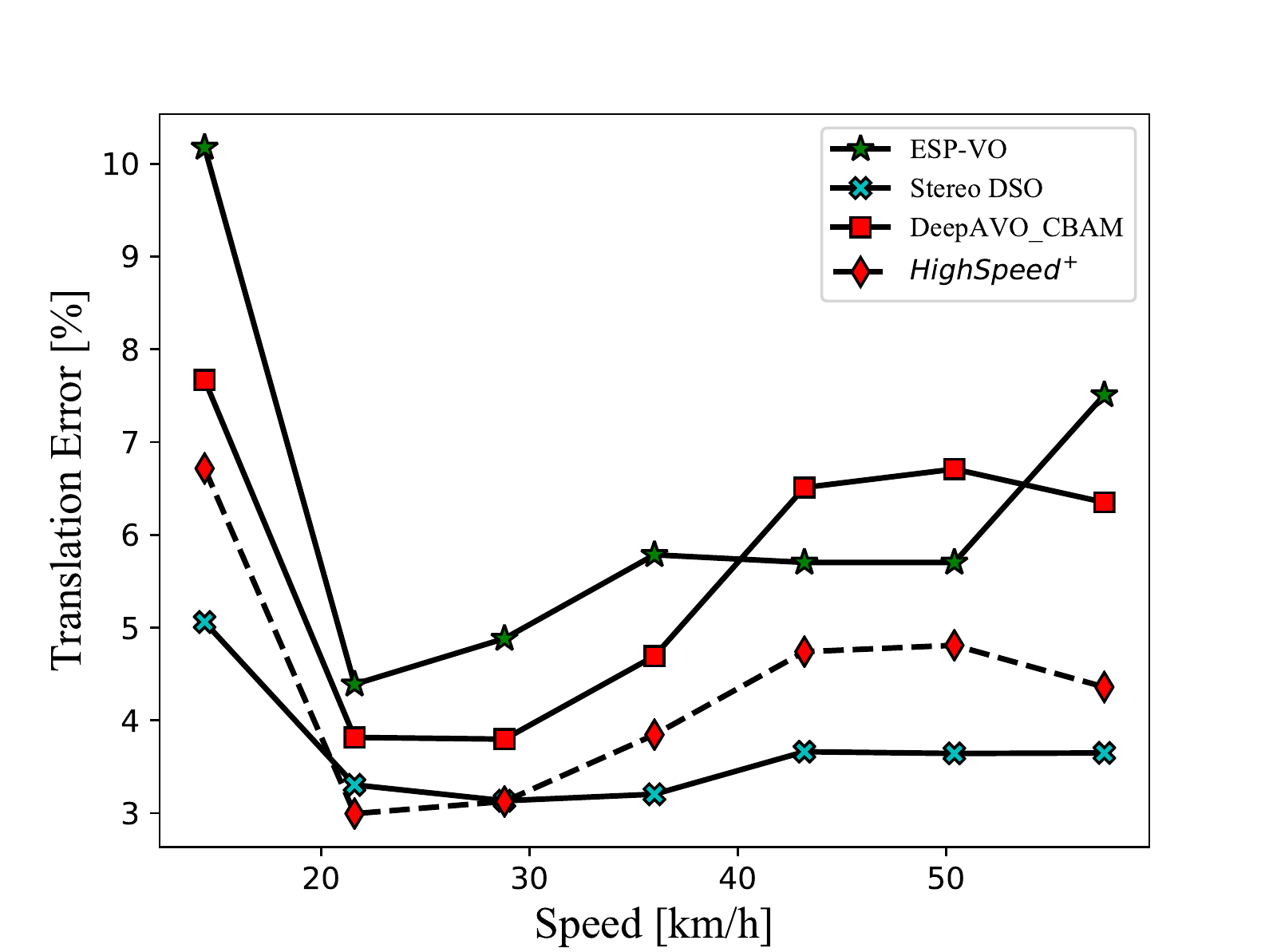}}
	\subfigure[Rotation against speed]{
		\label{Fig.RS}
		\includegraphics[width=0.4688\textwidth,height=0.73\columnwidth]{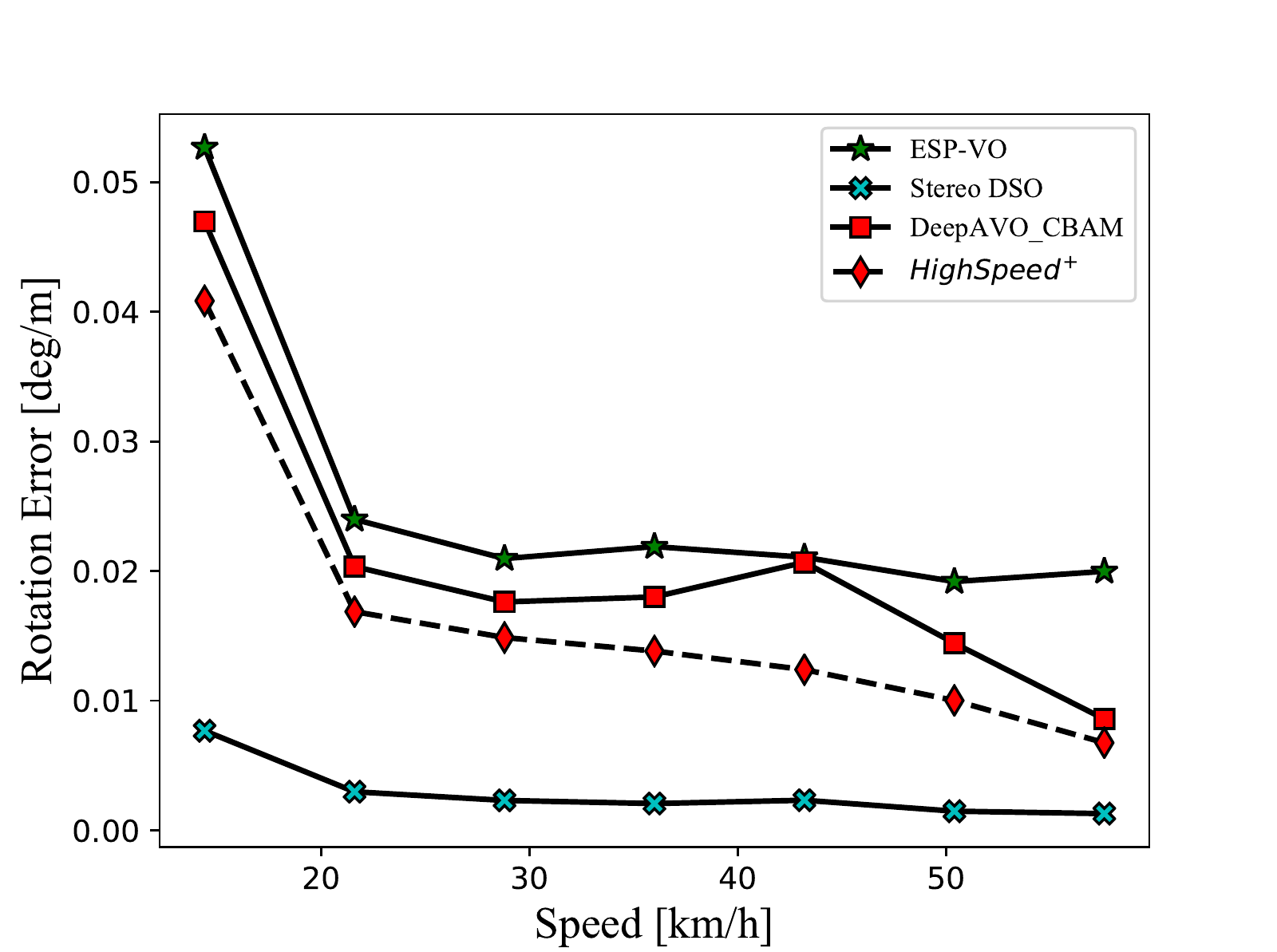}}
	\caption{Average errors across sequence lengths (a and b) and speeds(c and d) on test sequences of the proposed models and competitive approaches.}
	\label{Fig.error}
\end{figure*}

\subsection{Results on KITTI dataset}
\label{subsec:kitti results}

We compare the DeepAVO with several state-of-the-art VO algorithms, including the traditional stereo method DSO \cite{wang2017stereo}, monocular ORB-SLAM2-M \cite{mur2017orb} and the learning-based monocular models such as ESP-VO \cite{wang2018end}, NeuralBundler \cite{2019Pose}, CL-VO \cite{2019Learning}, DAVO \cite{2020Dynamic}. Although the direct method DSO is also capable of conducting ego-motion estimation in a monocular way, it consistently loses tracking while being tested on the KITTI dataset. To highlight the efficiency of the attention mechanism, we also consider the DeepAVO\_Less (i.e., our model without attention) and DeepAVO\_SE and DeepAVO\_Nloc using different attentions as the competitive methods. We follow the error metrics where averaged Root Mean Square Errors (RMSE) of the translational and rotational errors are adopted for different lengths of sub-sequences, ranging from 100, 200 to 800 meters, and different speeds (the range of speeds varies in different sequences). The detailed performance of the algorithms on the testing sequences is summarized in Table. \ref{tab:t2}.

\subsubsection{Qualitative and quantitative analysis}
Traditional monocular VO methods cannot recover the absolute scale and require pose alignment with ground truth. To achieve a fair comparison, the ORB-SLAM2-M is modified with its global loop-closure detection being disabled. Since the ORB-SLAM2-M does not recover the absolute scale, its keyframe trajectories are aligned to ground truth by using similarity transformation. Note that for DeepAVO, the scale learned in end-to-end training is completely maintained by the model itself without considering any prior knowledge and pose alignment. This indicates that the learning-based VO has an appealing advantage over other monocular VO. Table. \ref{tab:t2} suggests that our model, even with the vanilla version (i.e., DeepAVO\_Less), outperforms ORB-SLAM2-M in terms of the translation estimation, and the attention usage widens this margin further.  We also supplement the high-speed situations in the training set by adding the subsampled data of Sequence 00 of which the velocity shows the highest dynamic range, so as to alleviate high drifts in such scenarios. The visualization of trajectories corresponding to the previous testing is illustrated in Fig. \ref{Fig.test}. $HighSpeed^{+}$ outperforms DeepAVO\_CBAM and achieves very close performance to the stereo DSO. For DeepAVO\_Less, although achieving promising performance in regular environments (Sequence 03), it still suffers from the large scale drift under complicated scenes (Sequence 05, 07, and 10).

Table. \ref{tab:t2} also compares the proposed DeepAVO series with the other four learning-based methods. The rotation error of DeepAVO\_Less is slightly higher than the compared VOs, and the translation estimation still does not come up to the accuracy of baseline methods. It reveals that the distinct analysis of pixel motion in different quadrants of optical flow can elevate the performance when the model estimates the rotation. We assume that extracting motion-sensitive features directly from the encoded features may limit the accuracy. Fortunately, this deficiency is compensated by our proposed architecture that combines the attention mechanism to distill features, which are conducive to motion estimation. It is observed that DeepAVO\_CBAM outperforms most of the baseline methods and delivers comparable performance to DAVO \cite{2020Dynamic}, which additionally employs a semantic segmentation module for weighting semantic categories as well as a dilated pose estimation module for aggregating them in its architecture. The averaged $t_{rel}$ of DeepAVO\_CBAM is slightly (0.68\%) higher than that of DAVO. However, DeepAVO\_CBAM delivers a lower (7.84\%) averaged $r_{rel}$ than DAVO. $HighSpeed^{+}$, as the best one among the proposed DeepAVO series, further improves the performance of the proposed model, especially for sequence 04, 06, and 10 containing many high-speed samples. Compare with DAVO, the averaged $t_{rel}$ and $r_{rel}$ of $HighSpeed^{+}$ are 20\% and 26.47\% lower than those of DAVO, respectively.

In order to find out the attention mechanism that is preferable in guiding the VO task, we also discuss the performance of models with different attention modules. Among these models, SE and CBAM, unlike Nloc, exploit the correlation and dependence between features to distill information that is of great value to ego-motion estimation. The experimental results in Table. \ref{tab:t2} demonstrate the effectiveness of these two mechanisms for the VO task. Furthermore, the additional spatial constrain by CBAM, which preserves the valuable spatial features and suppresses the useless ones, allows DeepAVO\_CBAM to give the best performance to the DeepAVO framework.

We further evaluate the average RMSE of the estimated translation and rotation against different path lengths and speeds in Fig. \ref{Fig.error}. As the length of the trajectory increases, the errors of both the translation and rotation of the DeepAVO\_CBAM decrease, far exceeding other monocular methods, as shown in Fig. \ref{Fig.TP} and Fig. \ref{Fig.RP}. In term of the comparisons 
between the monocular VOs, DeepAVO\_CBAM consistently outperforms the other two competitors (i.e., ESP-VO and CL-VO) regardless of the travelled length increasing. Nevertheless, the translation estimated by the DeepAVO\_CBAM is slightly defective at high speed (Fig. \ref{Fig.TS}). It is attributed to the limited high-speed training samples in Sequence 00, 02, 08, and 09, of which the maximum speeds are all below 60km/h. As shown in Fig. \ref{Fig.TS}, $HighSpeed^{+}$ effectively alleviates the serious translational drifts in high moving speed. Compared with DeepAVO\_CBAM, the averaged $t_{rel}$ and $r_{rel}$ of $HighSpeed^{+}$ are reduced by 20.54\% and 20.21\% respectively. Especially for Sequences 04, 06, and 10, which contain many high-speed samples, the performance of $HighSpeed^{+}$ is significantly improved. By contrast, the rotational error of the DeepAVO\_CBAM shows a downtrend with the increasing speed in Fig. \ref{Fig.RS}. We presume that this is because the KITTI dataset recorded during car driving tends to go straight at high speeds. Moving forward at high speed, as a state without an obvious change in rotation, can be easily learned to model.

\subsubsection{The influence of balance parameter $\alpha$ in the Loss function}
\label{subsubsec:balance}
For the KITTI benchmark, the rotation in the F2F pose is two orders of magnitude smaller than the displacement. In order to balance the estimation in translation and rotation better, we test the influence of balance parameter $\alpha$ in the loss function (\ref{subsec:loss_function}) on the results. Theoretically, the rotational error can be reduced by our model when given a larger balance parameter to raise the weight of the rotational portion in the loss function. We compare the results of the balance parameter set to 10, 50, 100, and 150.

\begin{figure}[!h]
	\centering
	\subfigure[Sequence 03]{
		\includegraphics[width=0.226\textwidth,height=0.35\columnwidth]{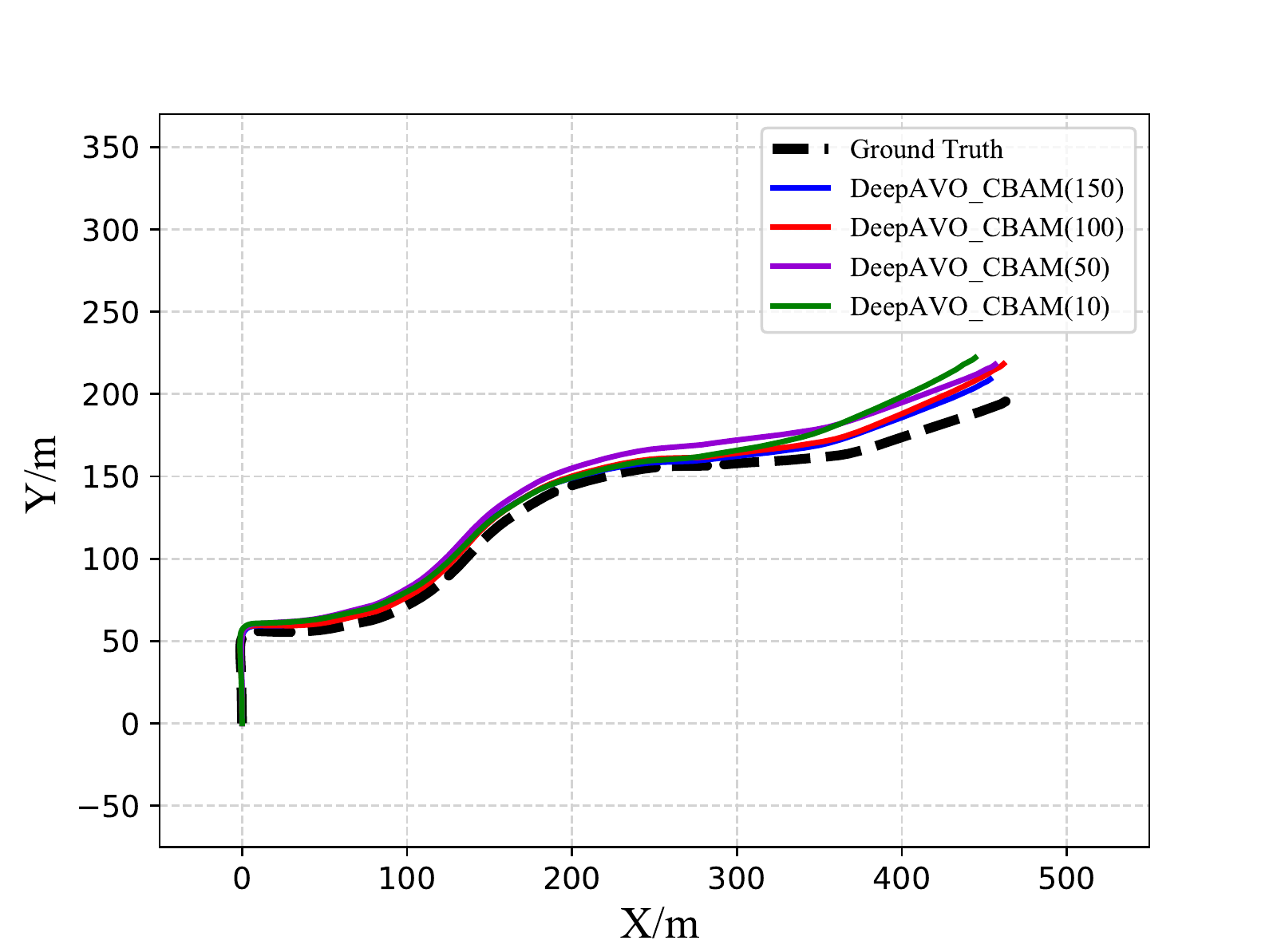}}
	\subfigure[Sequence 05]{
		\includegraphics[width=0.226\textwidth,height=0.35\columnwidth]{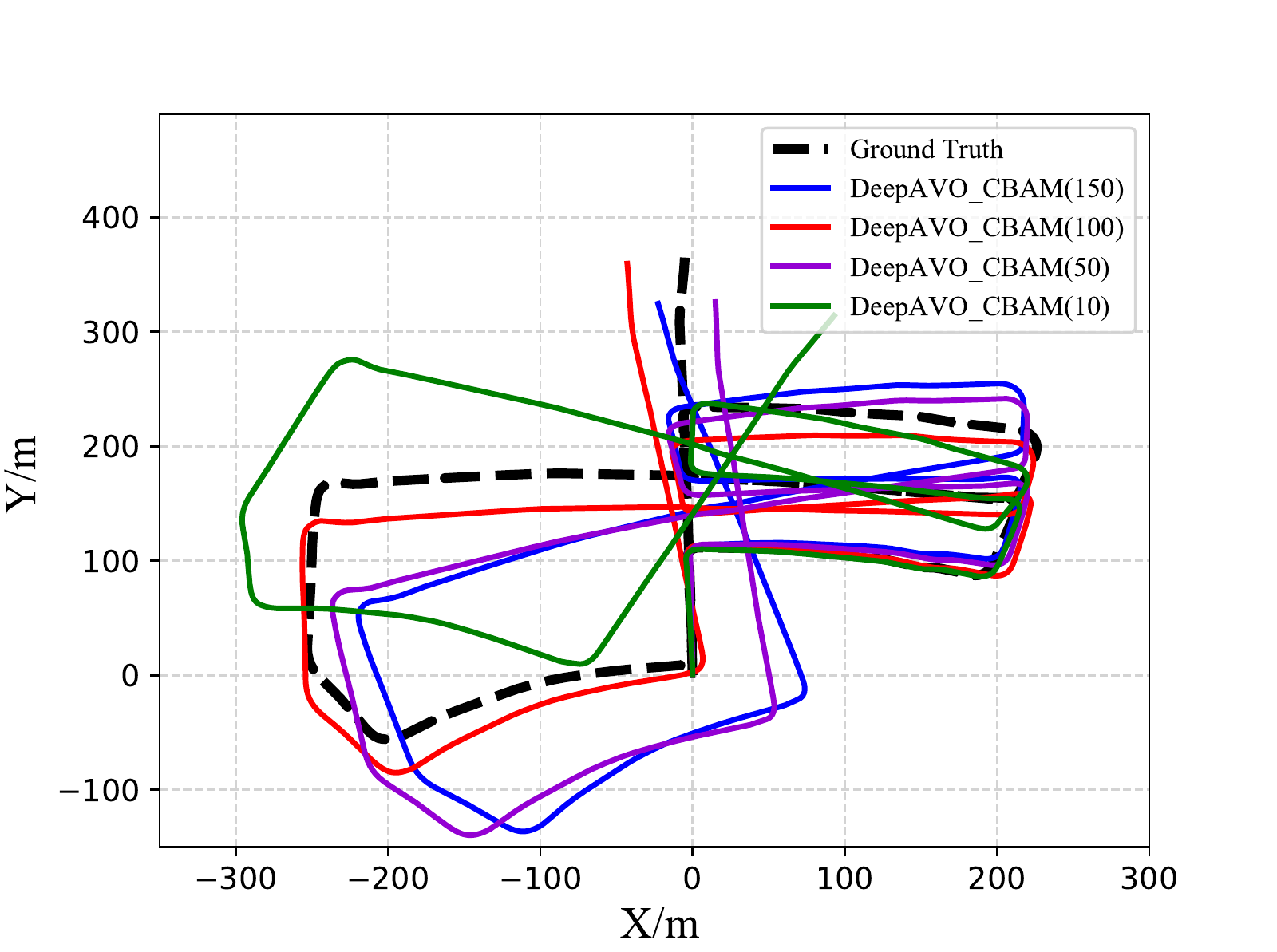}}
	\subfigure[Sequence 07]{
		\includegraphics[width=0.226\textwidth,height=0.35\columnwidth]{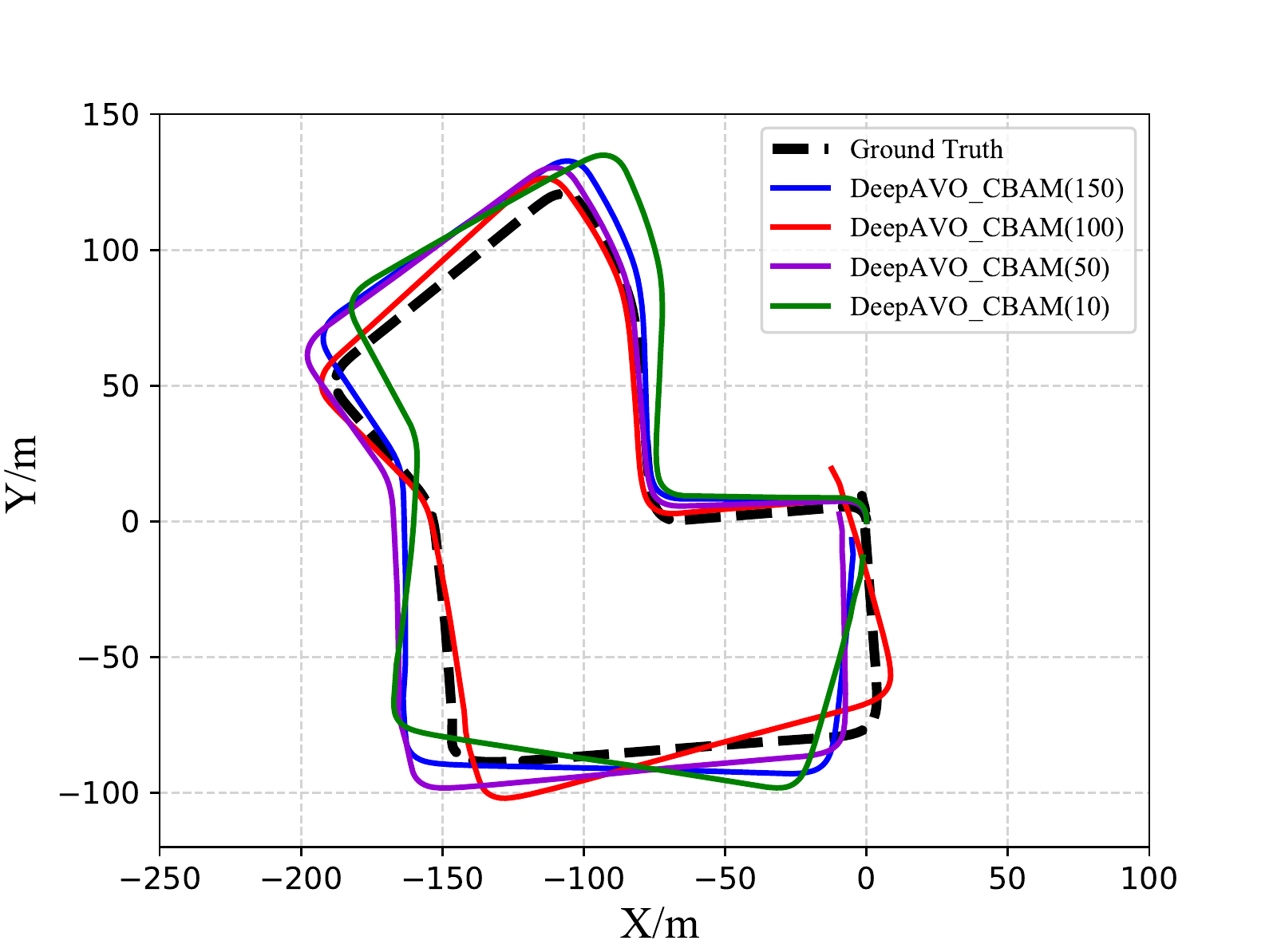}}
	\subfigure[Sequence 10]{
		\includegraphics[width=0.226\textwidth,height=0.35\columnwidth]{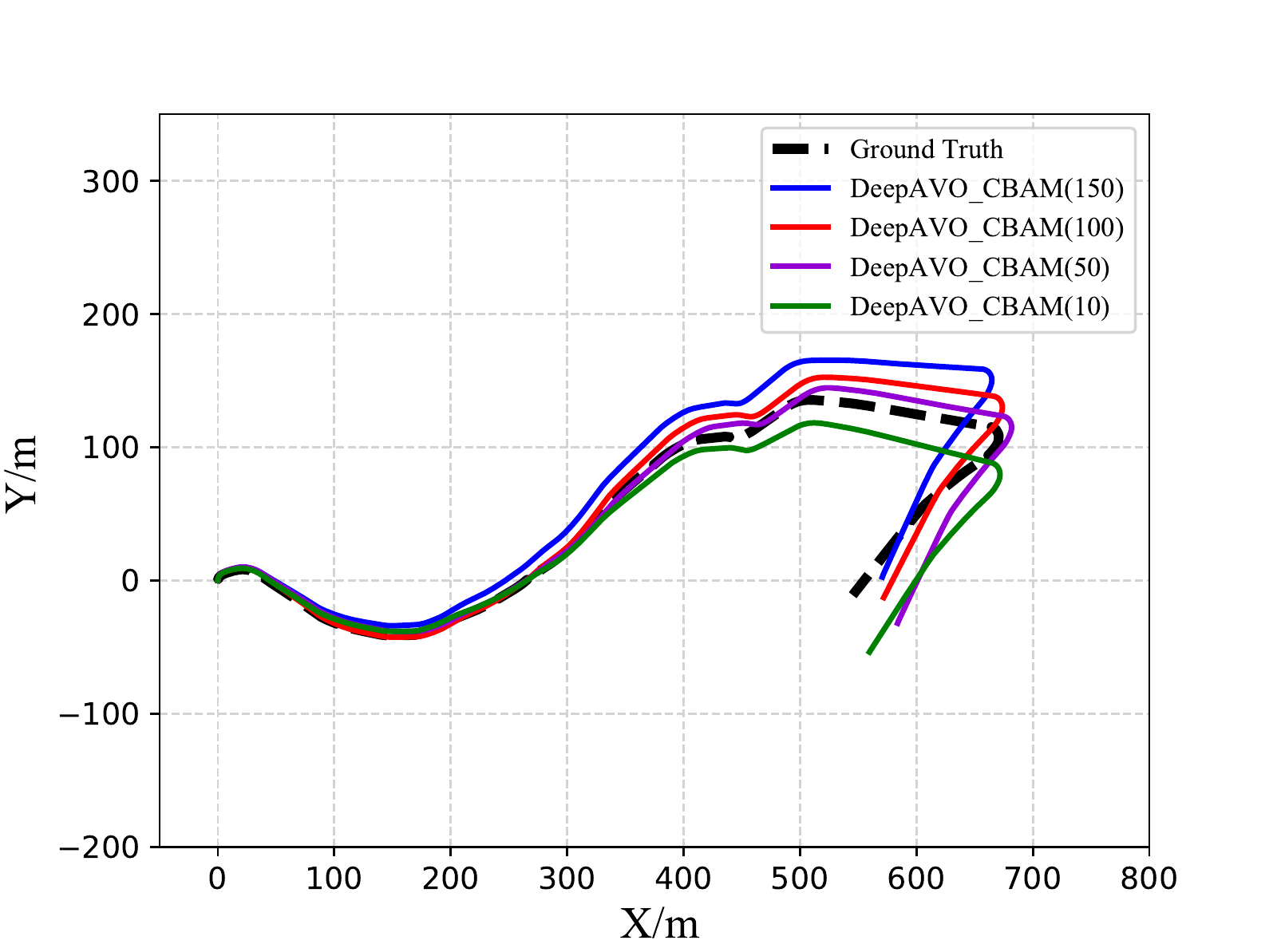}}
	\caption{The trajectories estimated by the models trained under the balance parameter ${\alpha}$ set to 10, 50, 100, and 150.}
	\label{Fig.balance}
\end{figure}

Fig. \ref{Fig.balance} illustrates the qualitative comparison. When the factor $\alpha$ is varied between 10 and 150, the performance of our model remains stable for the trajectories with less intense change in rotation (i.e., Sequences 03 and 10). In terms of the complex scenes (i.e., Sequences 05 and 07), however, the ego-motion estimation is sensitive to the factor $\alpha$. Through a trade-off between the accuracy in rotation and translation, we adopt $\alpha = 100$ as the final setting considering its promising results.
\begin{figure*}[!ht]
	\centering
	\subfigure[sequence 11]{
		\label{Fig.11}
		\includegraphics[width=0.31\textwidth,height=0.49\columnwidth]{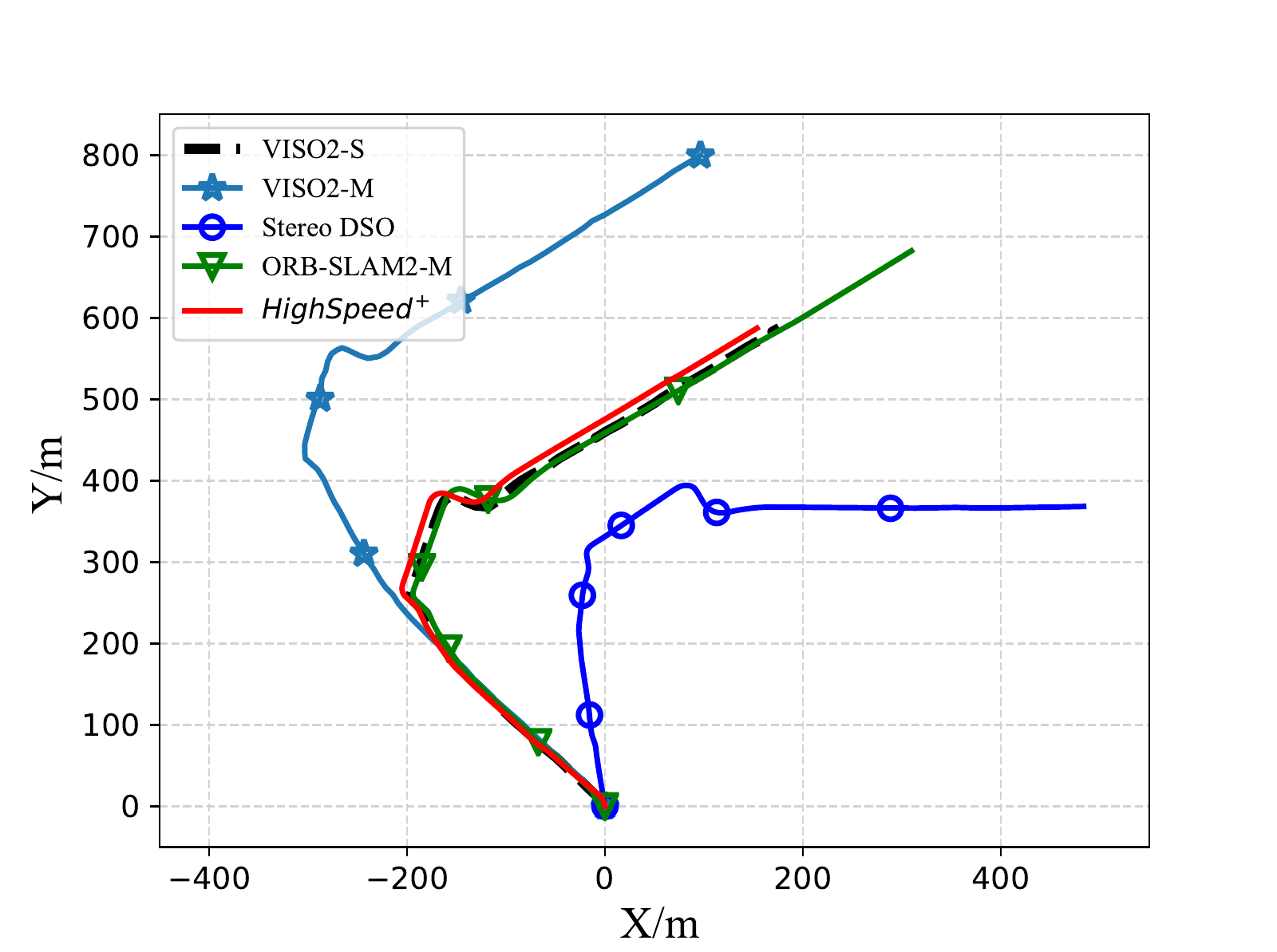}}
	\subfigure[sequence 12]{
		\label{Fig.12}
		\includegraphics[width=0.31\textwidth,height=0.49\columnwidth]{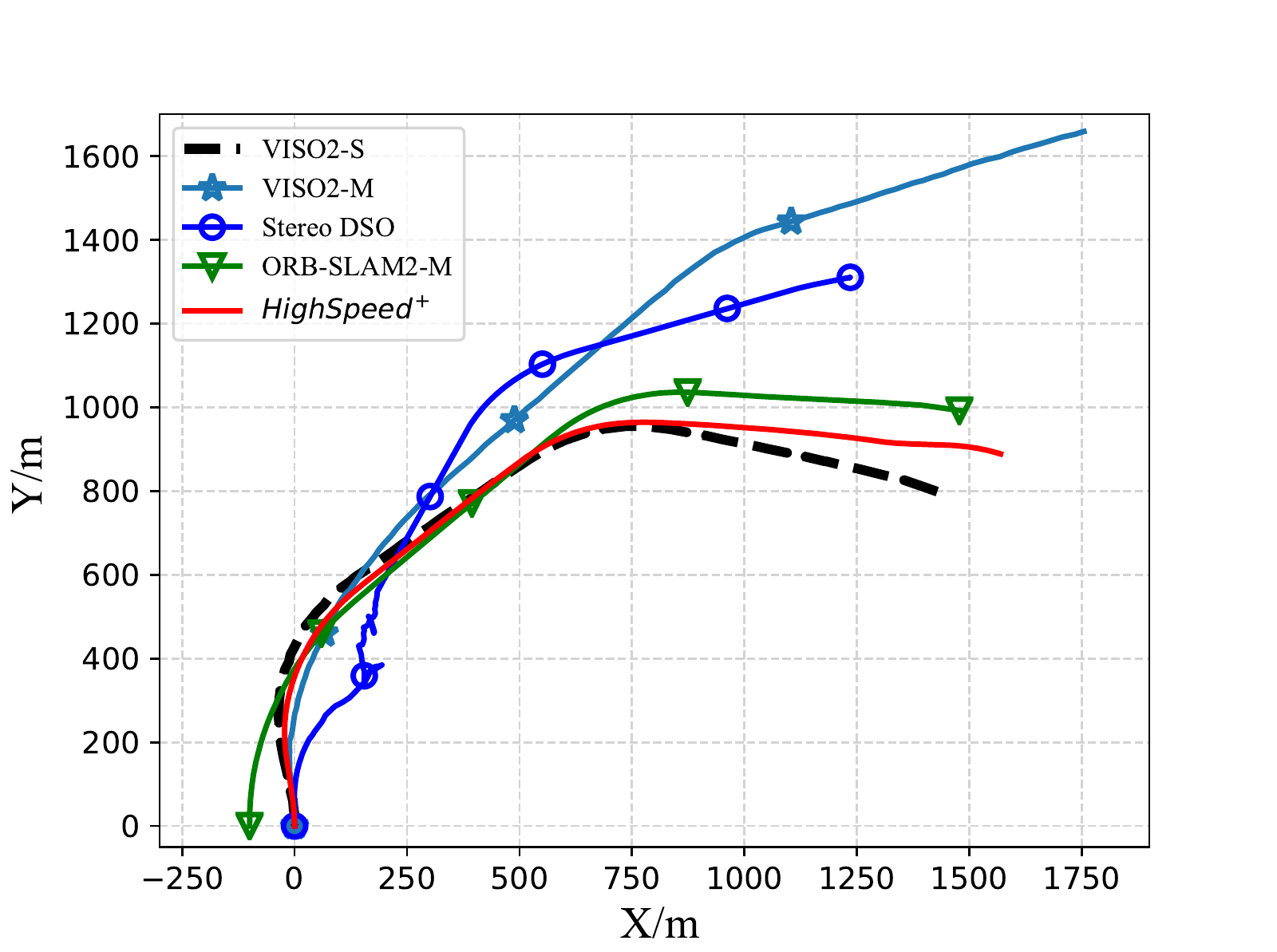}}
	\subfigure[sequence 13]{
		\label{Fig.13}
		\includegraphics[width=0.31\textwidth,height=0.49\columnwidth]{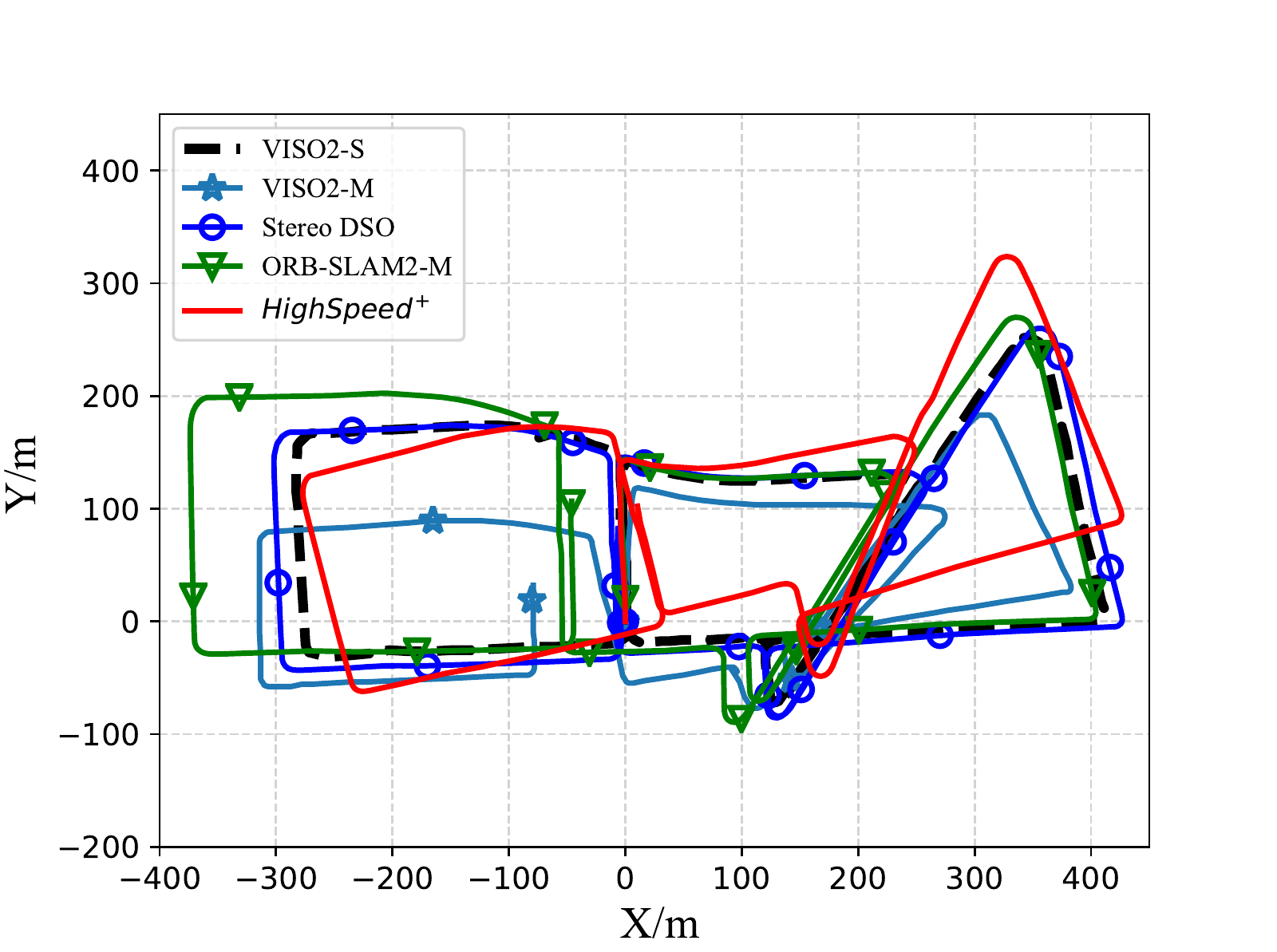}}
	\subfigure[sequence 14]{
		\label{Fig.14}
		\includegraphics[width=0.31\textwidth,height=0.49\columnwidth]{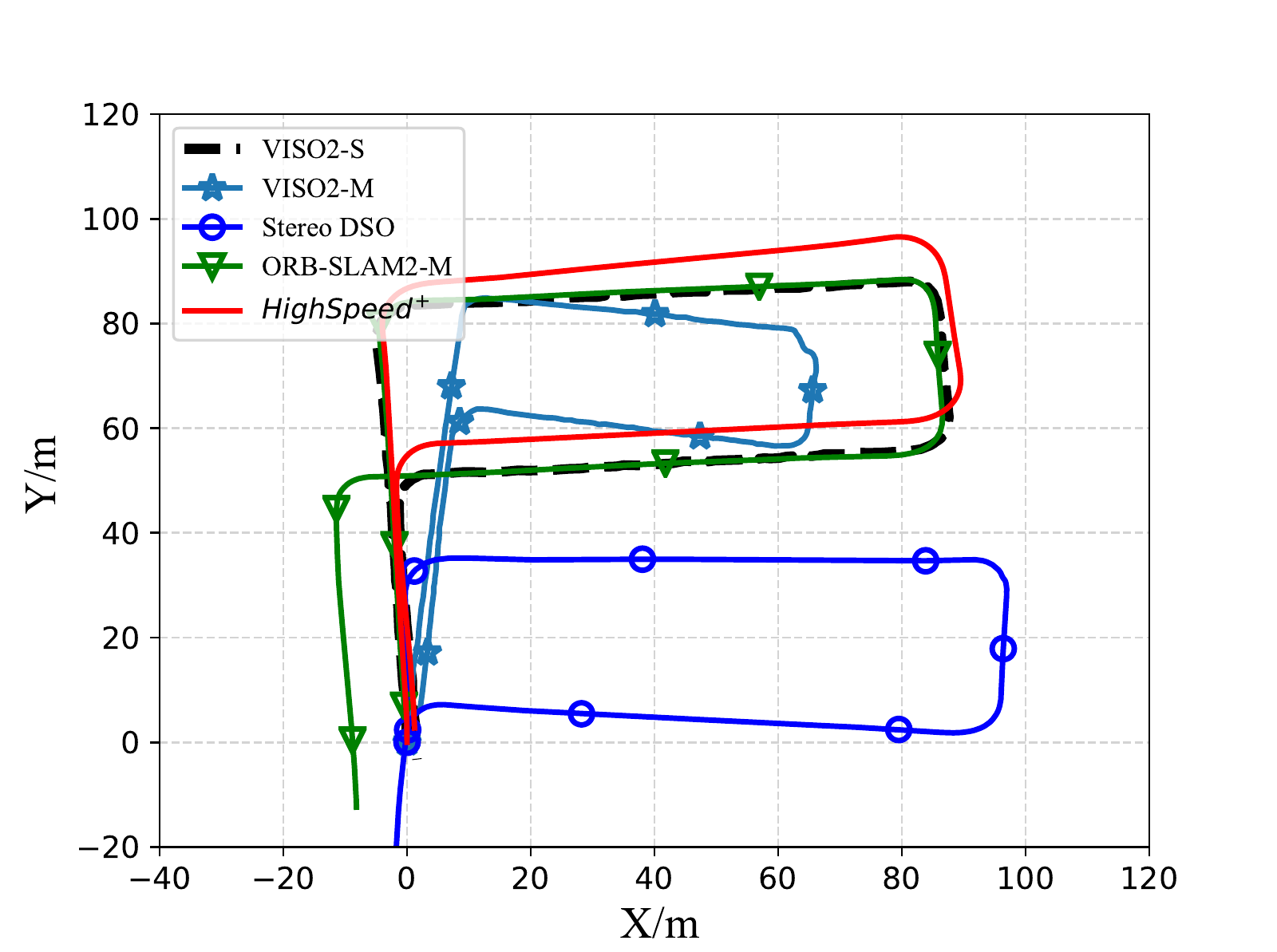}}
    \subfigure[sequence 15]{
		\label{Fig.15}
		\includegraphics[width=0.31\textwidth,height=0.49\columnwidth]{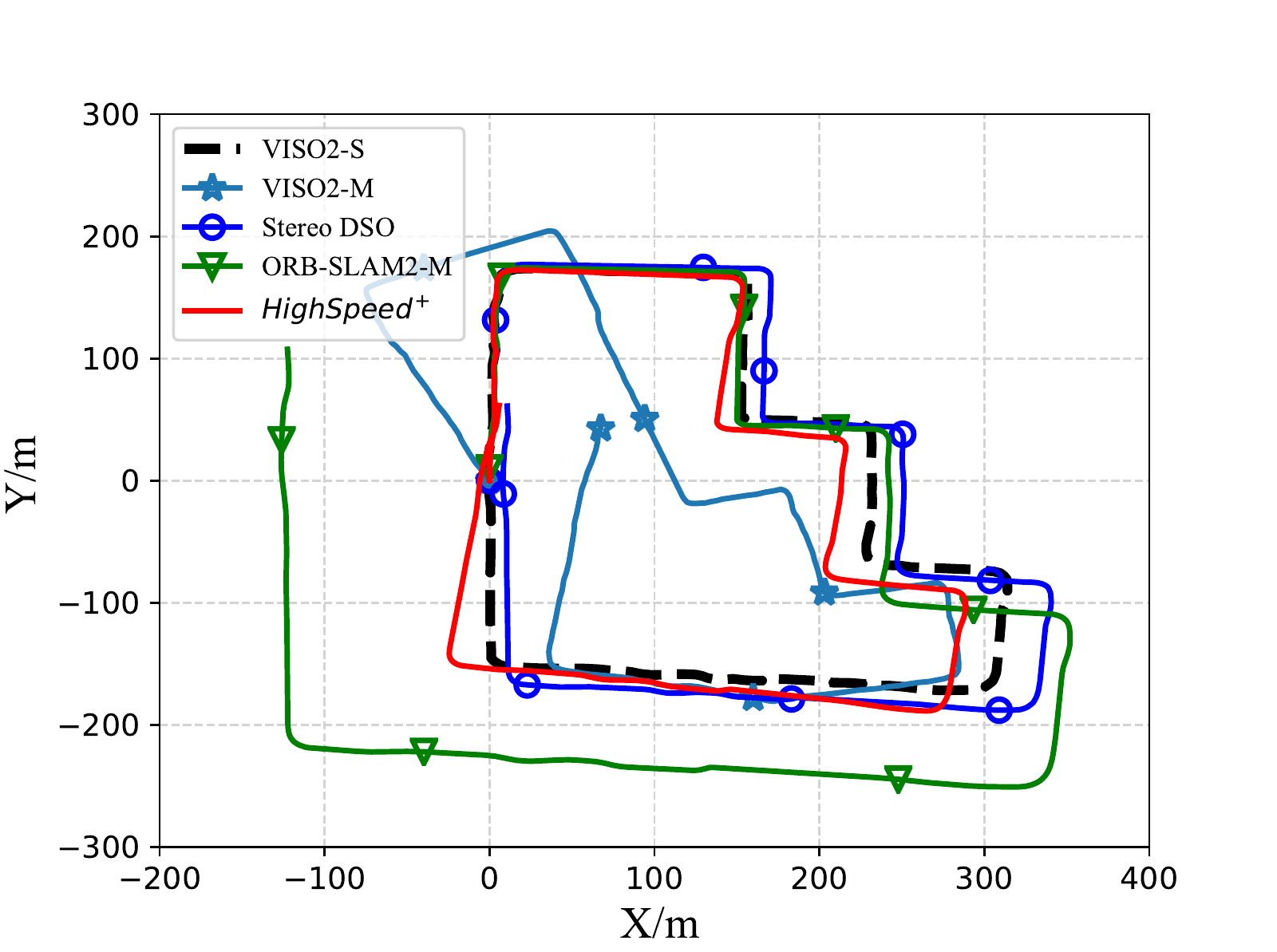}}
	\subfigure[sequence 16]{
		\label{Fig.16}
		\includegraphics[width=0.31\textwidth,height=0.49\columnwidth]{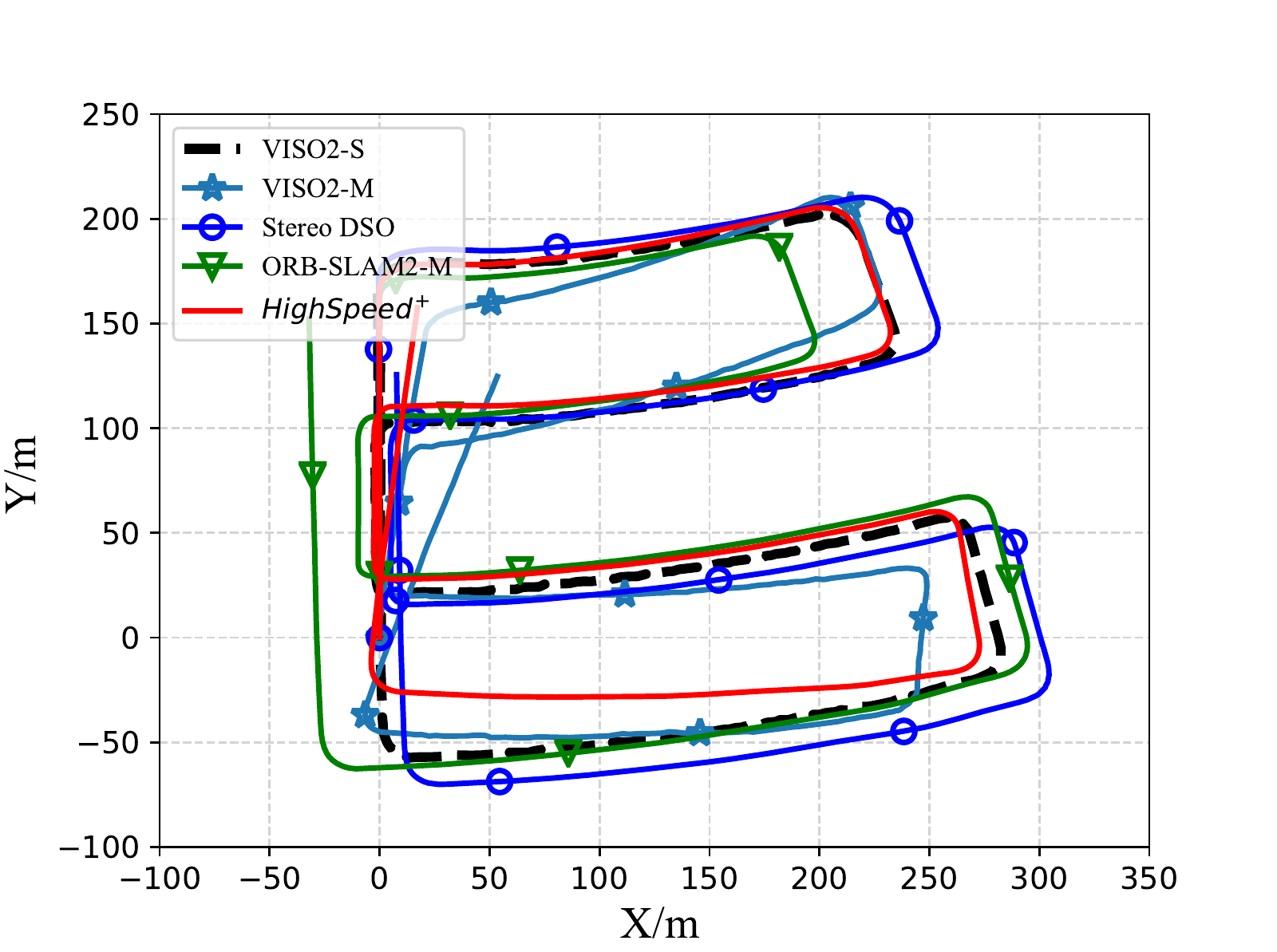}}
	\subfigure[sequence 17]{
		\label{Fig.17}
		\includegraphics[width=0.31\textwidth,height=0.49\columnwidth]{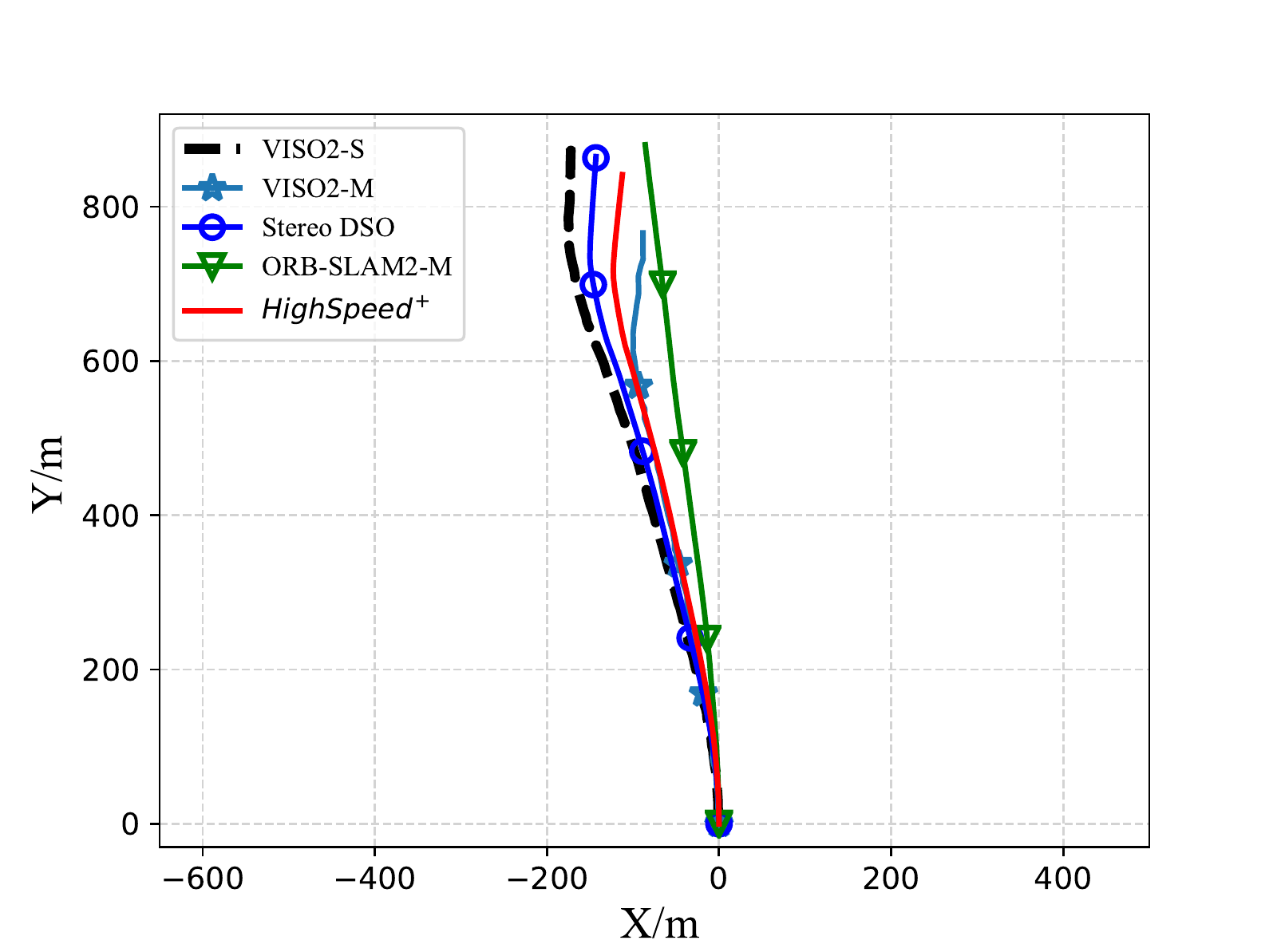}}
	\subfigure[sequence 18]{
		\label{Fig.18}
		\includegraphics[width=0.31\textwidth,height=0.49\columnwidth]{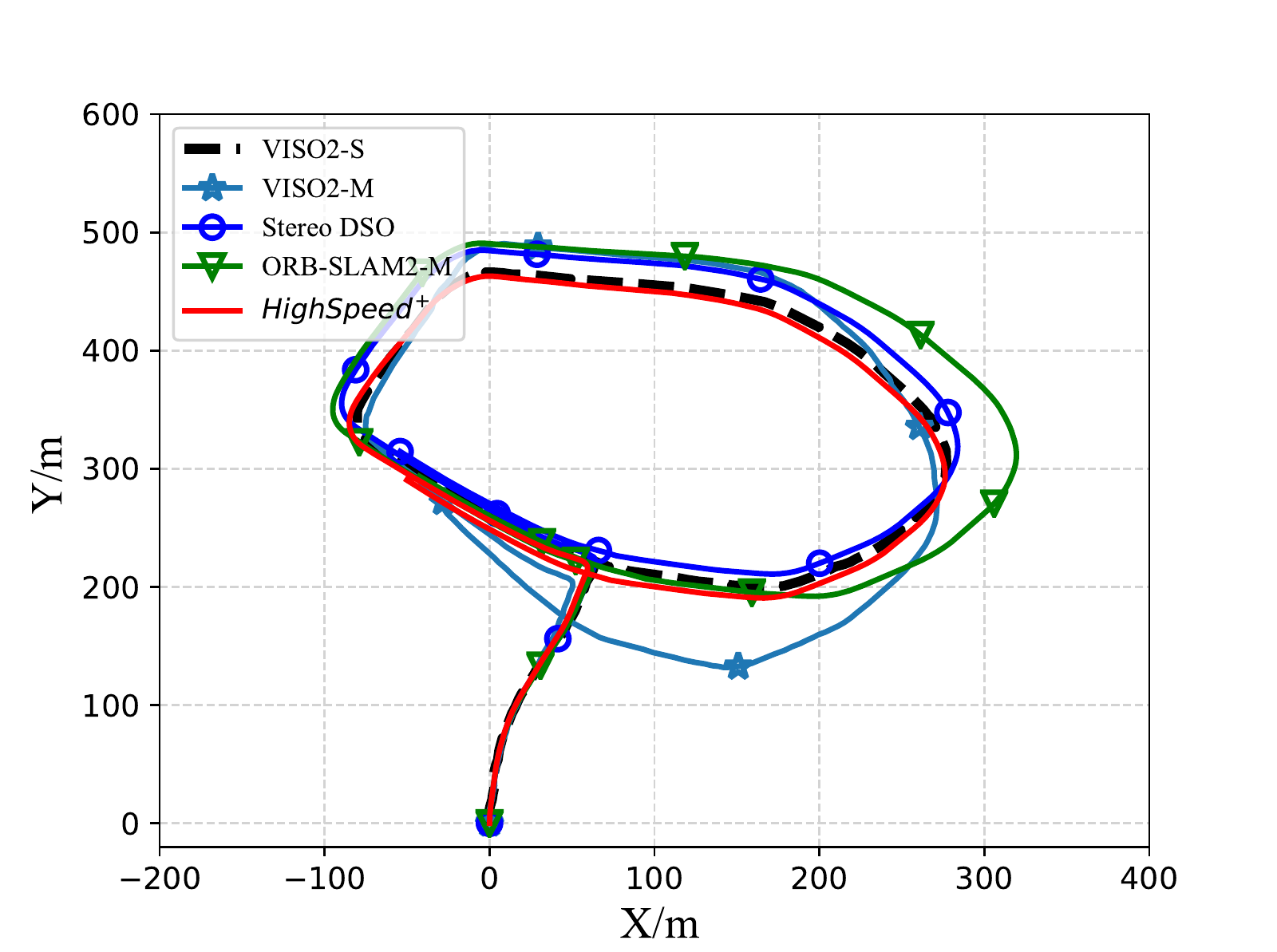}}
    \subfigure[sequence 19]{
		\label{Fig.19}
		\includegraphics[width=0.31\textwidth,height=0.495\columnwidth]{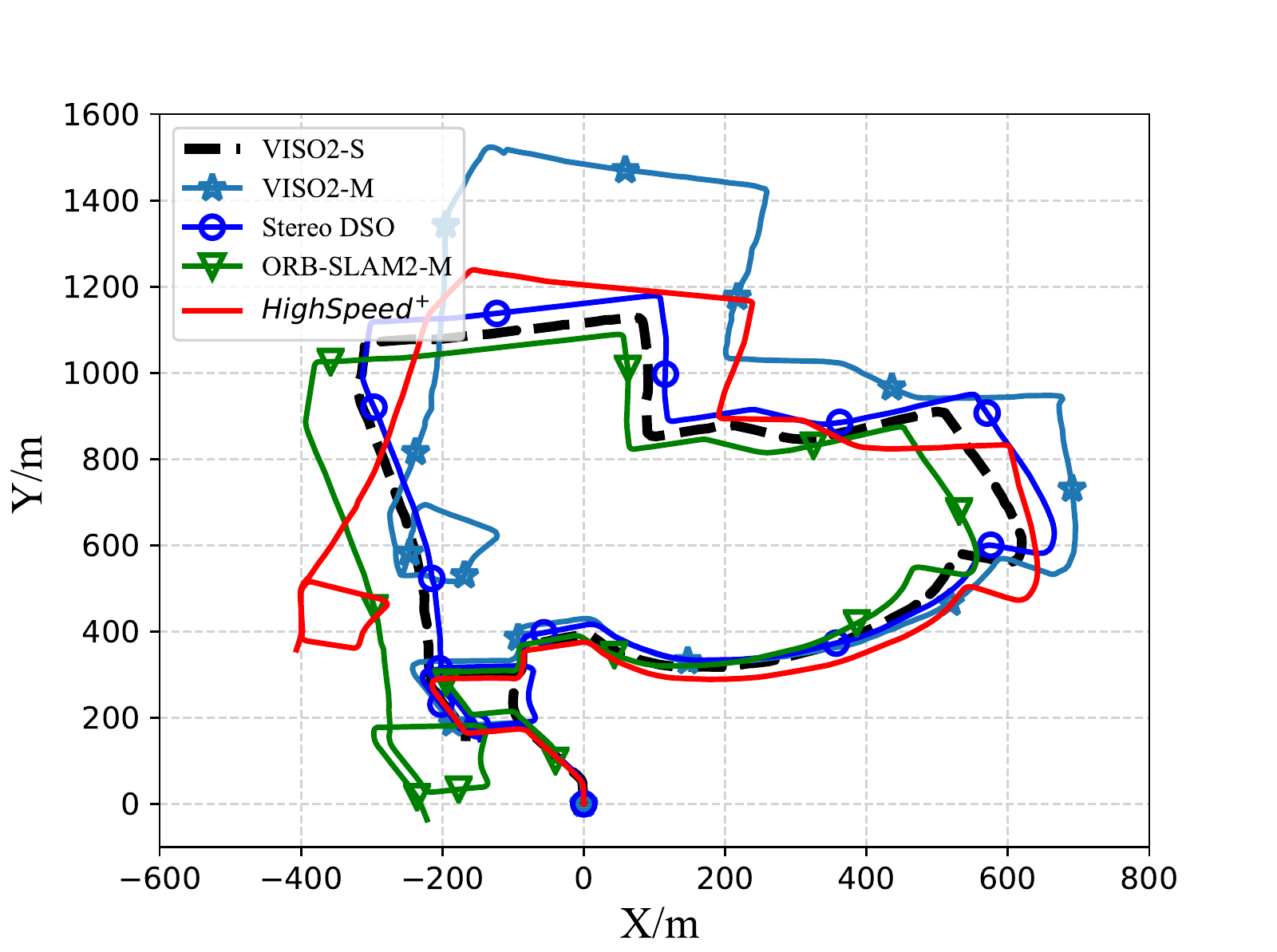}}
	\caption{Trajectories of VO results on the testing Sequence 11-19 of the KITTI VO benchmark (no ground truth is available for these testing sequences). The $HighSpeed^{+}$ model used is trained on the whole training dataset (00-10) and subsampled Sequence 00 of the KITTI VO benchmark, Its scales are recovered automatically from the neural network without alignment to ground truth. The \href{http://www.cvlibs.net/datasets/kitti/eval_odometry_detail.php?&result=88abc6589c77d89ee68a4080bb53d40094f22984}{results} have been submitted to KITTI website.}
	\label{Fig.11_20}
\end{figure*}

\begin{figure*}[!ht]
	\centering
	\subfigure[Malaga 03]{
		\label{Fig.m3_1}
	\includegraphics[width=0.31\textwidth,height=0.49\columnwidth]{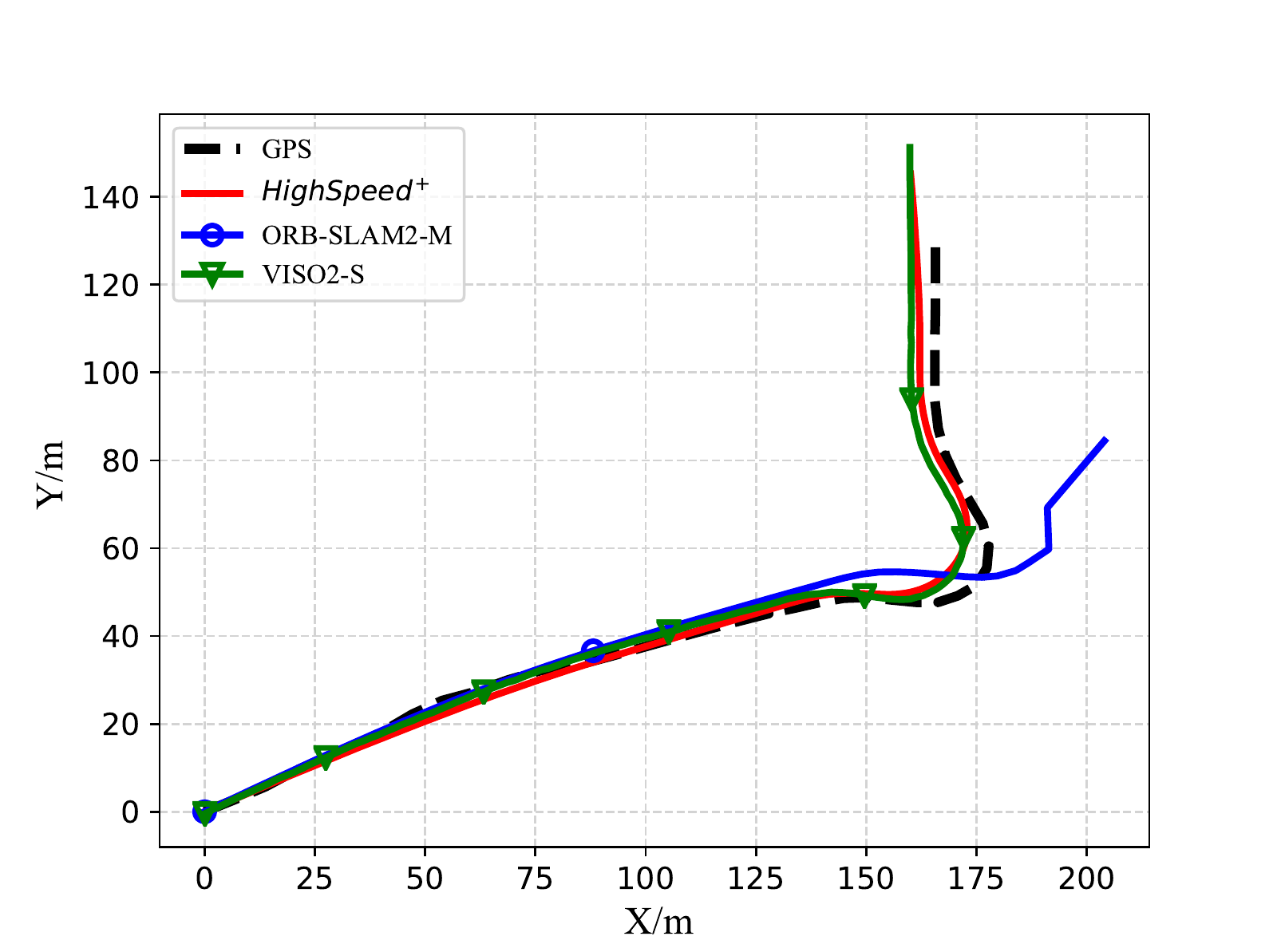}}
	\subfigure[Malaga 07]{
		\label{Fig.m5_1}
	\includegraphics[width=0.31\textwidth,height=0.49\columnwidth]{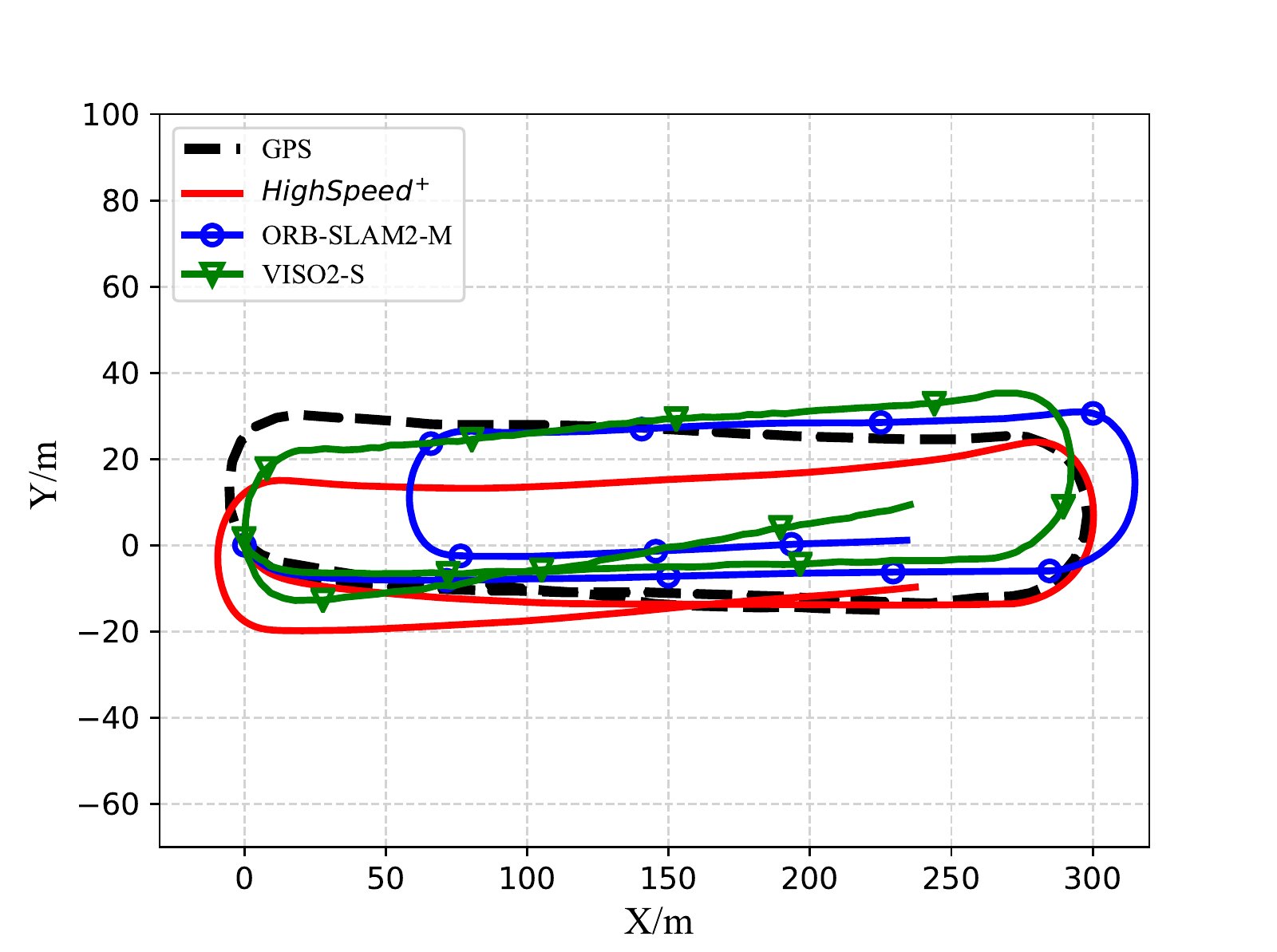}}
	\subfigure[Malaga 09]{
		\label{Fig.m7_1}
	\includegraphics[width=0.31\textwidth,height=0.49\columnwidth]{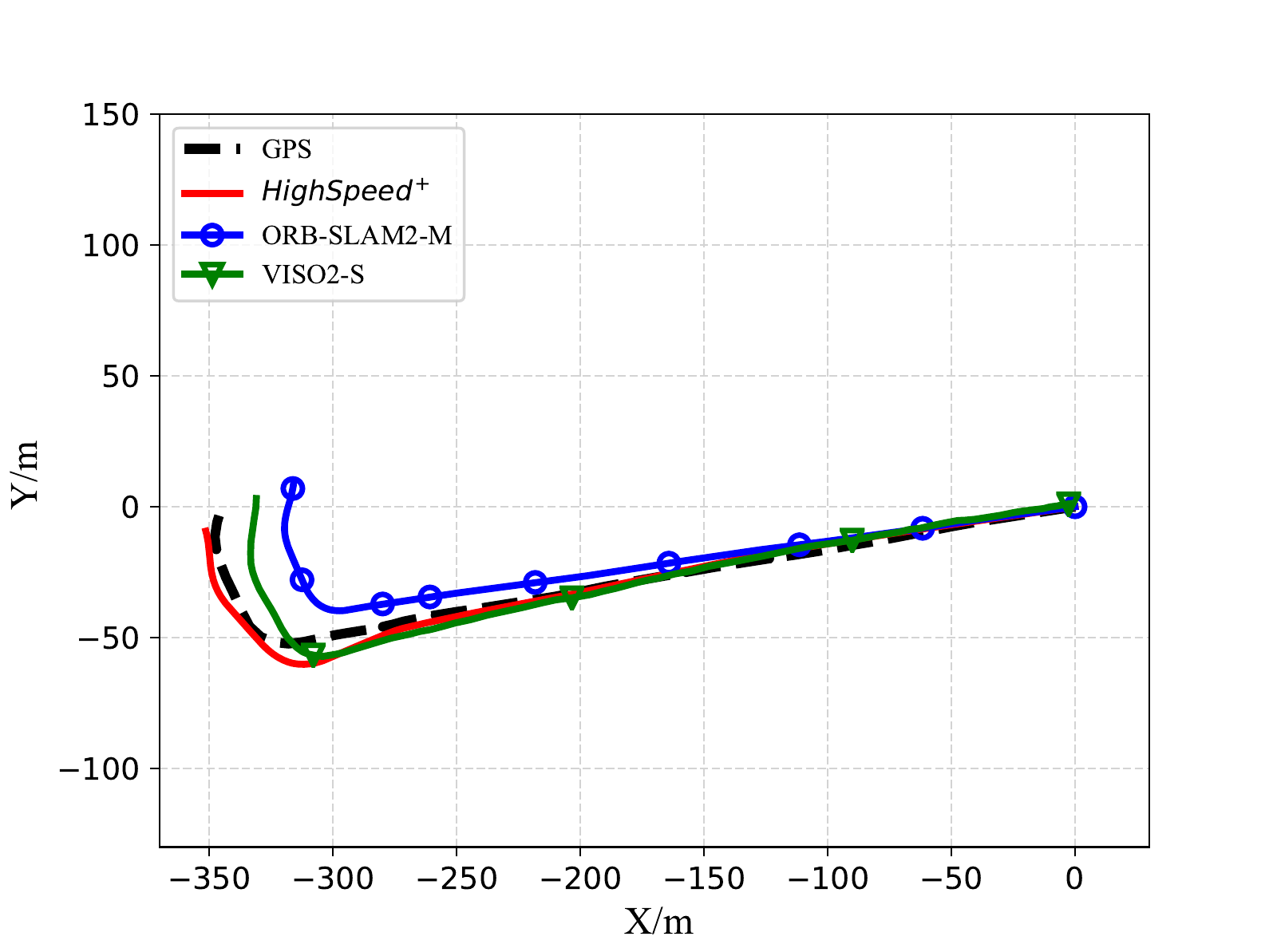}}
    \subfigure[Malaga 03]{
		\label{Fig.m3_11}
	\includegraphics[width=0.31\textwidth,height=0.49\columnwidth]{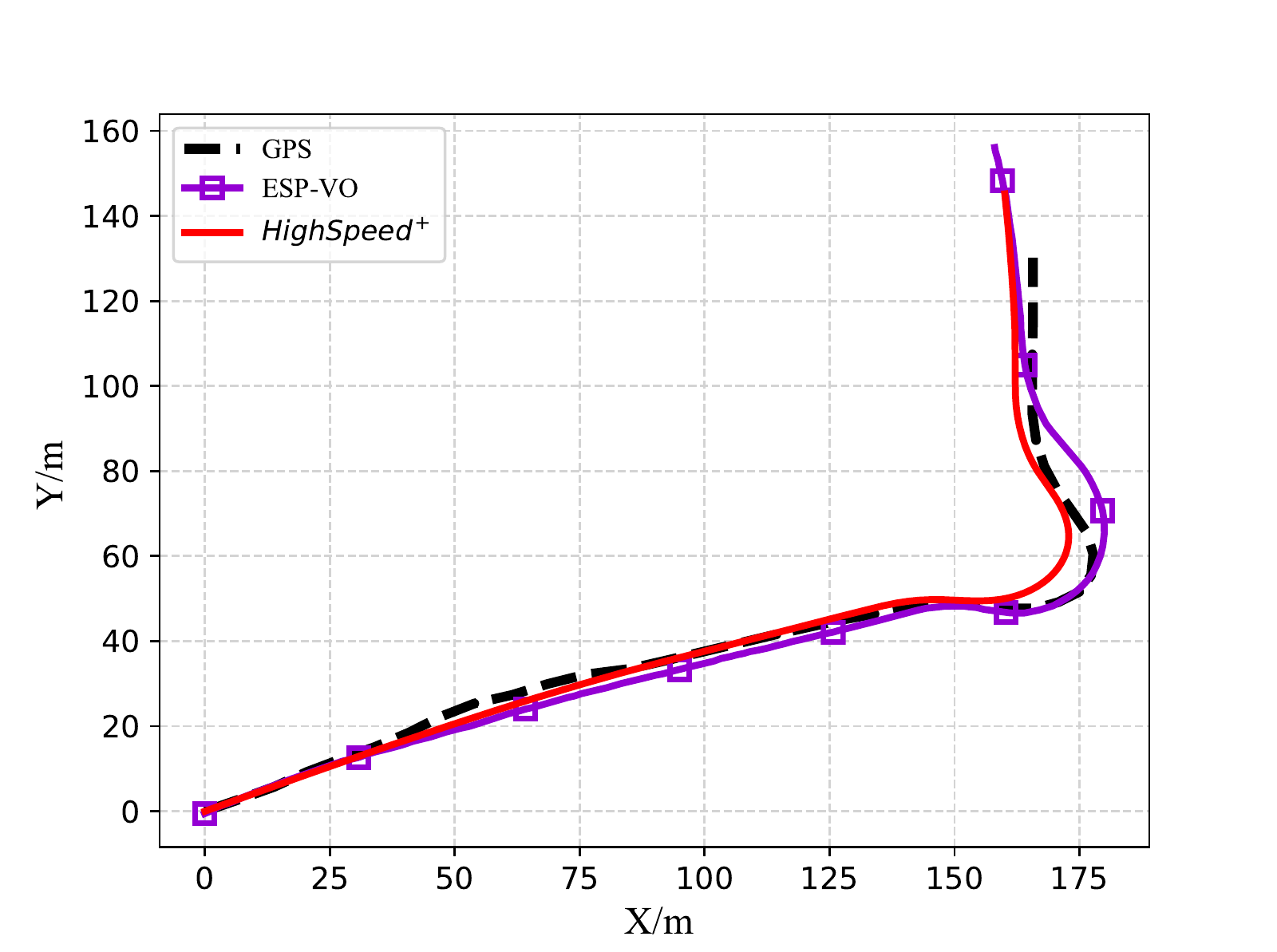}}
	\subfigure[Malaga 07]{
		\label{Fig.m5_11}
	\includegraphics[width=0.31\textwidth,height=0.49\columnwidth]{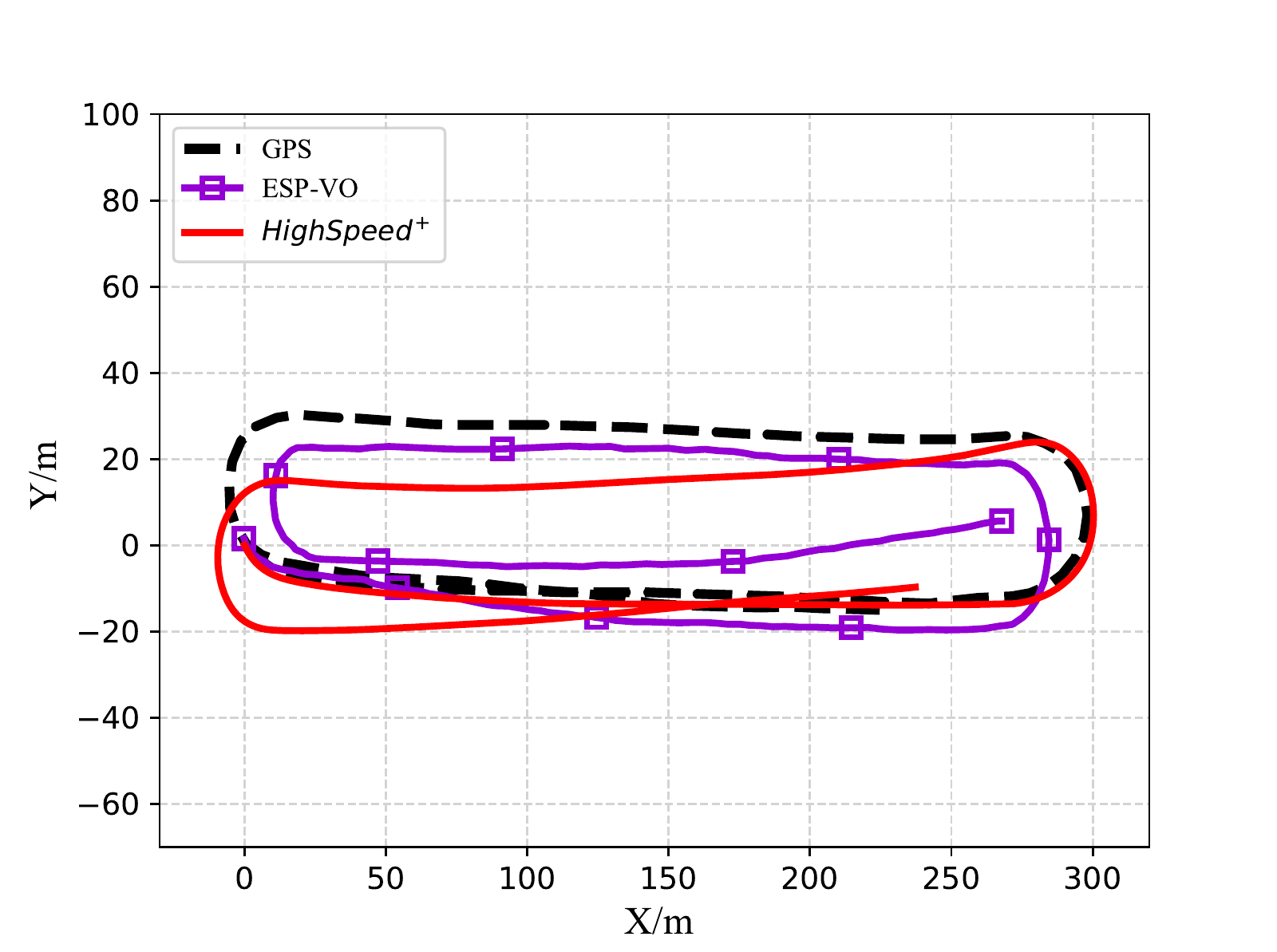}}
	\subfigure[Malaga 09]{
		\label{Fig.m7_11}
	\includegraphics[width=0.31\textwidth,height=0.49\columnwidth]{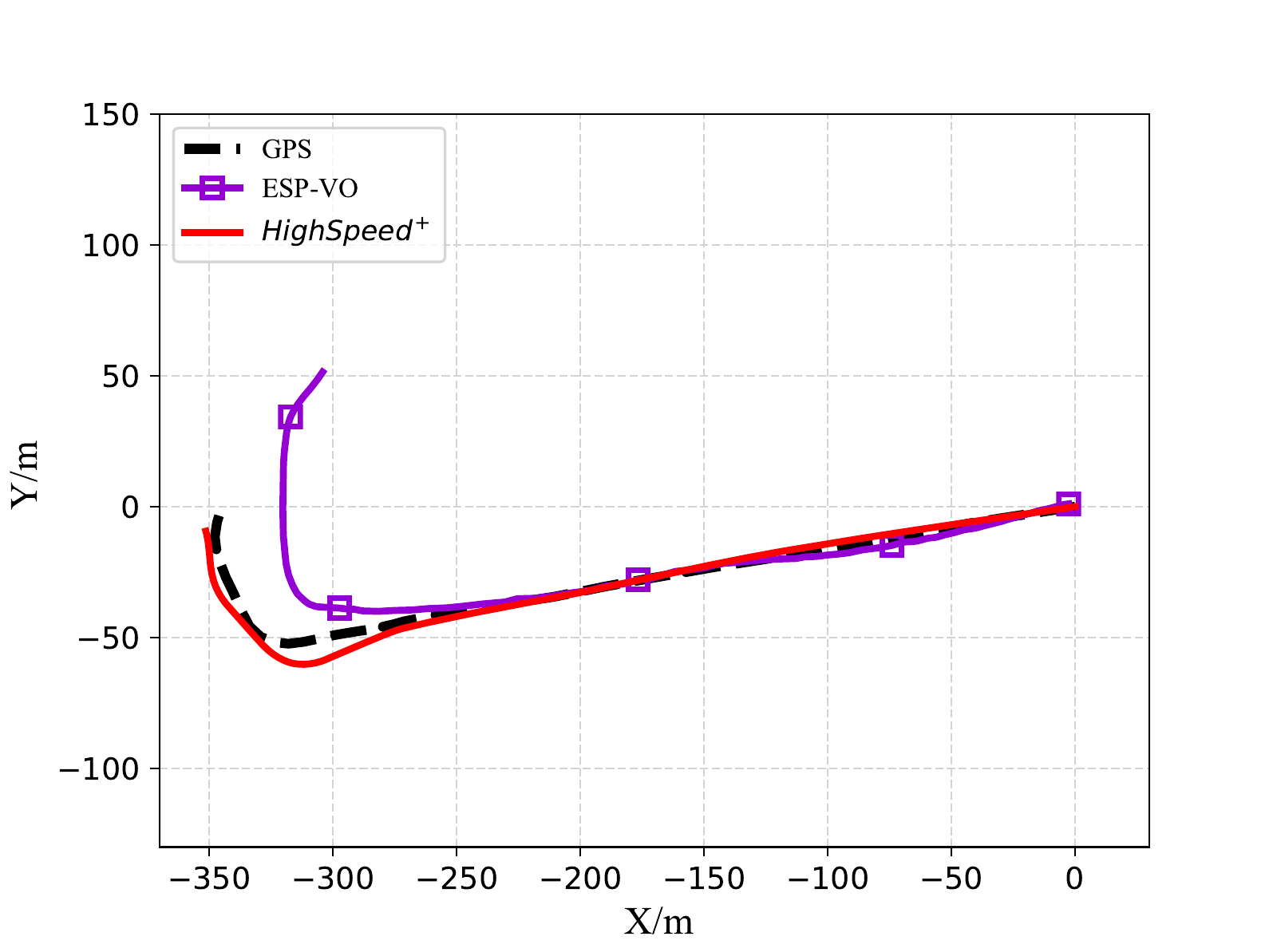}}
	\caption{Testing results on the Malaga dataset without any training or fine-tuning. The $HighSpeed^{+}$ used is only trained on Sequence 00-10 and the subsampled Sequence 00 of the KITTI.}
	\label{Fig.Malaga}
\end{figure*}

\subsubsection{Model generalization ability in the 11-19 sequence of KITTI}
\label{subsubsec:11-19}
Although the generalization of the $HighSpeed^{+}$ has been evaluated in the previous experiments, in order to investigate further how it performs in different motion patterns and scenes, the model is tested on Sequence 11-19 of the KITTI dataset. In this case, the $HighSpeed^{+}$ model is trained on Sequence 00-10 and the subsampled Sequence 00, providing more training samples to avoid overfitting and maximizing the generalization ability of the network. Due to the lack of ground truth for these testing sequences, similar to ESP-VO\cite{wang2018end}, we use stereo VISO2-S\cite{Geiger2012Stereoscan} as reference. Note that the \href{https://github.com/HorizonAD/stereo_dso}{stereo DSO} adopted in this paper is released by the \emph{Horizon Robotics} since its official version is not available.

The predicted trajectories are illustrated in Fig. \ref{Fig.11_20}. VISO2-M suffers from severe error accumulation, while monocular ORB-SLAM2-M\cite{mur2017orb}(without loop closure detection) partially alleviates the problem with the assistance of local bundle adjustment and a global map. Stereo DSO that can perform promising pose estimation on most test sequences has good generalization ability. It can be seen that the results of $HighSpeed^{+}$ are much better than VISO2-M's and roughly similar to the stereo VISO2-S's. It seems that this larger training dataset improves the performance of $HighSpeed^{+}$. Considering the stereo characteristics of stereo VISO2-S, $HighSpeed^{+}$, as a monocular VO, has achieved appealing results, indicating that the trained model has a good generalization ability in new scenes. We have submitted the \href{http://www.cvlibs.net/datasets/kitti/eval_odometry_detail.php?&result=88abc6589c77d89ee68a4080bb53d40094f22984}{reconstructed trajectories} on Sequence 11-21 to the odometry benchmark of the KITTI website for an open and fair comparison with existing methods.

\subsection{Results on Malaga and ApolloScape datasets}
\label{subsec:Malaga results}
Malaga urban dataset\cite{blanco2014malaga} and ApolloScape dataset\cite{wang2018dels}, similar to the KITTI dataset, are gathered entirely in urban scenarios by the sensors mounted on the vehicle. Malaga dataset provides stereo images captured at 20Hz along with data from IMU, GPS, etc. Note that the images of Malaga in the size of $1024 \times 768$ have to be resized and then cropped to fit the resolution in KITTI. ApolloScape dataset contains a large number of monocular video clips captured in different lighting conditions (i.e., morning, noon, and night) for self-localization. Similar to Malaga dataset, the ApolloScape dataset is only used to test models.

\begin{figure}[!h]
	\centering
	\subfigure[Road 11]{
		\includegraphics[width=0.226\textwidth,height=0.35\columnwidth]{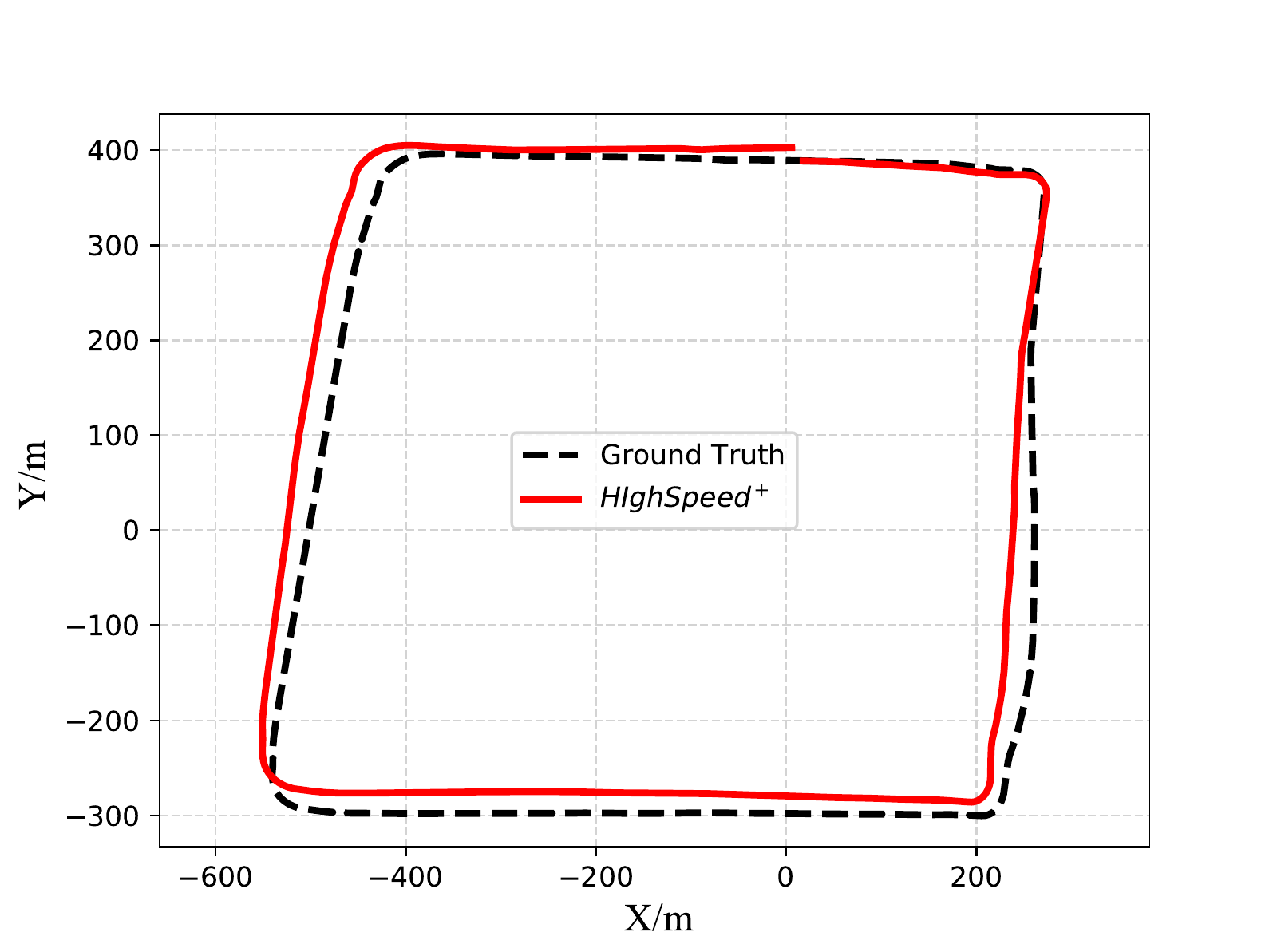}}
	\subfigure[Road 12]{
		\includegraphics[width=0.226\textwidth,height=0.35\columnwidth]{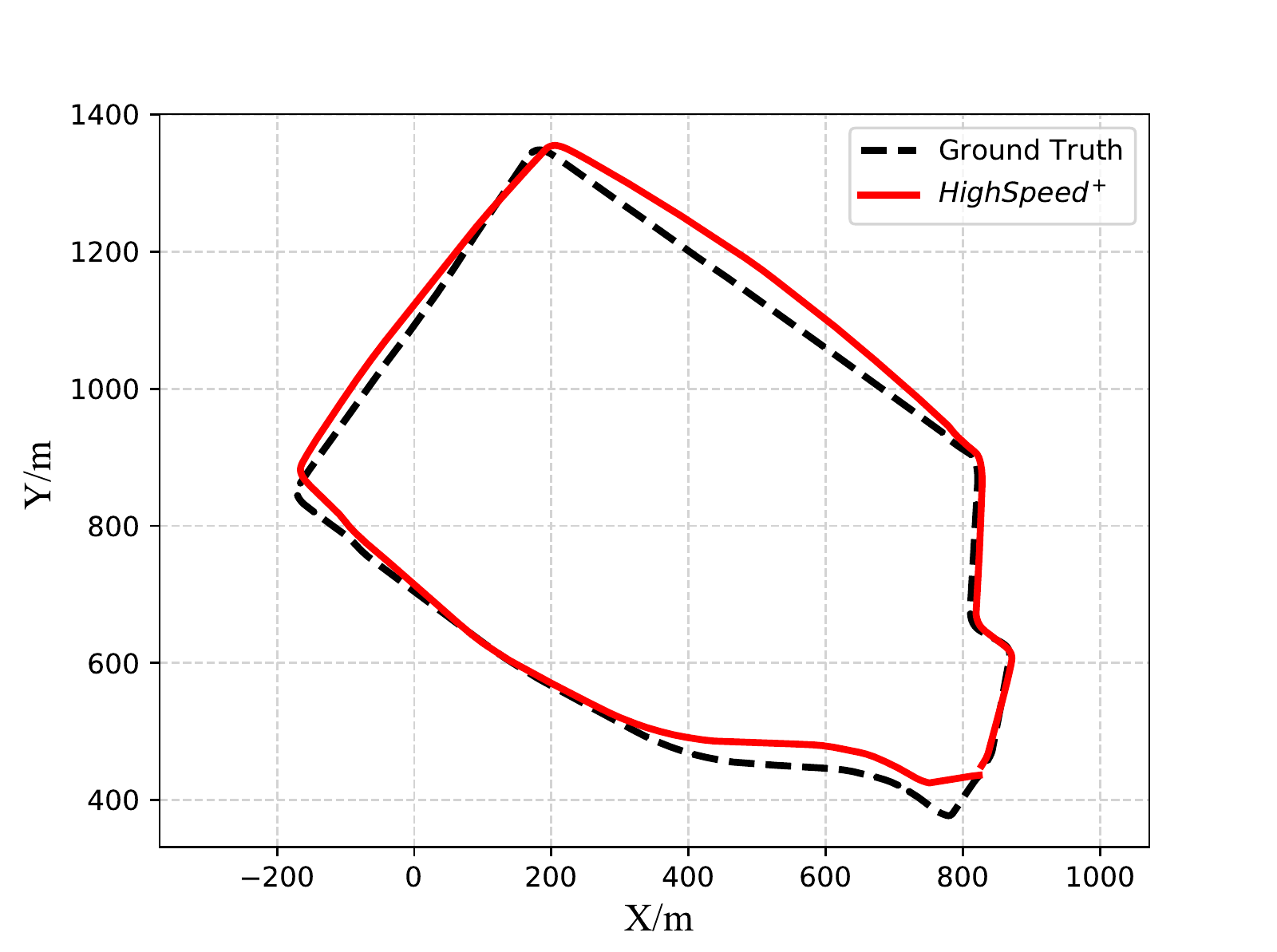}}
	\subfigure[Road 14]{
		\includegraphics[width=0.226\textwidth,height=0.35\columnwidth]{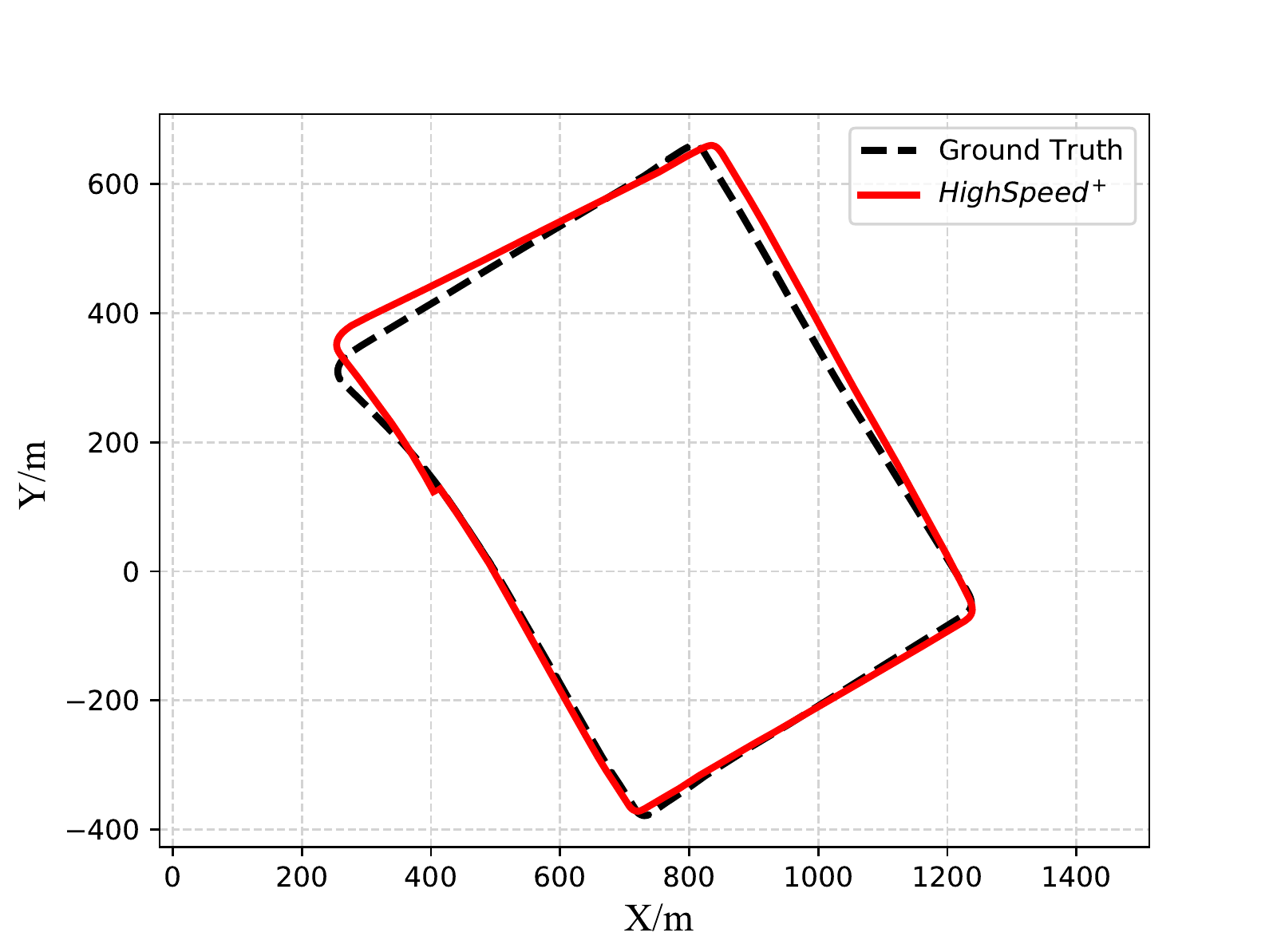}}
	\subfigure[Road 15]{
		\includegraphics[width=0.226\textwidth,height=0.35\columnwidth]{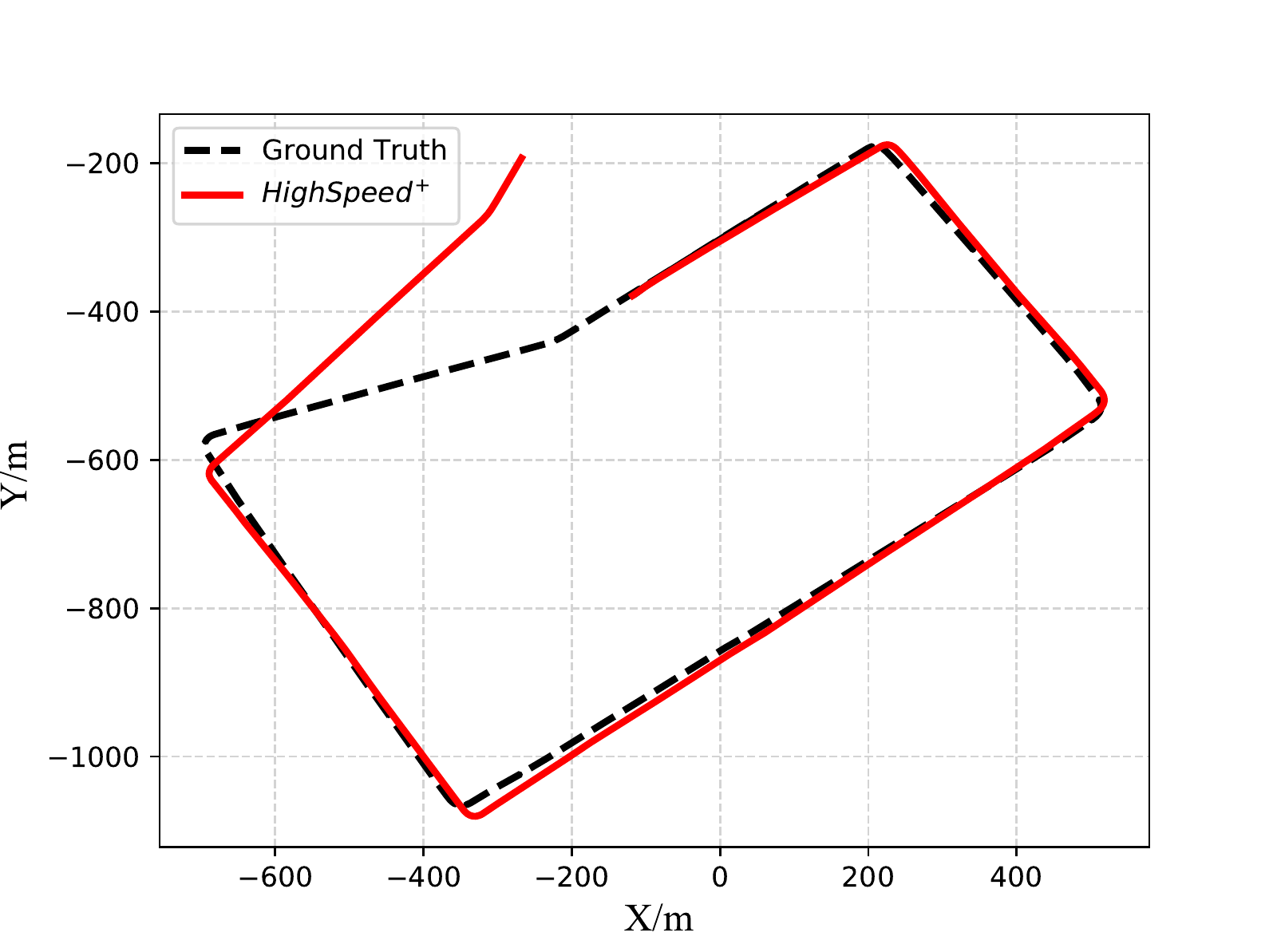}}
	\caption{Testing results on the ApolloScape dataset without any training or fine-tuning. The $HighSpeed^{+}$ used is only trained on Sequence 00-10 and the subsampled sequence 00 of the KITTI.}
	\label{Fig.apollo}
\end{figure}

Compared with the experiments in Sec. \ref{subsubsec:11-19}, the verification of generalization ability of $HighSpeed^{+}$ through Malaga and ApolloScape datasets is more convincing, since 1) it is the cross-dataset validation under entire new scenarios and hardware platform for data collection; 2) the $HighSpeed^{+}$ model in this experiment is consistent with the one in Sec. \ref{subsubsec:11-19} of which the training dataset is only derived from KITTI dataset (i.e., Sequence 00-10 and subsampled Sequence 00) without extra data augmentation or fine-tuning. Fig. \ref{Fig.Malaga} and Fig. \ref{Fig.apollo} shows the testing results on the Malaga (Malaga 03, 07, and 09 sequences) and ApolloScape (Road 11, 12, 14, and 15 sequences) datasets. Sparse GPS ground truth is available for Malaga sequences, while ApolloScape dataset provides the ground truth calibrated by multiple combined sensors.

As for Malaga dataset, we can see that $HighSpeed^{+}$ outperforms the ORB-SLAM2-M and learning-based ESP-VO. The pose estimated by $HighSpeed^{+}$ is close to VISO2-S's, both of which approximate the trajectories reconstructed by GPS, no matter in the regular or complicated scenes. As for ApolloScape dataset, we only compare the trajectories provided by $HighSpeed^{+}$ and ground truth as the unusual size of images ($3384\times2710$) always introduces failed initialization for stereo DSO and ORB-SLAM2-M. As shown in Fig. \ref{Fig.apollo}, it can be observed that the performance of $HighSpeed^{+}$ is outstanding in handling various Road sequences except for the last big turn in Road 15. 

\subsection{Results on self-collected indoor dataset}

We also conduct the experiment based on the self-collected dataset to evaluate the $HighSpeed^{+}$ model for indoor positioning. The monocular RGB images are collected in an office building environment using Intel Realsense D455 camera running on the Ubuntu system, as
\begin{figure}[!h]
	\centering
    \includegraphics[width=0.4\textwidth,height=0.63\columnwidth]{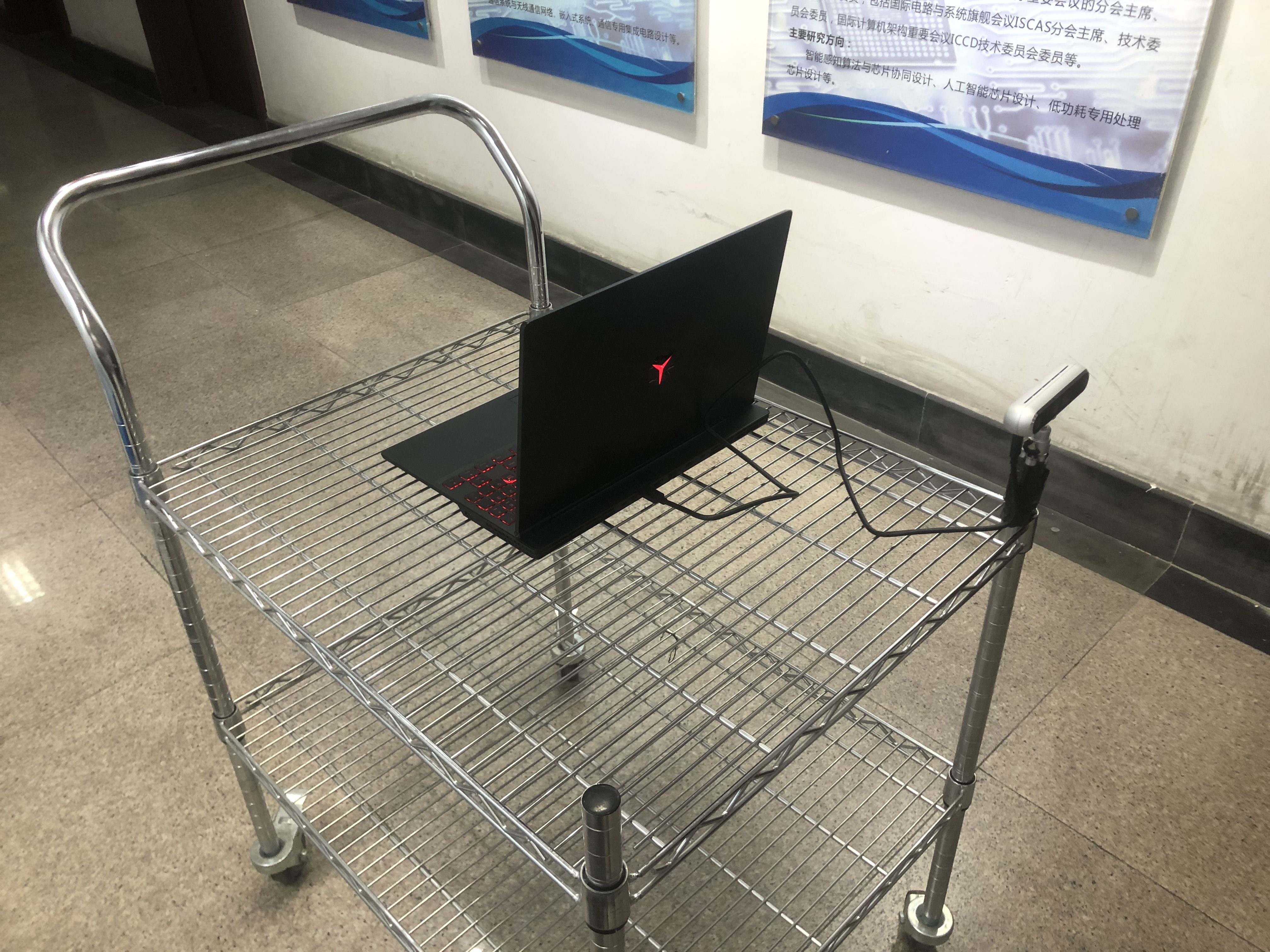}
	\caption{Indoor dataset collecting platform. }
	\label{Fig.selfpic}
\end{figure}shown in Fig. \ref{Fig.selfpic}. Unlike the previous experiments, the constructed trajectories by $HighSpeed^{+}$ cannot recover the absolute scale for the new indoor data collecting platform. Therefore, its predicted poses are aligned to ground truth by using similarity transformation.

\begin{figure}[!h]
	\centering
    \includegraphics[width=0.491\textwidth,height=0.83\columnwidth]{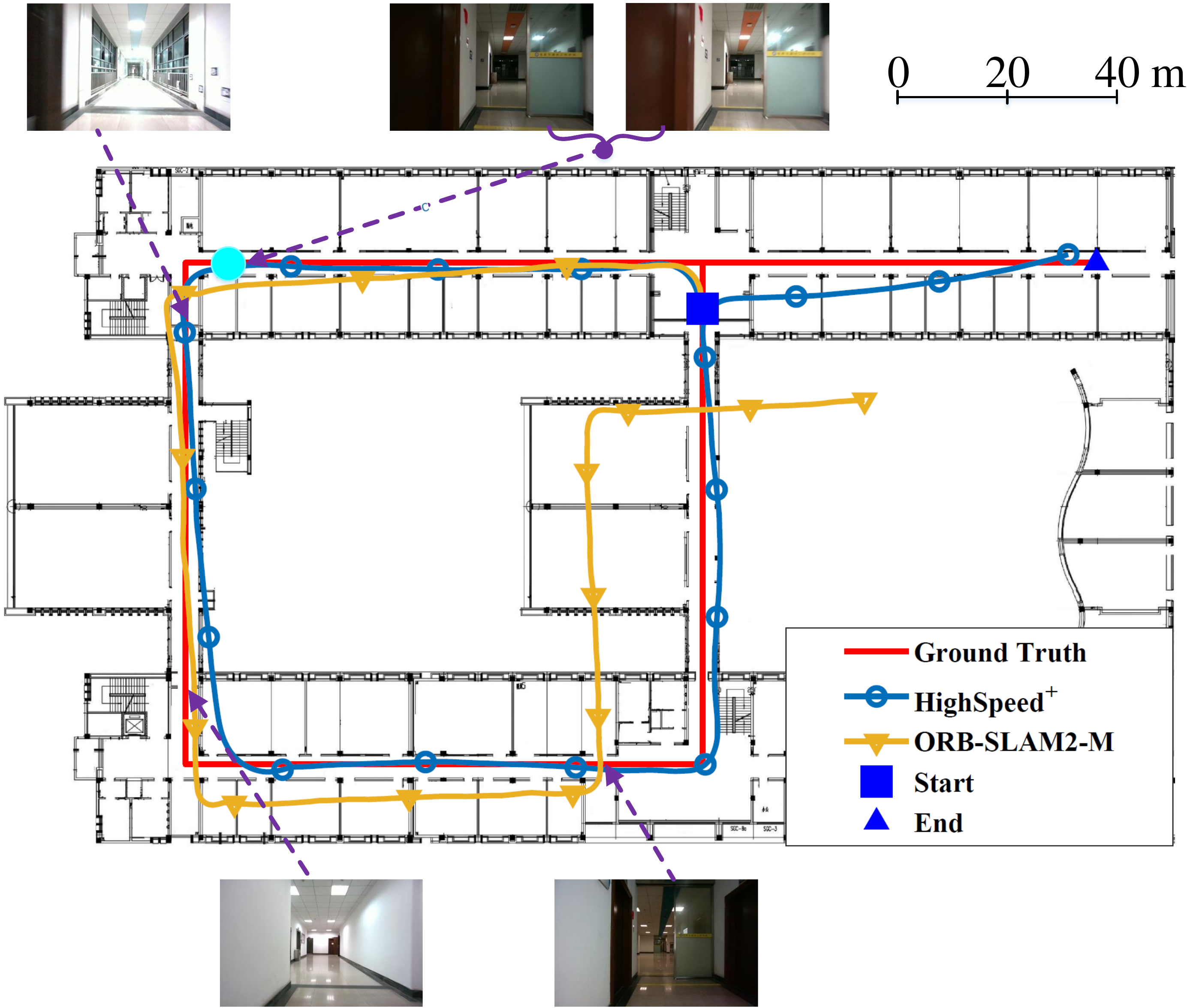}
	\caption{Testing results of $HighSpeed^{+}$ and sample images in an office building environment. The two consecutive images around the dot point show the drastic changes in illumination between frames.}
	\label{Fig.self}
\end{figure}

The reconstructed trails are shown in Fig. \ref{Fig.self} along with some sample images. It can be seen that the dataset is very challenging for monocular VO because the images are captured under different lighting conditions, and  some of them mostly contain texture-less white walls in narrow corridors. Nevertheless, $HighSpeed^{+}$ still maintains the tracking that suffers from light drifts. We also attempt to run monocular DSO and ORB-SLAM2-M on this dataset, but DSO failed to initialize and could not finish localization. Hence, we only provide the estimated results of ORB-SLAM2-M as the comparison.

\subsection{Computational Cost}
\label{subsec:Computation Cost}
Since real-time operation is critical for robotics applications such as autonomous driving, and learning-based methods are generally considered to be computationally expensive and time-consuming, we also compare the real-time performance of the DeepAVO model and ESP-VO. An NVIDIA Geforce Titan XP GPU and a desktop (Intel(R) Core(TM) i7-8700 CPU@3.20GHz and 16GB RAM) are used to compute the runtime of online inference on GPU and CPU, respectively.
\begin{figure}[!h]
	\centering
	\includegraphics[width=0.45\textwidth,height=0.27\textheight]{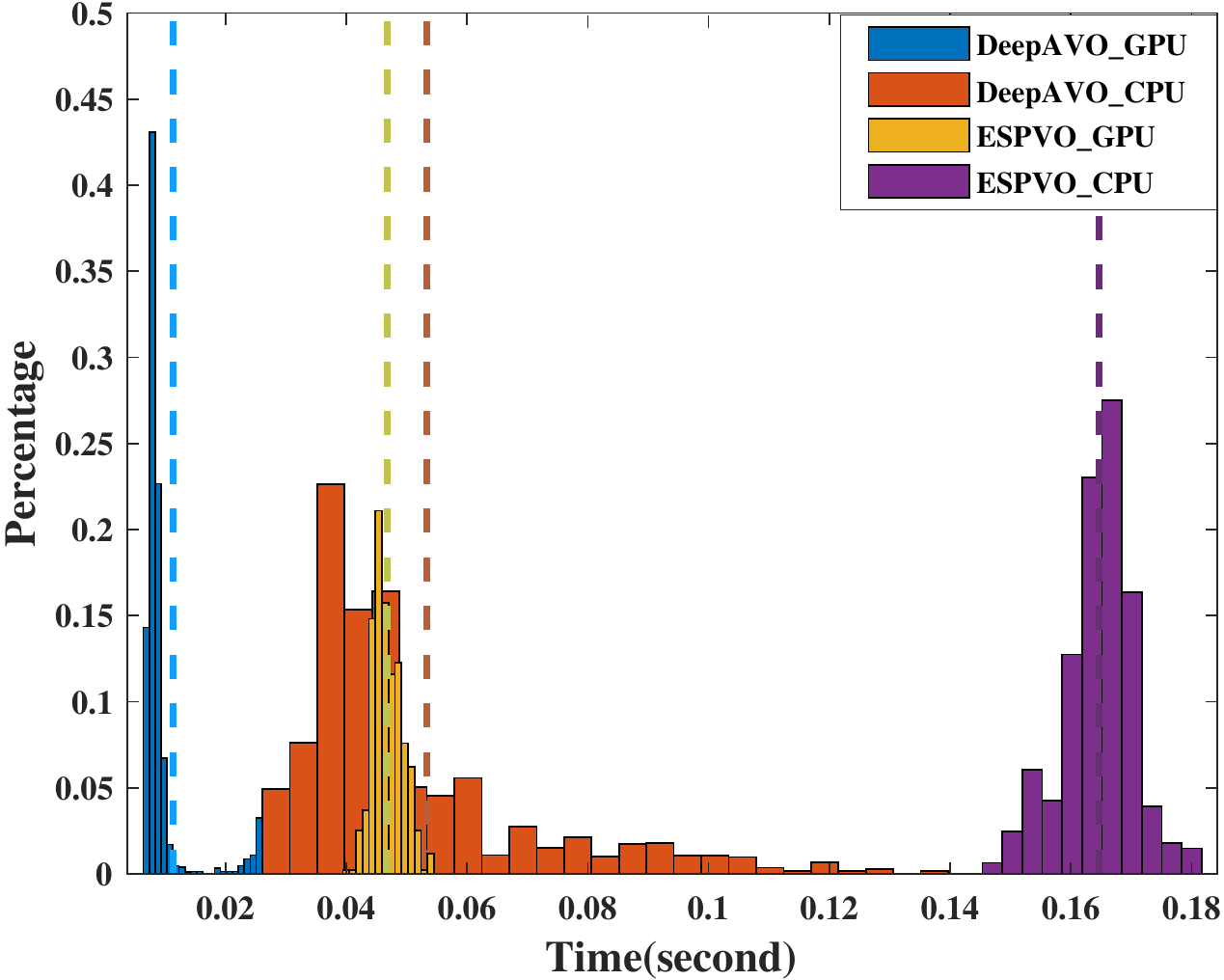}
	\caption{Time cost distribution of DeepAVO and ESP-VO on CPU and GPU.}
	\label{Fig.time}
\end{figure}

There are 1000 consecutive frames selected from the KITTI dataset involved in the time consumption statistics. The histogram of per-frame runtime in second on both GPU and CPU is shown in Fig. \ref{Fig.time}. Note that this time analysis for DeepAVO only involves odometry calculation. It can be seen that ESP-VO runs at about 20 Frame Per Second ($fps$) on GPU and 6$fps$ on CPU, while DeepAVO runs 5$ms$ to 30$ms$ per frame on the GPU and 30$ms$ to 140$ms$ on the CPU. The average per-frame runtime is about 12$ms$ and 53$ms$ on GPU and CPU, respectively. Optical flow calculation takes 30$ms$ per frame. Therefore, DeepAVO is capable of running up to 24$fps$ on GPU and 12$fps$ on CPU, which is faster than ESP-VO and far meet the demand for real-time positioning under the sampling rate of 10 $Hz$.

\section{Conclusion}
\label{sec:conclusion}

In this paper, we present a novel framework that contains four parallel CNNs focusing on four quadrants of optical flow for learning monocular visual odometry in an end-to-end fashion. In the framework, we incorporate a helpful attention component called CBAM, which distills the feature extracted by the Encoder in terms of channel and spatial aspects and ameliorates previous results. The refined features propagating global information through concatenating local cues of four branches further improve the pose estimation. The extensive experiments based on three datasets collected in outdoor environments by car and an indoor environment by cart verify that the DeepAVO outperforms many learning-based and traditional monocular VO methods and gives competitive results against the classic stereo algorithms, which highlights the promising generalization ability of the model. Besides, based on the computational cost analysis, it has been demonstrated that the DeepAVO can produce accurate and generalized results with low computational consumption.

In the future, we will focus on developing a complete SLAM system utilizing the attention mechanism and introduce sequential learning to consider the contextual information in the video sequences for better performance. 
\bibliography{mybibfile}

\end{document}